# A Review: Challenges and Opportunities for Artificial Intelligence and Robotics in the Offshore Wind Sector

Daniel Mitchell[1], Jamie Blanche[1], Sam Harper[1], Theodore Lim[1], Ranjeetkumar Gupta[1], Osama Zaki[1], Wenshuo Tang[1], Valentin Robu[1,2,3], Simon Watson[4] and David Flynn[1]

[1]Smart Systems Group, Institute of Sensors, Signals and Systems, School of Engineering and Physical Sciences, Heriot-Watt University, Edinburgh, EH14 4AS, U.K.
[2]Centre for Mathematics and Computer Science, Intelligent and Autonomous Systems Group, CWI, 1098 XG Amsterdam, The Netherlands.
[3]Algorithms Group, Faculty of Electrical Engineering, Mathematics and Computer Science (EEMCS), Delft University of Technology (TU Delft), 2628 XE Delft, The Netherlands.
[4]The University of Manchester, Department of Electrical and Electronic Engineering, Oxford Road, Manchester, M13 9PL.

Corresponding author: Daniel Mitchell (e-mail: dm68@hw.ac.uk).

This work was supported in part by the Offshore Robotics for Certification of Assets (ORCA) Hub under EPSRC Project EP/R026173/1 and EPSRC Holistic Operation and Maintenance for Energy (HOME) for offshore wind farms.

## ABSTRACT

A global trend in increasing wind turbine size and distances from shore is emerging within the rapidly growing offshore wind farm market. In the UK, the offshore wind sector produced its highest amount of electricity in the UK in 2019, a 19.6% increase on the year before. Currently, the UK is set to increase production further, targeting a 74.7% increase of installed turbine capacity as reflected in recent Crown Estate leasing rounds. With such tremendous growth, the sector is now looking to Robotics and Artificial Intelligence (RAI) in order to tackle lifecycle service barriers as to support sustainable and profitable offshore wind energy production. Today, RAI applications are predominately being used to support short term objectives in operation and maintenance. However, moving forward, RAI has the potential to play a critical role throughout the full lifecycle of offshore wind infrastructure, from surveying, planning, design, logistics, operational support, training and decommissioning. This paper presents one of the first systematic reviews of RAI for the offshore renewable energy sector. The state-of-the-art in RAI is analyzed with respect to offshore energy requirements, from both industry and academia, in terms of current and future requirements. Our review also includes a detailed evaluation of investment, regulation and skills development required to support the adoption of RAI. The key trends identified through a detailed analysis of patent and academic publication databases provide insights to barriers such as certification of autonomous platforms for safety compliance and reliability, the need for digital architectures for scalability in autonomous fleets, adaptive mission planning for resilient resident operations and optimization of human machine interaction for trusted partnerships between people and autonomous assistants. Our study concludes with identification of technological priorities and outlines their integration into a new 'symbiotic digital architecture' to deliver the future of offshore wind farm lifecycle management.

## 1 INTRODUCTION

The United Nations Conference of the Parties 26 (COP26) aim is to secure global net zero by mid-century and keep 1.5 degrees within reach. In reaching these targets, countries will need to do the following: accelerate phasing out of coal; curtail deforestation; accelerate a switch to electric vehicles and encourage investments in renewables. Energy Sector activities are clearly associated  been clearly identified as the main cause of global climate change and underpin the current climate change emergency, as acknowledged by the Intergovernmental Panel on Climate Change (UN) [1], [2]. Greenhouse gases, specifically Carbon Dioxide ($CO_2$), are produced during the burning of fossil fuels, which currently represent two thirds of global greenhouse emissions [1], [3]. To meet the global obligation to mitigate the consequences of climate change, there has been significant investment into renewable energy generation [4]. The maturity of offshore wind technologies, combined with global political support for its expansion as part of international governments COVID recovery stimulus, has resulted in this market experiencing unprecedent global growth [5].

The UK represents a world leading energy economy driving the trend of strategic investment into offshore wind [6], where UK renewable electricity capacity reached 47.4GW at the end of 2019, a 3GW, or 6.9%, increase on the previous year. Offshore wind farms also generated 31.9TWh of electricity, which represented a 19.6% increase in comparison to the previous year. The total energy output of the UK in 2019 for onshore and offshore wind energy production accounted for 9.9%, however in Q1 2021, this has increased to 25.6%, demonstrating a shift in energy provenance with a clear trend in favor of wind power [7]–[9]. Additionally, in 2020, the UK produced approximately 6.8GW via operational offshore wind farms, with the future energy generation growth trajectory including pre-planned and, consented projects under construction, expected to reach 27.2GW; an increase of 74.7% of installed turbine capacity. This is attributable to the launch of the fourth round of Crown Estate offshore wind leasing, with a potential of 7GW of new seabed rights offshore of England and Wales to power 6 million homes [10]. Crown Estate Scotland also approved their first round of offshore wind leasing, with the potential to generate in excess of the energy requirements for every Scottish home, offsetting 6 million tonnes of $CO_2$ each year and allowing for the investment of around £8 billion in the Scottish offshore wind sector [6]. On a continental level, the 2050 European Commission agenda estimates that between 250-450GW of offshore wind power will be required to limit global warming to within 1.5°C of pre-industrial levels and estimates that up to 30% in future global electricity demand could be supplied by offshore wind generated electricity [11], [12]. To meet the high end of the 2050 European Commission agenda, this would amount to an additional 40,910 11MW turbines which would be required to cover an area of 75,250km$^2$ with a forecasted capital expenditure for subsea cables reaching $16.1 billion and 2.5 million kilometers by 2030 [13].

Continued growth in offshore wind is dependent on effective policies and licenses, technological advancements and cost reductions in Operation and Maintenance (O&M). The UK has further strengthened its economic position within the offshore wind sector by reinforcing the relationships between revenue support and an active UK supply chain. UK offshore wind projects currently being installed and operated have an estimated 32% UK content, generating £1.8 billion per year, and where this is expected to rise to 65% by 2030, to generate £9.2 billion per year [14]. As a result of the COVID-19 pandemic, investments have been accelerated and increased globally [15]. Due to COVID-19 restrictions, greenhouse gas emissions were projected to drop approximately 6% in 2020 [16]. In November 2020, the UK government detailed a ten-point plan to accelerate a "*Green Industrial Revolution*" [17]. This plan presented the government approach to increase green sector resilience in the post COVID-19 economy, with support for green sector jobs and a roadmap to accelerate the UKs transition to net zero. Previous government support has reduced the cost of offshore wind by two thirds in the last five years, where the projected 2021 plan aims to double the offshore renewable infrastructure. In addition to increasing floating capacity twelvefold to 1GW, and with a commitment to produce 40GW of offshore wind by 2030, these targets would require private investment to reach £20 billion in the UK and will double employment within the sector [18]. The government will also invest £160 million into modern ports and manufacturing infrastructure to ensure the UK is the leader of manufacturing larger wind turbines. This will deliver a projected 60% UK content in offshore wind projects through more stringent requirements for supply chains within the contracts for difference auctions (UK government initiative for supporting net-zero targets). The net result of these initiatives is an increase in global competitiveness and expertise while attracting inward investment for UK manufacturing [17], [19]–[21].

To secure a reliable, affordable and resilient supply from offshore wind farms, offshore infrastructure requires a continuous and complex engineering cycle, associated with inspection, repair, logistics, maintenance and removal of subsystems [22], [23]. There have been a number of advancements in technology which have reduced O&M costs. However, wind farm operators face several challenges which prevent them from achieving their roadmap to an efficient and sustainable offshore wind farm. While developments in Robotics and Artificial Intelligence (RAI) have the potential to positively shape the future offshore wind sector the challenges include:
- **Reduction of O&M costs** - O&M expenditure accounts for up to 25% of the total lifecycle cost of an offshore wind farm; a barrier which needs to be addressed to further develop the offshore wind sector [24]. Financial risks which inhibit the development of O&M include turbine downtime, managing vessels and personnel, hazardous weather conditions, sea state and increasing distances to shore. To reduce costs, new standard operational procedures must be developed for RAI deployment for asset inspection and maintenance. This enhancement in procedures allows for

- **Removal of personnel from hazardous working environments** - The need to develop RAI for offshore wind sector O&M is reflected by the concurrent need to reduce human presence from the offshore environment, minimizing exposure of the human workforce to dangerous weather conditions and sea states [26].
- **Ecological issues in expanding offshore wind** – The deployment of carbon intensive field support vessels including helicopters, crew transfer vessels, heavy jack up vessels and service operation vessels which are used for maintenance of offshore wind farms and as a hub for engineers to live offshore. This leads to disruption and a potential for lasting detrimental effect on the habitats and species which inhabit these areas [27].
- **Recruitment challenges** - Substantial recruitment shortages, which currently exist in the sector at all skill levels due to competition from mainland and other offshore employment sectors [28], [29] In addition to issues relating to training where challenges exist in capturing knowledge effectively to upskill engineers and how technology such as mixed reality devices can be used to create virtual offshore environments where skills can be put to the test in a safe *'fail safe'* environment.
- **Lifecycle challenges** - The decommissioning of WT blades typically occurs after a 25-year lifecycle. The current method of disposal of these complex components is burial in landfill facilities. A key challenge to the renewable energy sector is the upscaling of the recycling process for these materials [30], [31].
- **Emergent challenges** include the reliability of supply chains from shore, as wind farms are increasingly situated further offshore and improvements to efficiency, while reducing risk, via the application of big data analysis methods and the identification of erroneous datasets from offshore systems [32]–[34].

With this complex asset base growing and with distributed examples of how RAI can engage at different points of an offshore windfarm lifecycle, it is now vital that we conduct a detailed review into the opportunities and barriers relating to RAI for offshore wind farm lifecycle management. Our review undertakes both a detailed ground up analysis of existing robotic and AI capabilities, including industrial and academic sources [35]. We also apply a top-down analysis based on the lifecycle needs of offshore wind, as to identify key capabilities, barriers and enablers for RAI in offshore services. Industry within the ORE sector, academia and government agencies all envision RAI as an enabler to transform many current methods and procedures [36]–[38]. This review provides quantitative and semi-qualitative analysis of key research activities and outputs alongside expert analysis over the last 5-10 years. This review highlights that recent advances in autonomy have acted as critical enablers for several commercial platforms. While the hardware for these autonomous operations already existed, the primary advances relate to the novel system integration and data analysis from these technologies. However, enabling autonomy as a service represents a major bottleneck in the deployment of RAI. There is a misconception that robotic platforms are certified as "safe" for their first use [39]. In addition, there is no framework in place that ensures these platforms are safe to use throughout their lifecycle. Therefore, the development of quantitative methodologies for the assessment of the reliability and resilience of a robotic platform, via self-certification, will be a key metric in future successful deployments. This will significantly advance the roadmap for a trusted autonomous offshore energy field operated, inspected and maintained from onshore monitoring facilities via resident robotics [40].

The remainder of this review is structured as follows. Section 2 presents an overview of current offshore wind farms. Section 3 provides critical analysis of the required improvements in the Offshore Renewable Energy (ORE) sector due to RAI for each of the lifecycle functions. The state of the art in RAI is discussed next where advantages and disadvantages for robots in air, land and sea are discussed in Section 4. Analysis of the commercial and patent database is presented in Section 5 using Scopus and Espacenet. A review of academic literature is used to identify trends in research with results discussed in Section 6. The capability challenges based on industrial insights is discussed in Section 7. Section 8 highlights a roadmap to resilient infrastructure offshore. Section 9 proposes viewpoint of future infrastructure by highlighting a digital architecture for the orchestration of resident multi-robot fleets and RAI offshore. This is followed by a summary of key findings in Section 10 with conclusions presented in Section 11.

## 2  OFFSHORE RENEWABLE ENERGY OVERVIEW

The offshore wind energy sector is undergoing transformative changes in the way that energy is collected and transferred to the National Grid. To date, this has relied on technological advances in wind turbine design. The amount of power a wind turbine can generate depends on the wind speed and the swept area of the blades. Therefore, increasing the size of a wind turbine has significant benefits for energy harvesting. In 2021, it was calculated that it would take 1200 turbines to make 10GW of energy. By 2030, it is estimated that this will reduce to 800 wind turbines due to the increases in wind turbine size [41].

Floating wind technology allows for offshore turbines to be positioned further from the shoreline in areas with deeper waters and steadier, stronger winds [41], [42]. The UK government set a target of 1GW of floating wind capacity by 2030, however the Carbon Trust estimates that this figure could double by 2035 due to increased benefits and reduced costs [41]. An overview of key advances in offshore wind farms is summarized in Figure 1 where current offshore wind farms are positioned close to the shoreline with small-medium power capacities. By 2024, it is expected that much larger 14MW turbines will be positioned offshore at larger distances from the shoreline and in deeper waters.

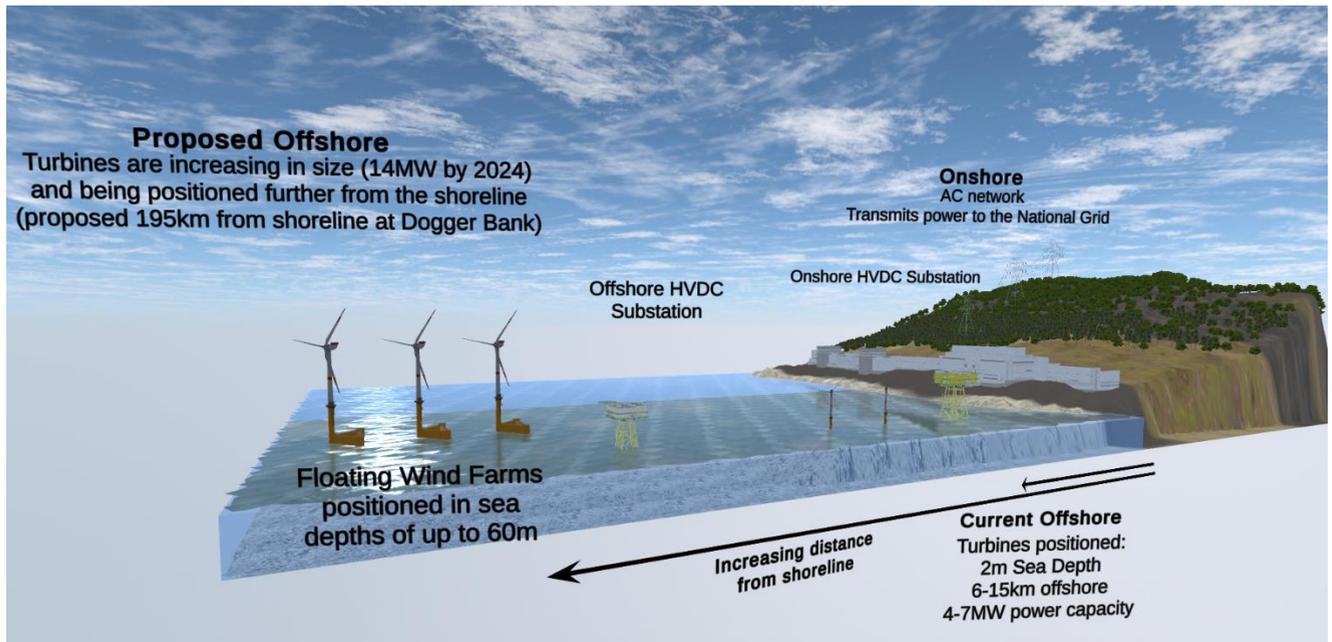

**Figure 1** An overview of key advances within an offshore wind farm [43]–[45].

The Crown Estate outlines their summary of the primary construction phases of an offshore wind farm until the commissioned in the following phases [46]:
1. Development - The Crown Estate has awarded the developer the development site.
2. Planning – The developer has submitted the planning consent for evaluation.
3. Contracts for difference (eligibility and secured)- Includes projects that have received planning consent and have a grid connection date agreed with the National Grid.
4. Pre-Construction – Planning permission approved but the project has not entered the construction yet.
5. Under Construction – Offshore construction has begun.
6. Operational - The offshore wind farm has been commissioned and is generating power.

The timeframes and current phases of several proposed and under construction offshore wind farms are displayed within Figure 2. This data has been merged with data collected in 2016 and 2020 to give an increased view of the timelines provided by Renewables UK [47], [48]. Data for the contracts for difference and operational phases was unavailable. Figure 3 charts the four remaining stages using the merged data from Figure 2, with the duration assigned for each phase of the wind farm project. The average of each of the four stages for each wind farm development is then represented and shows a 277% increase in the efficiency of the 'In Development' phase over 2016-2020 period. Industry has focused on optimizing O&M procedures but from this data, it can be identified that there is also an opportunity in improving and reducing the duration of the in planning, pre-construction and under construction phases. Especially with more ecological sensitive and less carbon intensive field support vessels and technologies.

Figure 3 depicts an improvement in the 'In Development' phase of an offshore wind farm. However, there are little/zero improvements in the other lifecycle phases. This could be attributed to offshore wind farms being positioned in more complex areas, further from the shoreline and with larger wind turbines. Therefore, more work is required during these phases due to increased transportation durations, difficulty during installation due to deeper waters and more treacherous conditions further offshore. A second point includes the 'In Planning' and 'In Development' phases taking up ~40% of wind farm deployment duration. This represents an opportunity for improvement within this lifecycle phase of an offshore wind farm.

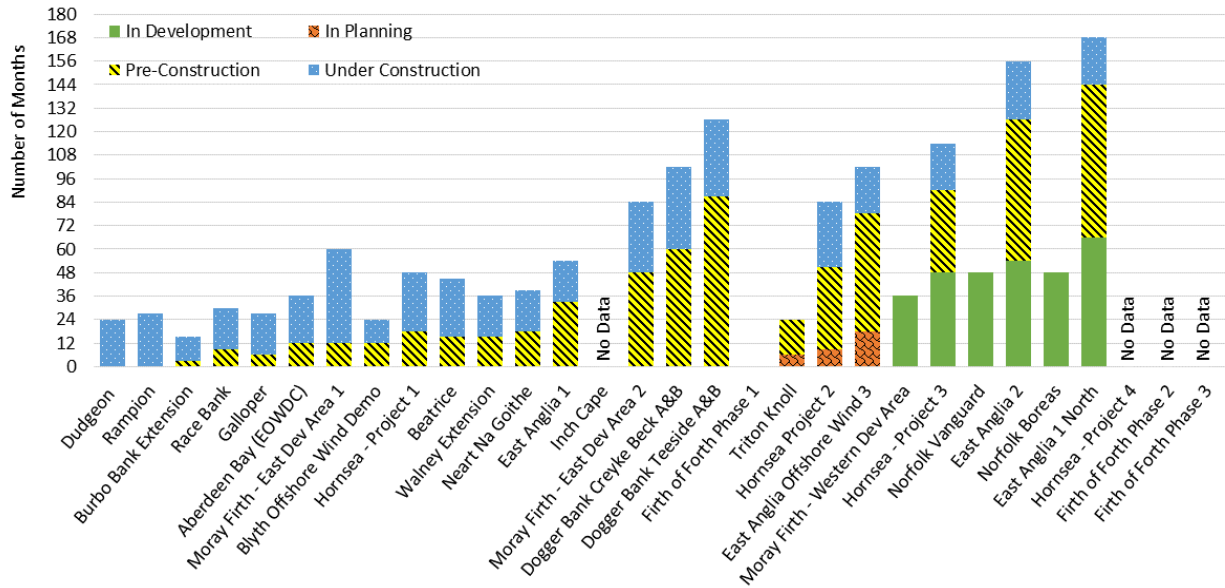

**Figure 2** Timeframes and current phase of offshore wind farm projects (2020) [47], [48].

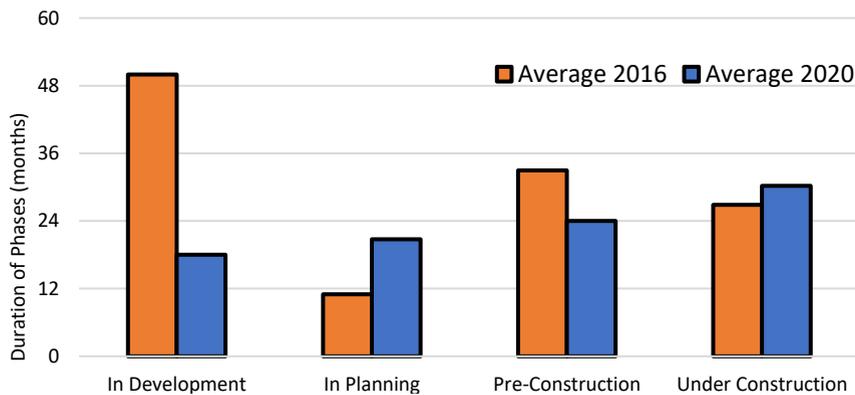

**Figure 3** Average phases of the lifecycle of an offshore wind farm [47], [48].

## 2.1 Investment

The approach taken by government, academia and industry has propelled the offshore wind energy sector into significant growth. This has been due to substantial investment and funding opportunities available throughout the UK. Between 2016 and 2021, approximately £19 billion has been invested in the ORE sector in the UK. This investment has contributed to jobs across the UK in manufacturing, project development, construction and operations [49]. The offshore wind sector published plans which would generate thousands of highly skilled jobs across the UK and to establish offshore wind as the core of reliable, clean and affordable energy systems. In November 2020, the UK Government pledged that 30 million homes in the UK will be powered by offshore wind by 2030 [50], with the aim of ensuring the UK maximizes the advantages for UK industry from the global shift to clean energy growth. The first round of future contracts for difference opened in May 2019 and will support £557 million of investment with the next round being supported in 2021 [20].

Investment in offshore wind energy will quadruple the level of global energy investment, despite the onset of the COVID-19 global pandemic, which resulted in a net deacceleration of the global economy. The report identified that 28 new offshore wind farms were approved with a total worth of £28 billion for the year 2020. This represents well above the total investment for 2019 and a fourfold increase compared to the first half of 2019 [15]. This accelerated economic growth is due to many governments implementing green recovery packages as part of their post COVID-19 recovery strategies. Further investment has been committed globally by the Iberdrola group plan to invest £67.2 billion into renewable energy generation worldwide. Additionally, Scottish Power pledged to invest £10 billion over the next five years (2020-2025) to aid the UKs green recovery plans and will double the renewable generation capacity of Scottish Power, adding a further 2.4GW of renewable technology to its portfolio by the end of the investment period in 2025 [50].

The Offshore Wind Growth Partnership, in partnership with the Offshore Wind Industry Council, has played a key role in the sector deal between industry and government and will install 30GW of offshore wind by 2030. This includes an investment of up to £100 million in a new industry program conceived to help expand UK business in the rapidly growing global offshore wind market [51], [52].

*2.2     The Primary Lifecycle and Support Functions of Offshore Windfarm*

The Crown Estate, industry and academia have identified key lifecycle stages, as highlighted in phases 1-6 (page 5). However, a shift in the wider approach is required to reflect the full lifecycle of an offshore wind farm array. This will optimize operations throughout the entire lifecycle, from commissioning to decommissioning, as the sector matures. This review highlights areas which will retain focus and other areas which require to be considered in more detailed throughout the lifecycle of an offshore wind farm. These include several integrated and complex phases as summarized in Figure 4.

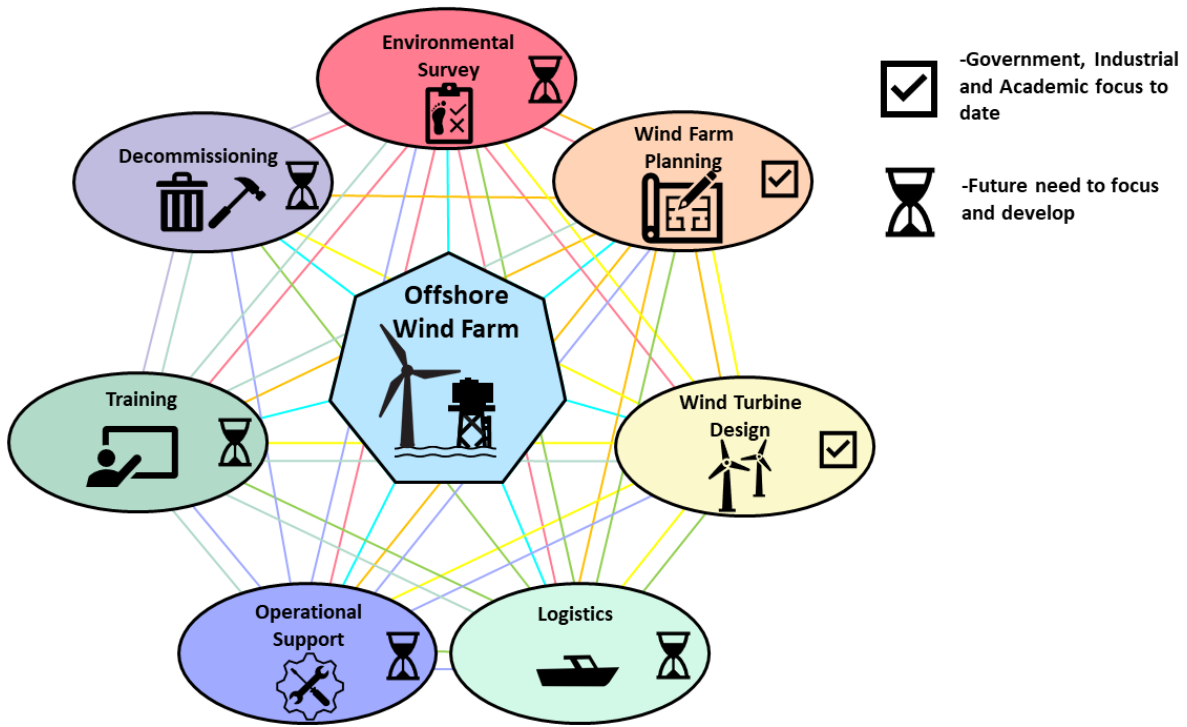

Figure 4 The primary lifecycle and support functions of an offshore wind farm.

2.2.1     ENVIRONMENTAL SURVEY

When developing a proposal for an offshore wind farm, careful considerations are made for all aspects of planning. Development and consenting services are undertaken until financial close or full commitment from the operators to begin construction. This can only be achieved once all surveys, assessments and tests have been completed and include environmental surveys, metocean assessment and geological and hydrographical surveys to minimize the risk to any sea life that inhabit the waters during construction [46].

Offshore wind farm installations have the potential to cause a detrimental effect on wildlife due to inconsiderate man-made construction and operation. There is a drive for a more ecologically strategy via the development of sustainability-centric planning. To reduce all negative effects, assessments on the physical, biological and human environment during the lifecycle of a wind farm are undertaken. The environmental survey stage includes an evaluation of the impact of the wind farm at the intended location and surrounding areas. This includes ornithological studies, studies of benthic species within sediment and seabed, fish and shellfish studies and marine mammal surveys to establish the diversity, abundance, distribution and behavior of local wildlife species. Although there are several negative implications, there are also notable environmental benefits, such as shellfish utilizing wind turbine foundations as artificial reefs which create ecosystems for multiple species and the animals which prey on them [27].

### 2.2.2 WIND FARM PLANNING

To ensure a successful and seamless construction phase, a well-ordered plan must be in place for the installation of an offshore wind farm. These plans must consider all aspects of the lifecycle and support functions and can be further summarized to consider features of a wind farm array such as grid connection, subsea cabling, foundations, number of turbines, installation companies, substations, operational base and operational port. The planning phase defines the full scope of the project, outlining clear objectives, achieving a minimal impact on the environment, a high standard of design, optimum levels of safety and a recovery plan to encourage wildlife after commissioning.

Wind farm operators are required to know the depths of the seabed and identify any areas which may be considered dangerous to ensure foundations are properly and safely fabricated. Guidance for geophysical survey of unexploded ordnance and boulders was recently published by the Carbon Trust Offshore Wind Accelerator [53]. This guidance supports subsea cable installation by improving the effectiveness and efficiency of surveys. Two 500lb wartime explosive devices were discovered in the surrounding area of Rampion wind farm during construction in 2016 and were retrieved and disposed of by E.On [54]. In the previous year, seven unexploded devices were discovered on the seabed at the location of a planned offshore wind farm in the wash of the Lincolnshire coast [55]. To ensure the safe construction and commissioning of future offshore wind farms, in unexplored, deeper waters, planning is an essential phase of the lifecycle and support functions of an offshore wind farm. This highlights the requirement to accurately utilize data from geophysical surveys in the planning stages to form a knowledge base for risk evaluation and management of these operations, which can reduce life-threatening and costly accidents. This is an ongoing risk when working offshore as submerged obstacles and threats can be located, including unexploded wartime ordnance and shipwrecks [56].

### 2.2.3 WIND TURBINE DESIGN

The global requirement to develop clean and reliable energy sources is a key driver for the evolution of wind turbine design. The state-of-the-art in wind turbine development is manifested in high-pace competition between large, multinational manufacturers and represents edge technology refinements in mechanical and electrical designs to ensure the successful and reliable, long-term operation of a wind turbine asset. The cost of offshore wind has declined from £150 per Megawatt hour (MWh) to £40 and, as a result, wind turbine designs have evolved to capture higher windspeeds further offshore. However, factors that have limited development further offshore include safe remote operation, cost of logistics, harsher weather conditions and deeper waters. Several risks and challenges result from operations further offshore, however, more efficient energy generation and more affordable electricity for consumers remain key drivers for continued development [57]–[59]. While wind turbine design has evolved significantly, the stringent specifications for regulating the quality of power to the grid has remained the same and address aspects including flicker and voltage change, steady state operation and control systems to ensure a reliable connection [46], [60].

Competition between major wind turbine manufacturers, such as GE Renewable Energy, MHI Vestas and Siemens Gamesa, has driven the rapid pace of efficiency and reliability improvements of offshore wind turbines. Resulting in technology development of increasing blade lengths to capture larger swept areas of wind, in turn achieving larger generated power capacities, as detailed in Table 1. A by-product of this development is the need to situate larger turbine structures further offshore, exposing these highly sophisticated assets to harsher conditions, in deeper waters and increased average windspeeds, resulting in a new generation of floating wind turbines with the capability to meet the requirements for sustained operation further offshore [58], [59].

**Table 1** Wind turbine manufacturers and the increasing size of wind turbines with respect to date.

| Company [Reference] | Offshore Turbine Name | Current Offshore Distance from Shoreline UK | Tower Height (m) | Blade Length (m) | Swept Area (m²) | Power Capacity (MW) | Serial Production |
|---|---|---|---|---|---|---|---|
| Siemens Gamesa [61] | SWT-3.6-120 Offshore | 6 km (Burbo Bank) [44] | 90 | 58.5 | 11,300 | 3.6 | 2007 |
| Siemens Gamesa [62] | SWT-7.0-154 | 13km (Beatrice) [63] | 90 | 75 | 18,600 | 7 | 2017 |
| GE Renewable Energy [64] | Haliade-X 12 | 195 km (Teesside A at Dogger Bank) | 260 | 107 | 38,000 | 12 | 2019 |
| MHI Vestas [65] | V174-9.5 | N/A | 110 | 85 | 23,779 | 9.5 | 2019 |
| Siemens Gamesa [66] | SG 8.0-167 DD | 89 km at Hornsea 2 | Site Specific | 81.4 | 21,900 | 8.0 | 2019 |
| GE Renewable Energy [67] | Haliade-X 13 | 195km (Proposed 190 units at Dogger Bank [43]) | 248 | 107 | 38,000 | 13 | 2020 |
| MHI Vestas [65] | V164-10.0 MW | 6 km (Burbo Bank) | N/A | N/A | 21,124 | 10 | 2021 |
| Siemens Gamesa [68] | SG 11.0-200 DD | N/A | Site specific | 97 | 31,400 | 11 | 2022 |
| Siemens Gamesa [69] | SG-222 DD | 195 km (Teesside B at Dogger Bank) | Site Specific | 108 | 39,000 | 14 | 2024 |

The constraints placed on offshore wind turbine design differs to that of onshore designs and allows for designs focused on larger, more efficient and powerful wind turbines. However, the pace of advancement in design has not been met with commensurate improvements in inspection procedures. The standard approach to wind turbine inspection has consisted of personnel and remote-controlled robots overcoming the barriers of height and weather conditions to complete inspections [70], [71]. This initial challenge must be overcome to access areas at height such as substation, nacelles, tower and blade.

### 2.2.4 LOGISTICS

To overcome the difficulties of the offshore environment, offshore wind farm operators require a modern logistical service system. Logistics in an offshore wind farm includes organization and movement of equipment or personnel from the shoreline to the offshore site. Current offshore logistics solutions and supporting infrastructure within the lifecycle of offshore wind farms is illustrated in Figure 5. The offshore lifecycle is heavily reliant on installation vessels (A), crew transfer vessels (B), subsea divers (C) and helicopters (D) throughout the phases from the environmental survey to decommissioning [72]. The main challenge in any heavy lift vessel is finding the correct vessel and preparing it by ensuring it is equipped for the task. In addition, skilled technicians are crucial to operate the equipment.

An offshore commute often includes rough sea states for up to four hours before the work begins, therefore, Service Operations Vessels (SOV) are utilized and act as a floating hub for any offshore technicians to live on and to store spare parts for a wind farm array. An SOV can hold up to 88 technicians living there for 4 weeks whilst completing maintenance on a wind farm. An SOV typically includes a gangway system where the effects of the waves are counterbalanced to provide safe and rapid access for offshore technicians to complete remedial maintenance [73]. Helicopters can also be utilized for rapid deployment or in emergencies which require medical treatment. This aids in overcoming sea troubles to maximize yield, however, all logistics vehicles are very expensive to deploy and so should be deployed strategically to minimize this cost [74]. In addition, these methods are expensive, carbon intensive and present several risks to the personnel deployed in the hazardous offshore environment. Therefore, it is necessary to reduce the carbon footprint of offshore logistics vessels to more efficient vessels which are battery or hydrogen powered.

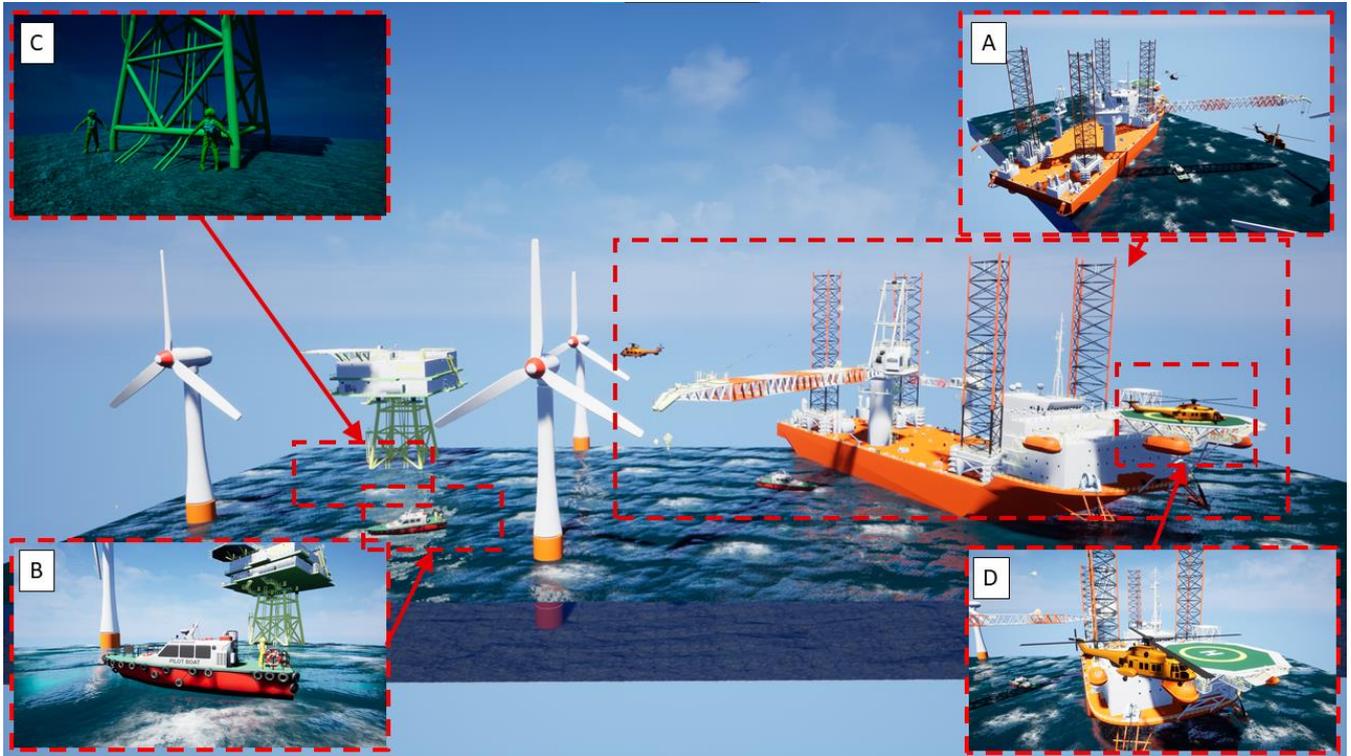

**Figure 5** Current assets and the supporting infrastructure within the lifecycle of an offshore wind farm array.

2.2.5   OPERATIONAL SUPPORT

The offshore wind sector has evolved through many primary lifecycle functions (Figure 4) due to improvements from a deepening of knowledge. Research is being completed to improve O&M procedures to make improved decisions for the remaining useful life of an asset to reduce costs and risk [75], [76]. Operational support remains a large proportion of a wind farm and represents not only visual aspects such as the wind turbine, but also subsea cables, foundations, substation and cables on land which connect to the national grid. For example, for a 300-MW wind farm, loss of revenue from a power outage due to a fault in one subsea cable is approximately £5.4 million per month, and the cost of locating and replacing a section of damaged subsea cable can vary from £0.6 million to £1.2 million [77]–[82].

2.2.6   TRAINING

Safety is paramount in harsh environments. When serious injuries occur in an offshore setting, time is a critical factor and can make the difference between life and death. First aiders and first responders are required to maintain control over a situation when under severe pressure to perform lifesaving actions. Over forty paramedics from the north-east of England completed an intensive working at height and rescue training course, delivered using a 27m high wind turbine training tower operated by the ORE Catapult at National Renewable Energy Centre in Blyth, UK. This training included techniques for self-recovery and casualty recovery, and emergency procedures when using vertical ladders and fixed vertical fall arrest systems [83]. Training at these facilities ensures high levels professional capability gained from learnings attained from previous accidents and offshore emergencies.

The AI Sector Deal introduces several measures to ensure the primary role of offshore wind energy in the future. These statistics are presented in Figure 6 and aims to increase the skilled workforce from 10,000 in 2017 to around 36,000 by 2032 [36]. This increase will be primarily due to an expected threefold increase in the number of turbines, from 1660 to potentially 5,358 by 2032, and with an expected fivefold increase in installed capacity from 6.4GW to 35GW over the same period [84]. There are also many transferrable skills from other sectors, where the main entry routes into the industry will continue to be from technically related industries, such as the armed forces and those from the wider energy sector. Apprentices and graduates will provide the foundations of new experience for future generations and people with cross-sector skills such as commercial, IT, data analytics and ROV operators of drones and subsea vessels [84]. It is likely that the UK will be short of 20,000 engineering graduates per year to meet this demand for a green future [84], [85]. For the future workforce, it was identified that 400,000 new energy workers will be required to ensure that the UK reaches net zero. This will allow for opportunities for skilled tradespeople, engineers and other specialists across the country to help with installing low-carbon heating and the development of new technologies [86]. This will require government, industry and academic institutions to promote STEM educational progression, opportunities and clear career pathways. This includes programs to develop curriculums, increase job mobility between sectors and increase apprenticeship opportunities [20]. This must be facilitated by a partnership between these institutions to create effective engagement for students. They must also develop innovative methods to attract a diverse range of backgrounds. To create and encourage investment in training and skills throughout the supply chain for the existing workforce, a procurement skills accord must be created. To facilitate the flow of skills between technologies, companies and other sectors, the implementation of common training standards must also be implemented. This would generate and encourage various pathways for individual development [84].

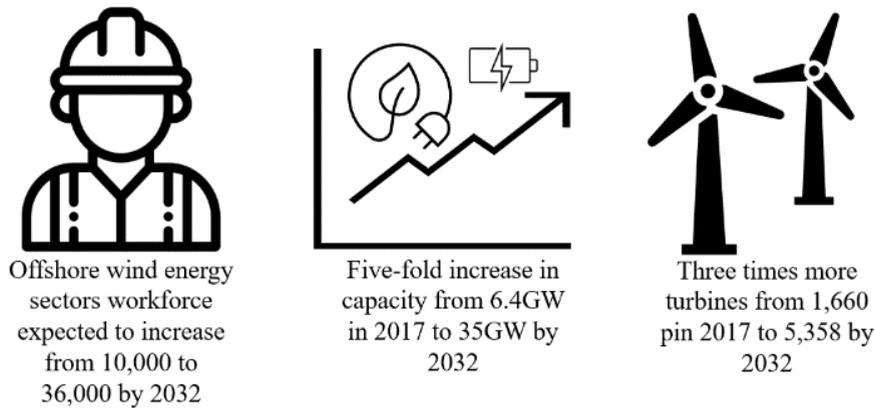

**Figure 6** Key statistics within the offshore wind energy sector [84].

### 2.2.7 DECOMMISSIONING

The decommissioning of offshore wind farms is a relatively new concept and is a growing market as wind farms reach the end of their lifecycle. Although well-structured plans are in place to perform decommissioning, engineers will continuously be learning and improving processes. To date, the decommissioning process for offshore wind turbines has only been performed at depths of less than 50m, representing a learning opportunity for decommissioning teams in less difficult and hazardous environments nearer shore [87]. For wind farms at the end of their planned lifecycle, there is the potential for reuse of key components, therefore complete dismantling and disposal is not necessarily the first option. This includes the refurbishment of minor components, such as drivetrain or rotor and, where viable, reuse of major components, such as the tower, foundations and cables. This allows existing projects to extend remaining useful life and increase energy production. The implementation of wind turbine decommissioning represents a new milestone for the offshore wind sector and consists of many of the learnings from lifecycle development of wind farm arrays. However, with the first farms decommissioned in 2016, the process requires development, with many processes still to be standardized [87].

Current wind turbine design, despite their role in clean energy production, create many tons of waste per turbine at the end of their lifecycle. While approximately 85% of turbine components are highly recyclable, the predominantly fiberglass-constructed blades present difficulties for sustainable disposal [30]. Between September 2019 and March 2020, 1000 end of lifecycle fiberglass wind turbine blades were retired to for burial at a landfill facility at Casper, Wyoming, USA (Figure 7). In general, each blade element (shear webs, load-carrying beams, leading and trailing edges, and the aerodynamic shell) are manufactured into a single-piece component. This construction process results in difficulties in component separation for disposal and typically requires a diamond-tipped saw blade with sufficient water cooling [88]. Consequently, the strength of these materials, so well suited for asset longevity during the energy production phase, makes them very difficult to recycle. To address this, companies such as Global Fiberglass Solutions offer green-product manufacturing and fiberglass recycling [89]. However, the relative availability of these companies are currently insufficient to cope with the future demand for decommissioning and waste disposal. The current growth in wind farm are being installed construction will lead to an increase in decommissioning offset by the life expectancy of a wand farm (currently 20-25 years. This will require significant upscaling of the methods employed by companies, such as Global Fiberglass Solutions, to cope with the global demand for waste disposal. Global Fiberglass Solutions, located in Bellevue, Washington, USA, transform fiberglass composites into small pellets for conversion into injectable plastics or waterproof sheets, which can be re-utilized within new construction projects [31], [89]. Pyrolysis represents an alternative for the recycling of wind turbine blades, where the blades are sliced up and then placed in ovens at 450-700 Celsius. The resulting material can be used for glue, paint and concrete. Bi-products of the pyrolytic process include syngas, a fuel source for combustion engines, and charcoal, which can be used as fertilizer [31].

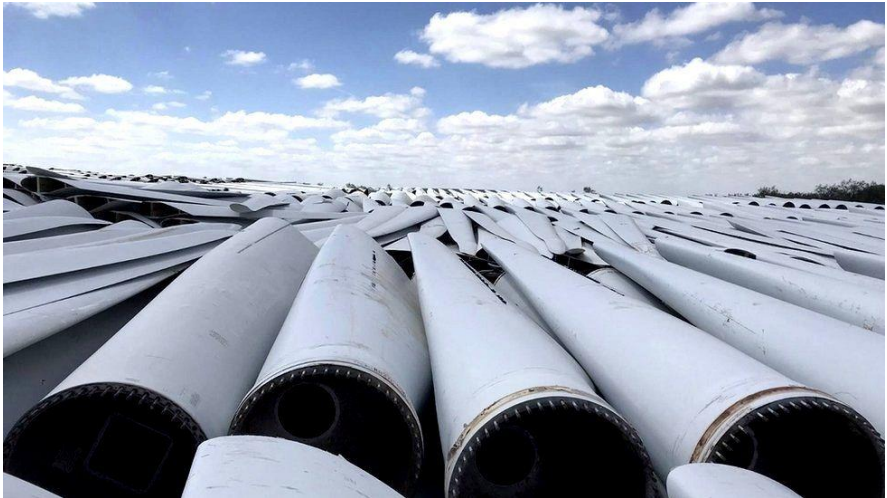

**Figure 7** Decommissioned wind turbine blades awaiting burial at Casper, Wyoming [31].

## 3 CRITICAL ANALYSIS OF THE REQUIRED IMPROVMENTS IN ORE DUE TO RAI

The offshore wind sector has created several opportunities and benefits, however, there are several outcomes which result in barriers and challenges which need to be addressed. Currently offshore O&M activities amount to 25-30% of the total lifecycle costs of offshore windfarms [90]. According to the Crown Estate, 80% of the cost of offshore O&M is a function of accessibility during inspection resulting in the requirement to get engineers to remote areas in the field. Therefore, minimizing the need for human intervention offshore is key to minimize cost whilst maximizing potential. The solution/demand to this problem includes readily deployable, resident offshore BVLOS robotic assistants [91]. As identified previously, there remains opportunities for wind farm operators to improve certain phases of the development of an offshore wind farm. In addition, via the admission of the UK government itself, that the UK faces a shortfall of 20,000 engineering graduates each year to ensure the UK reaches net zero by 2050 [92], [93]. This is made more difficult as there is sector wide competition for engineers.

To support the future optimization of offshore wind farms, the role of RAI must be considered within, and across, each of the strategic phases of wind farm development. Robotics and artificial intelligence are considered as two enabling elements of autonomy. For the example of the ORE Sector, a robot is a physical device which navigates around an offshore wind farm with an objective to gather data or interact with the environment. The level of autonomy of the robotic platform can vary from teleoperated to fully autonomous, however the objective remains the same. AI techniques are commonly implemented onboard a mobile robot or computer and used for the data analysis. Therefore, a robot collects the data and AI analyzes the data. This is how the combination of robotics and artificial intelligence can enhance operations significantly. Big data analytics results in the combination of all functions, assets and systems, where data streams create a whole-systems approach, with the potential for increased activity.

It is important to emphasize that RAI represents an additional tool to be used by engineers to ensure safe practice. The expertise from these skilled engineers will still be required over the lifetime of a wind farm, albeit retraining will be required to support post-processing of data collected from inspections. The aim of these procedures is to support monitoring and extend the lifetime of the wind farm array by utilizing a wide range of sensing and inspection methods to optimize electricity generation whilst ensuring safe operations [46]. Figure 8 highlights the critical analysis and demand for RAI within offshore windfarms for each of the lifecycle phases identified.

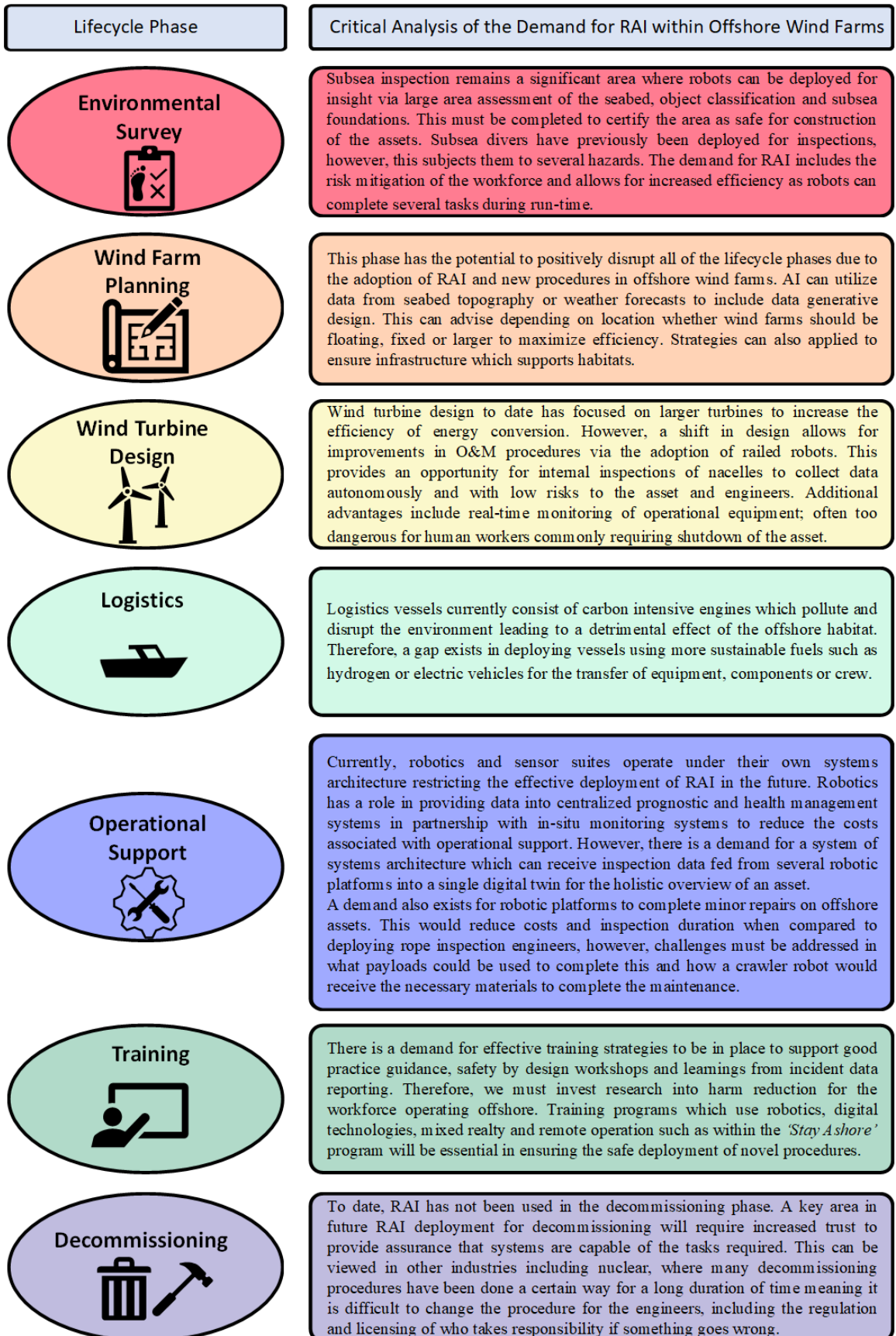

**Figure 8.** Critical analysis of the demand for RAI in Offshore wind farms per each lifecycle phase [94]–[98].

### 3.1.1 REALIZING THE POTENTIAL OF RAI AND EARLY ADOPTER INITIATIVES

The potential of RAI is being captured by early adopters within industry. The uptake of RAI is minimal currently, however is on the increase as technology advances. Early adopters of RAI include the nuclear and manufacturing sectors. The Siemens Electronic Works plant in Amberg, Germany represents a showcase *'smart factory'* via the implementation of digital twins to accelerate product design and manufacturing alongside the workforce. The facility manufactures more than 1200 different products, resulting in a production line that changes configuration approximately 350 times per day. With more than 75% of the facility working autonomously where manufacturing is 5-10 times faster and with 99.5% reliability to meet 24-hour delivery times. Due to the significant level of automation, the employee roles within the factory have changed, yet the factory still employs 1200 people who maintain the robots, write code and run simulations of new products in the digital twin [99]–[101].

Every country will be affected differently with the implementation of RAI, with each country having different local and regional economic concerns. The transition of RAI into the workforce leaves employers vulnerable if not executed with the correct level of diligence. It is clear, however, that not all employers and employees will be affected in the same way. An OECD report '*Policy on the Future of Work*', has suggested that significant imbalances may occur for those that prosper and those left at an economic disadvantage by the rise of robotic automation in the workplace [102]. The report highlights that low skilled and young people are at most risk, with those remaining in work experiencing a resultant reduction in salaries. In rural areas, robotics will be a significant boost to local economies, with job creations in various support activities and allied sectors. The fear of robot-induced unemployment may worry many in cities and suburbs, however, the roles of the workforce will change to manage the upkeep of the robots. This reflects a dual-transition which is occurring currently in the expansion of the offshore wind sector alongside the challenges of adapting and implementing new roles and skills within the offshore workforce. A portion of these skills must pertain to energy and AI within this sector and include the training of robot technicians during the concurrent development of robust and resilient robotics deployed offshore to support the infrastructure. The deployment of RAI serves to reduce the inherent risks to personnel operating in the hazardous offshore environment whilst also having a direct positive impact on productivity.

## 4 STATE OF THE ART ROBOTICS AND ARTIFICIAL INTELLIGENCE

To support our analysis of the current state of the art in RAI, we have structured our analysis to decouple robotic platform type, application, level of autonomy and influencing trends from alternative industrial applications for offshore robotics. An in-depth analysis of academic and commercial state of the art in RAI is performed using a criteria that allows us to partition the application, level of autonomy, underpinning business case the following criteria of assessment;

**Service Robot Trends** – Identification of adoption trends in robots with respect to the most popular type of deployment. This provides insight to both market acceptance type of robotic platforms as well as the business case..
**The Spectrum of Autonomy** – Different robotic platforms have different levels of autonomy driven by functionality and application dependent constraints e.g. risk (safety case), environment. This reveals trends in terms of levels of autonomy, thereby levels of human engagement and/or intervention, and varying attitudes towards autonomy.
**Service Robots** – This provides an overview of the current capabilities, (land, sea and air), in offshore robotics including function/role within the lifecycle of an offshore wind farm.
**Hype Curve of Robotics –** This allows for an informed analysis of robotic platform maturity, adoption, and acceptance into specific markets and applications.

### 4.1 Service Robot Trends
Offshore environments present several key challenges for the offshore wind sector, however, the rapid development of RAI will enable significant advancement in RAI autonomy as a service across multiple sectors. The ORE sector continues to benefit from the myriad of markets influenced by RAI, such as automotive, manufacturing and logistics. This has enabled RAI development to be well positioned to adapt and upscale system manufacturing to set the standard for global growing demand in wind turbines and RAI required to minimize costs. For example, remote inspection operations conducted by robotic systems on offshore facilities will require the collaboration of several different robotic platforms; comprising of UAVs, crawlers, ASVs and AUVs.

When considering service robots, at 41%, autonomous guided vehicles represent the largest proportion of all units sold. This type is well established in non-manufacturing environments, completing tasks within logistics, and has vast potential in manufacturing. At 39%, inspection and maintenance robots represent the second largest category of all units sold and cover a wide range of robot types, from low-priced units to expensive custom solutions. Service robots for defense applications accounted for 5% of the total number of service robots sold in 2018, with UAVs representing the highest share [103]. Figure 9 highlights the expected increasing trend of service robots, where the leading applications are within logistics, inspections and defense sectors respectively. These statistics highlight a significant year-on-year increase in service robots being used as tools across industry sectors. The relatively slow adoption of service robots within the defense sector indicates the more stringent nature of their regulatory framework and protocols. This is due to the sensitivity of hardware, software and assets, highlighting the requirement of fully trusted autonomy and secure robotic platforms.

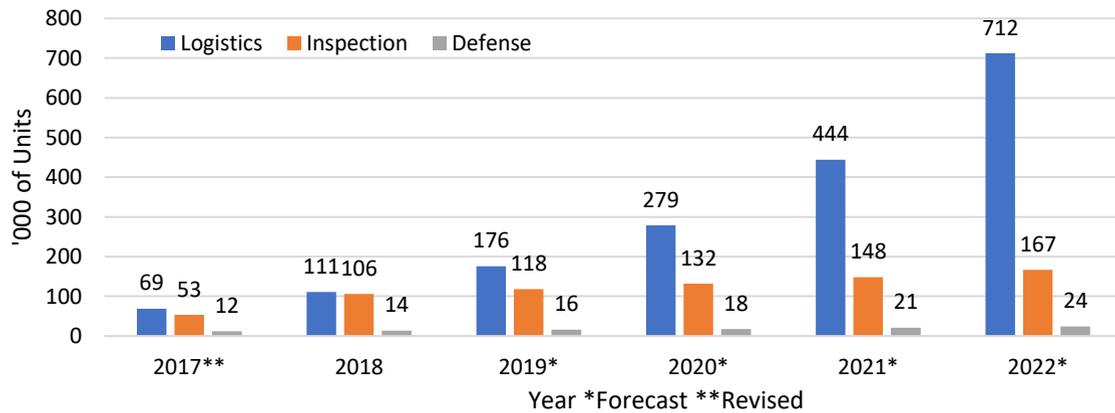

Figure 9 **Service robots for industrial applications** [104].

### 4.2 The Spectrum of Autonomy

The term '*autonomous system*' is widely used within academia and industry. However, this is a term that requires a concise definition, depending on the scenario and application, such as software systems, robotics, AI and autonomous automobiles. In this instance, a full and trusted autonomous system is defined as a system with the capability to complete autonomous operation with independent onboard decision making while adhering to a set of rules which ensure safety and self-certification [105]. Trust is also a developmental area of research which is extremely important. An operator must be able to trust the autonomous systems onboard a robotic platform when working in shared workspaces and trust that a robot will operate as intended during a BVLOS mission.

The following levels of autonomy can be viewed in Figure 10 and are discussed in more detail in the following:

**Level 0 - No Autonomy -** A robotic platform must be teleoperated via a control system or remote control. The human operator is responsible for the safe operation of the robotic platform. These typically operate within Visual Line of Sight (VLOS) conditions.

**Level 1 - Operational Assistance -** The robot is again deployed via teleoperation, however, there are minor safety features which may consist of collision avoidance (SLAM) or automatic emergency stop.

**Level 2 - Partial Automation -** The robot may be considered as semi-autonomous at this stage. The human operator is responsible for the safety of the robot; however, the robot can perform some autonomous tasks with human supervision.

**Level 3 - Conditional Operation -** The robot can perform under well-defined variables. This is currently applied to autonomous automobiles on motorways or within restricted *'robot only'* areas within warehouses. The robots can operate autonomously in these scenarios as the likelihood of several risks are minimized, such as unpredictable events.

**Level 4 - High Automation -** The robot can operate fully autonomously in many scenarios. The robot can also provide suggestions for mission optimization or to ensure that safety governance is achieved. The robot has a larger amount of control over the previous autonomous systems, however, a human operator can retrieve control of the robotic platform at any point.

**Level 5 – Full and Trusted Autonomy -** The human operator can assign a robot a mission and the robot can ensure the mission is achieved in a safe and self-optimized manner. This includes the robot overcoming unforeseen challenges and events, which may otherwise impact the capacity of the robot to complete a set task. The robot must have the ability to make all decisions onboard and, in extreme cases, defer the decision making to the human authority.

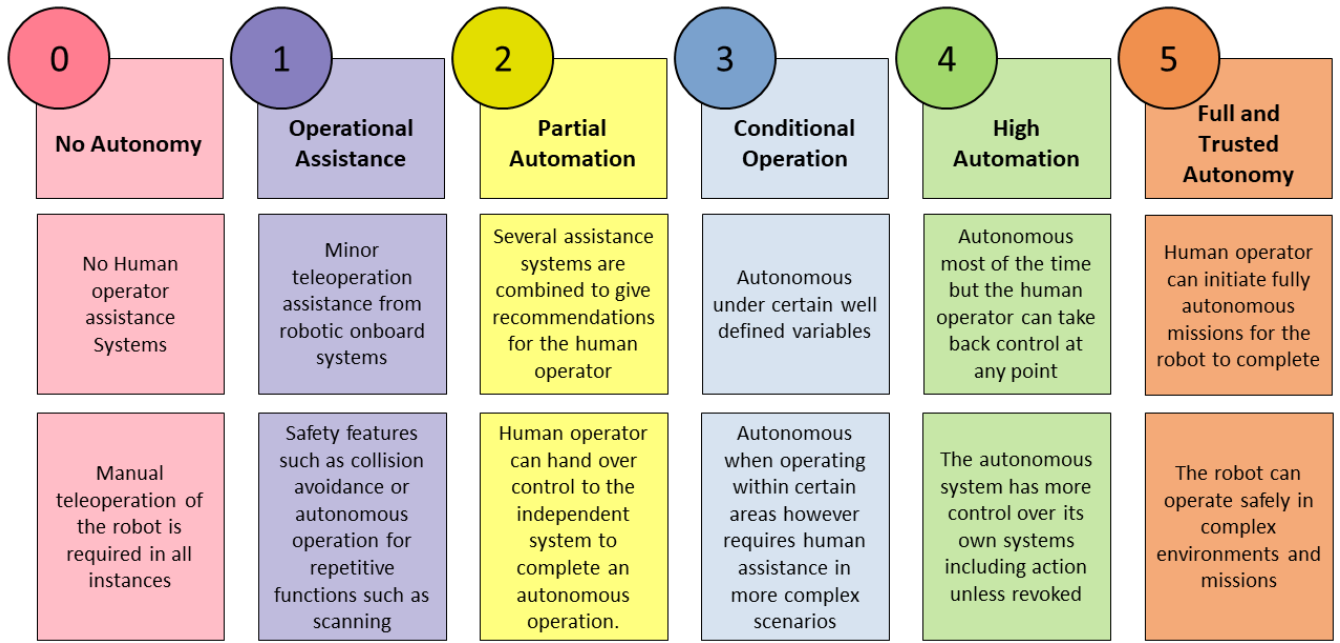

**Figure 10** **The spectrum of autonomy** [105], [106].

### 4.3   Robotics in the Field

To date, underwater inspections typically been conducted via towed array sonars coupled to a surface vessel. The Hugin is an Autonomous Underwater Vessel (AUV), which can automate the inspection process (Table 2, Figure 11) [107], [108]. However, whilst the Hugin is an excellent inspection platform, this AUV is unable to perform any activities that require manipulation, which could include repairs or maintenance tasks. Though, the AUV can be maneuvered effectively, a manipulator would allow sensing payloads to be positioned within confined underwater spaces that prohibit access to the large Hugin AUV. This AUV platform can also be deployed in rough seas (up to sea state 5). However, to increase productivity during the deployment and recovery phases of the platform, this should be improved for operation in higher sea states [109]. No indication of deployed performance factors or case studies were reported for the Hugin at the time of writing. Robotic platforms such as Hugin AUV, will ensure this process becomes more automated via semi to persistent autonomy.

**Table 2** A list of Hugin AUV robotic capabilities, applications, key features and tasks.

| Application | Planning | Development | Maintenance | Array Assessment | Lifecycle Stages |
|---|---|---|---|---|---|
| **Key Features** | Stability | Maneuverability | Battery Endurance | Operating Depth | Flexible Payload |
| | | | 100 hours at 4 knots | 6000m | Capacity |
| **Tasks** | High Resolution Seabed Mapping | Geophysical Site Inspection | Oceanographic Assessment | Environmental Monitoring | Marine Countermeasures |

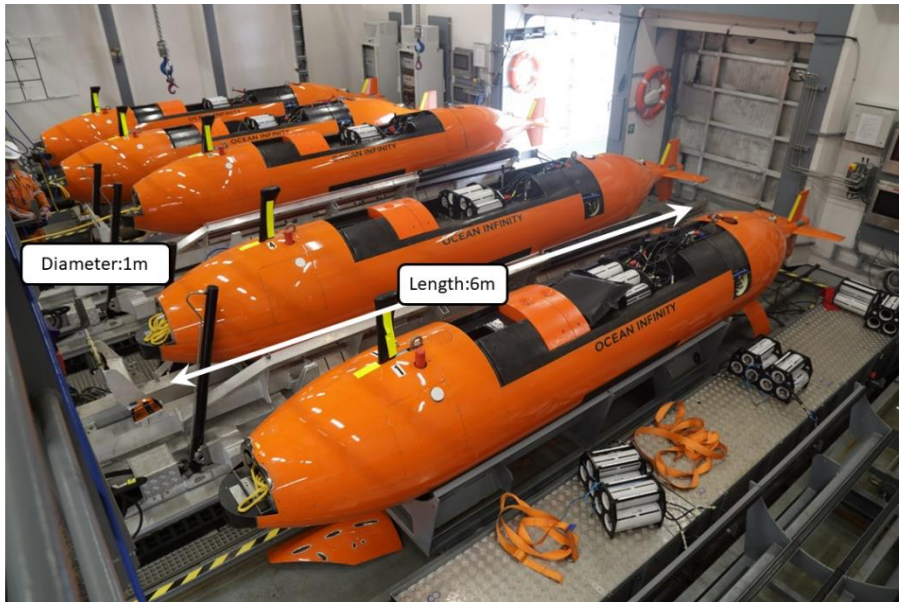

**Figure 11. Hugin AUV being deployed from its launch platform** [107], [109], [110]**.**

I-Tech recently developed the Centurion SP (Figure 12), representing the latest of the "Work Class" of Robotic Operated Vehicles (ROV). This series of ROV was specifically designed for service operations in deep-water projects with strong sea currents [111]–[113]. ROVs have typically been previously used solely within the Oil & Gas (O&G) sector. However, the latest variants of ROV are being deployed in the planning and construction phases of offshore wind farms, due to the key advantages detailed in Table 3. Limitations of these types of vessels include the requirement of a tethering cable to provide powering and control, in addition to severe restrictions in confined space mission capability.

**Table 3 List of iTech[7] Centurion SP capabilities and properties** [111]–[113].

| Application | Planning | Inspection | Repair and Maintenance | Asset Integrity | Production Enhancement | Decommissioning |
|---|---|---|---|---|---|---|
| Key Features | Payload 300kg | 4000m depths | 230 horsepower | Weighing 3,200 kg in air | Four vectored horizontal and four Curvetech thrusters | 8 cameras available HD as standard with optional 3D video |
| Tasks | Processing and reporting | Visual inspection and digital imaging | Cathodic protection | System repair solutions | Heavy Weather ROV Deployment Systems | Debris clearing and recovery support |

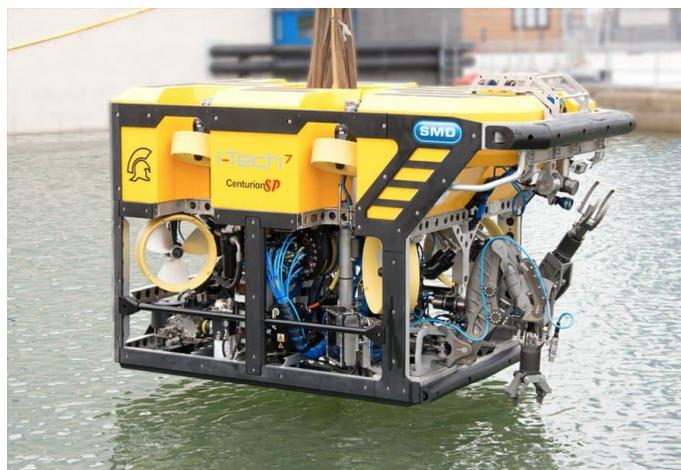

**Figure 12 iTech[7] Centurion SP ROV being deployed for a mission** [111]–[113]**.**

The feasibility of using robots for the internal inspection of nacelle structures in offshore wind turbines was studied by Netland *et al.*, where the robotic platform travels on a permanently installed rail system within the wind turbine nacelle structure as detailed in Figure 13 [95], [114]–[116]. The use of a rail system was demonstrated using a prototype system with a USB camera on a pan-tilt mechanism for a replication of an offshore wind nacelle and included the following advantages:
- A simple and reliable method of moving an inspection robot across key locations.
- An effective means of supplying power to the robot.
- A means of safe travel on a predetermined route, preventing departure and damage to any equipment.
- The inclusion of suites of multiple sensors, such as infrared for detection of friction or faulty electrical systems, microphones for monitoring machinery and other compact sensors for vibration and temperature.

Limitations of railed robots include:
- Requires early adoption/retrofit, to integrate into designs by the manufacturer.
- Due to a lack of freedom in maneuverability, faults detected in unforeseen areas are difficult to properly detect.
- Only cover certain elements and aren't suitable in several areas.

Developments in risk analysis may represent an ideal solution via the validation of the need for resident, railed inspection systems, leading to the implementation of these robotic systems in the early stages of nacelle design. However, continual development may lead to cost reductions in retrospective installation, leading to easier adoption of the technology by industry and resulting in earlier diagnosis of fault detection and maintenance scheduling. Accessibility challenges also exist for all types of platforms where these could simply be addressed in the design phase of many environments and include using ramps instead of stairs, replacing standard doors with automated sliding doors and accessible docking stations, highlighting further development.

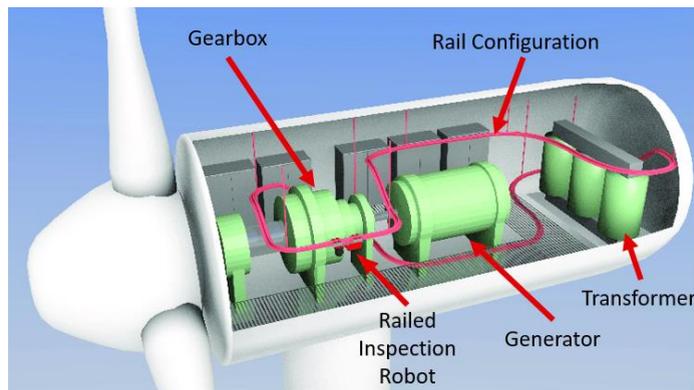

**Figure 13** A diagram of a rail-guided inspection robot within a wind turbine nacelle [95].

The Rolls-Royce, Advanced Autonomous Waterborne Applications (AAWA) initiative (Figure 14), represents one of the most significant investments in autonomous shipping. An industrial and academic consortium of partners have recognized that the ability of a ship to monitor its own health, identify and communicate with its surroundings and make decisions based on that information, is vital to the development of autonomous operations. The sensing technology required to realize autonomous shipping already exists, the challenge is to develop the optimal methodology to combine these technologies safely, reliably and cost-effectively [117].

The primary technical requirements of autonomous maritime logistics relate to:
1. **Sensor Fusion** - Sensing technologies including radar, high-definition visual cameras, thermal imaging and LIDAR, are proven technologies. The challenge requires reliable and efficient integration of these technologies within the context of the maritime environment.
2. **Control Algorithms** - Control algorithms lack maturity with respect to field deployments of robots. Current levels of autonomy are constrained to strictly "within envelope" exploration, however, autonomous vessels will require navigation and collision avoidance to perform real-time decision making based upon sensor data. These algorithms must accommodate the ability to interpret the maritime rules and regulations of the sea. The development of such control algorithms will be refined through an iterative and gradual process of laboratory testing, simulation and incremental field deployment and validation.

3. **Communication and Connectivity** – This will require regulated standards on how autonomous ships will coordinate docking and connectivity to a digital twin of the asset, which supports the operation and health status reporting of the asset.

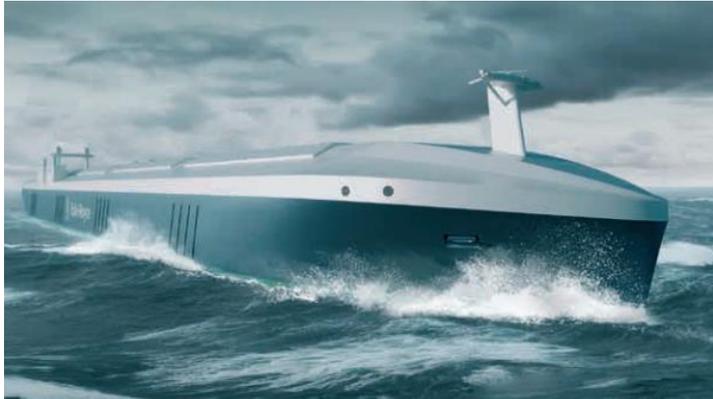

**Figure 14** Rolls-Royce, Advanced Autonomous Waterborne Applications initiative (AAWA) [117].

For smaller vessels, Thales recently reached a milestone using a Unmanned Surface Vessel (USV) named 'Halcyon' (Figure 15- Left) where the USV completed its first autonomous 150 nautical mile trip (over the horizon) across the English Channel to Plymouth. The USV is capable of deploying a payload of a towed synthetic aperture module, which is equipped as a mine countermeasure sonar for high resolution imagery and in water object detection [118]. The crossing demonstrated the robustness of USVs due to the long range ability of the mine countermeasures-class workboat [119].

The Windfarm Autonomous Ship Project (WASP) represents an integrated autonomous vessel and robotic cargo transfer mechanism for the delivery of spares to offshore wind farms, with the potential to transform offshore wind sector logistics. WASP identifies that vessels and logistics can account for as much as 60% of offshore wind farm operating costs, resulting in 25% of the total lifecycle costs [120]. RAI represents an option to reduce these costs, where Autonomous Surface Vessels (ASVs) with the capability for close proximity operation are well suited to serving offshore turbines. An example of this is the Global C-Workers (C-Worker 7, Figure 15- Right), with an operational endurance of up to 25 days without the need for refueling; resulting in lower costs and less fuel required than traditional vessels [121]. Future use of automated unmanned service vessels may provide logistical, surveying, monitoring support, in addition to aerial drone deployment [122]. Within the WASP feasibility study, the challenges for ASVs operating within wind farm arrays were identified, allowing for the characterization of potential ASV performance and the development of a sector route map for ASV integration into manned vessel operations [123].

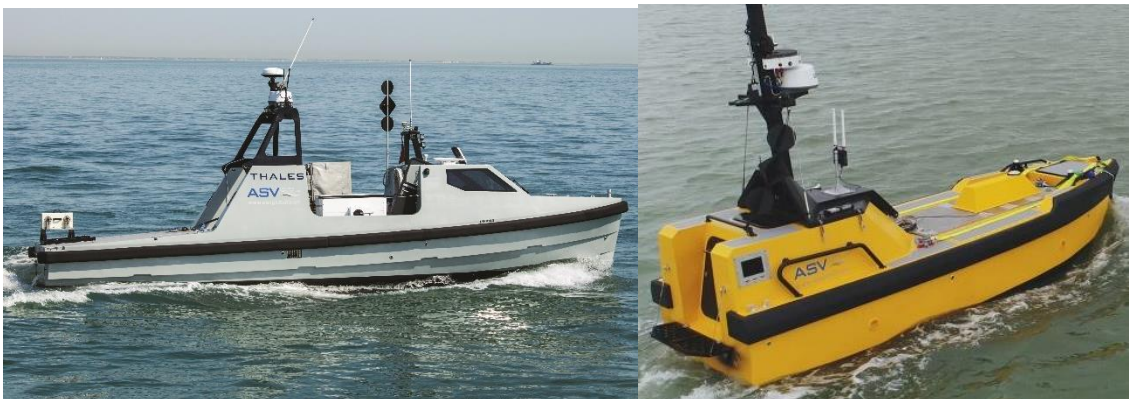

**Figure 15** Left- Thales Halcyon USV [124]. Right- Global's C-Worker 7 autonomous surface vessel [120].

A list of companies who have designed and manufactured their own robotic platforms is displayed within Table 4. The most prominently accepted within industry include AUVs and UAVs. Figure 16 displays images of these robot types alongside the challenges related to these platform types.

**Table 4** List of companies and their robotic platform listed in terms of function and deployment level.

| Robot Type | Figure 16 | Function | Company Name | Mission Type | Autonomy Level (Refer to Figure 10) |
|---|---|---|---|---|---|
| UAV | A | Inspection | Cyberhawk | Visual Line of Sight and some cases of Beyond Visual Line Of Sight (BVLOS) | 3 - Conditional Autonomy |
| | \multicolumn{5}{l}{40,000+ flights for over 300 clients. Qualified drone pilots and an industry experienced inspection engineer alongside Intel Falcon 8. Weather dependent up to 6 turbines can be inspected. Detect defects of ± 5mm. iHawk viewer utilized for post processing and localization of defects. 36.8% savings vs annual rope-access inspection costs. Recently secured a five-year multimillion-dollar contract with Royal Dutch Shell PLC to utilize [125]–[127].} |
| Crawler | B, C | Inspection & Maintenance | Bladebug | Currently VLOS | 2 - Partial Automation |
| | \multicolumn{5}{l}{A novel electronic skin, Wootzkin, allows the robots existing vacuum system to attach to the blade more accurately by using machine learning algorithms. Reductions of 30% in blade maintenance activities. Reduce the need for rope-access technicians. Recent deployment to maneuver for 50m on a 7MW Levenmouth turbine blade [128]–[132].} |
| AUV | D | Inspection, Maintenance & Repair | Eelume | BVLOS | 4 - High Automation |
| | \multicolumn{5}{l}{Self-propelled and can access confined spaces previously inaccessible to AUVs and ROVs. Subsea docking system enables for the system to be deployed resident to the ocean alongside housing of modular payloads. Reduces transport from the shoreline due to remote programming. 'Snake like' design allows for accurate hovering and maneuvering in ocean currents. Payloads- gripper, cleaning tools, sonar. Joints enable for U-shape perspective of a [133]–[135]} |
| Quadruped | E | Inspection | Anybotics | BVLOS | 4 - High Automation |
| | \multicolumn{5}{l}{Autonomous inspection deployment on a converter platform near the North Sea wind farms in a partnership with Tennet in 2018. The mission transmit run-time data collected by visual, thermal, microphones and gas detection sensors [136]} |
| Quadruped | F | Inspection | Boston Dynamics | BVLOS | 4 - High Automation |
| | \multicolumn{5}{l}{Spot deployed by British Petroleum around 305km offshore from the Gulf of Mexico on an O&G offshore rig. Successful navigation of open mesh grated flooring and mesh stairs. Deployed to scan for abnormalities, track corrosion, check gauges, map the facility and detect possible leaks of methane by using onboard sensors that allow for real-time understanding of the plant [137]–[140].} |
| Digital Twin | N/A | Operation & Condition Monitoring | Osberg H | BVLOS | 4 – High Automation |
| | \multicolumn{5}{l}{The world's first fully automated O&G platform, operated by Equinor. Installed in October 2018, the development is expected to yield 110 million barrels of oil via 11 wells. Only requires two maintenance visits a year. Construction was delivered ahead of schedule and at $750 million; 20% lower than the cost estimate of the plan for development and operation. Fully automatic, remotely operated plan utilizes digitalization via teleoperation, interconnected systems and a digital twin for operation of the platform [133].} |

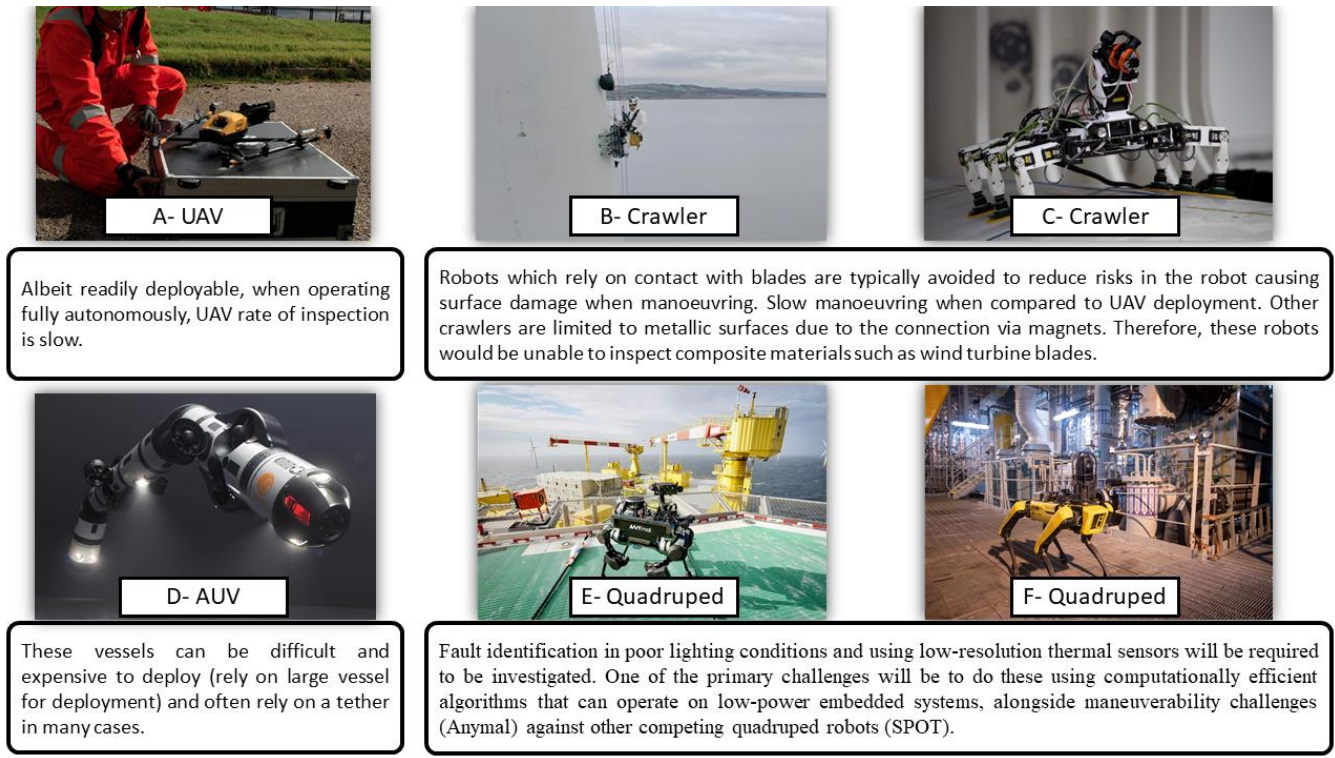

**Figure 16** Compilation of robotic platform deployments as discussed in Table 4 alongside challenges faced [129], [134], [136], [138], [141].

Quadruped robots (Figure 16E and F), have been the focus of significant research activity in 2020/21 due to demonstrably increased levels of robustness, agility and maneuverability. These attributes had previously represented significant challenges, where a large amount of public outreach has been achieved by manufacturers Anymal and Boston Dynamics. Further developments for quadrupeds, which address challenges in perception and autonomy, are expected to result in full deployment in industrial assets, resulting in increased safety during continuous deployment.

Digital twins are being redefined as digital replications of living as well as nonliving entities that enable data to be seamlessly transmitted between the physical and virtual [142], [143]. The continuing development of digital twins must also ensure that a human operator can trust that an asset is operating as intended. Within the offshore wind sector, wind farm arrays are being commissioned further offshore, and will require more BVLOS inspection and operational data analysis undertaken from remote facilities onshore. However, as offshore assets warnings are collected and transmitted for analysis there is the possibility that these warnings are underrepresented or overrepresented (where a fault warning does not accurately represent the level of severity). This could lead to a remote human operator trusting erroneous data, via the digital twin (simulation), over the real-world asset; leading to further damage or inefficiencies due to the discrepancy.

State of the art skills development training includes an immersive hybrid reality offshore wind training facility uses virtual reality and has recently been commissioned at Newcastle College Energy Academy in Wallsend to ensure future engineers gain the relevant experience. The system recreates the working environment for offshore wind turbine engineers and virtually subjects users to a realistic, yet safe environment to locate and diagnose simulated faults [144]. This allows a safe training environment to develop the vital skills required to work in the offshore wind industry environment.

Other, more general, robotic platform concerns and barriers include inspection robots required to operate near high voltage sources and must overcome electromagnetic interference of the circuitry, ensuring reliable communications and potential arcing from the asset to the robotic platform. Electromagnetically harsh environments require accurate localization in dark, GPS denied environments, and clear fault detection and verification, to support existing condition monitoring systems [145], [146]. However, a gap exists in ensuring that RAI can be deployed offshore regularly. This is due to there being no regulation which is uniform and adaptive that can keep up with the latest technology advancements, no standardization of designed infrastructure and equipment and lastly, a workforce that has been trained to utilize the capabilities of robots for seamless deployment. On an industrial level, companies have justified concerns over equipment malfunctions and cyber-attacks, however industry requires a mindset shift. Where a key enabler includes the workforce trusting autonomy and the reliability of RAI [147]. This is where the design of robotics becomes important as autonomous systems must be tailored to the mission and environment.

*4.4    Hype Cycle and Overview of Robotics and Artificial Intelligence*

Hype cycles provide an opportunity to evaluate the relative market promotion and perceived value of innovations, resulting in a useful comparison between consumer reality and consumer hype. The expectation of the performance of a robotic platform tends to be high upon the innovation trigger. In reality though, the actual performance does not often match the capability of the technology. As the technology develops from this point, the hype curve reflects what the potential of the technology is against what the reality of the capabilities of the technology are (representing technology readiness levels). Figure 17 demonstrates that the visibility of innovation can be represented against time, where the anticipated potential of results is reflected throughout the chart and is partitioned into the following segments, where a RAI platform exemplar is provided to aid understanding [148]:

- **Innovation Trigger**- Railed robots represent a platform type which has recently undergone the innovation trigger with a commensurately sharp rise in the number of applications. However, the offshore wind industry will be required to incorporate railed robots in the early design stages of wind turbine nacelles, a process difficult to retrospectively achieve in the lifecycle of a platform.
- **Peak of Inflated Expectations**- Climbing robots are designed to ascend metal and composite structures. However, many of these robots move very slowly and are also unable to complete maintenance with the level of versatility, dexterity and accuracy of a rope-access engineer. Climbing RAI also tend to be severely limited in terms of payload.
- **Trough of Disillusionment**- ASVs are reaching the trough of disillusionment, as they have reached the limit of their effectiveness in terms of current capabilities. ASVs will play a vital role in the logistics of other automated platforms, operating as a deployment platform for UAVs and crawler robots. However, until advancements in the levels of autonomy of these platforms increase, ASV deployment will remain restricted. AGVs have also reached this point of the hype curve. There is an increased availability of inexpensive payloads and improved manipulation to complete more tasks during O&M however, in many cases still require teleoperation to ensure trusted autonomy.
- **Slope of Enlightenment**- Quadruped robots have recently advanced to the slope of enlightenment. This is due to the solution of significant robustness barriers due to balance problems and the challenges posed via operation in different terrains and complex environments.
- **Plateau of Productivity**- UAVs have reached this segment of the hype curve as they are the most readily deployable robotic platform to date. However, there are still advances to be made in terms of the level of autonomy within these platforms, which predominantly operated via remote human control mode and within line of sight of a pilot, rather than via fully autonomous means.

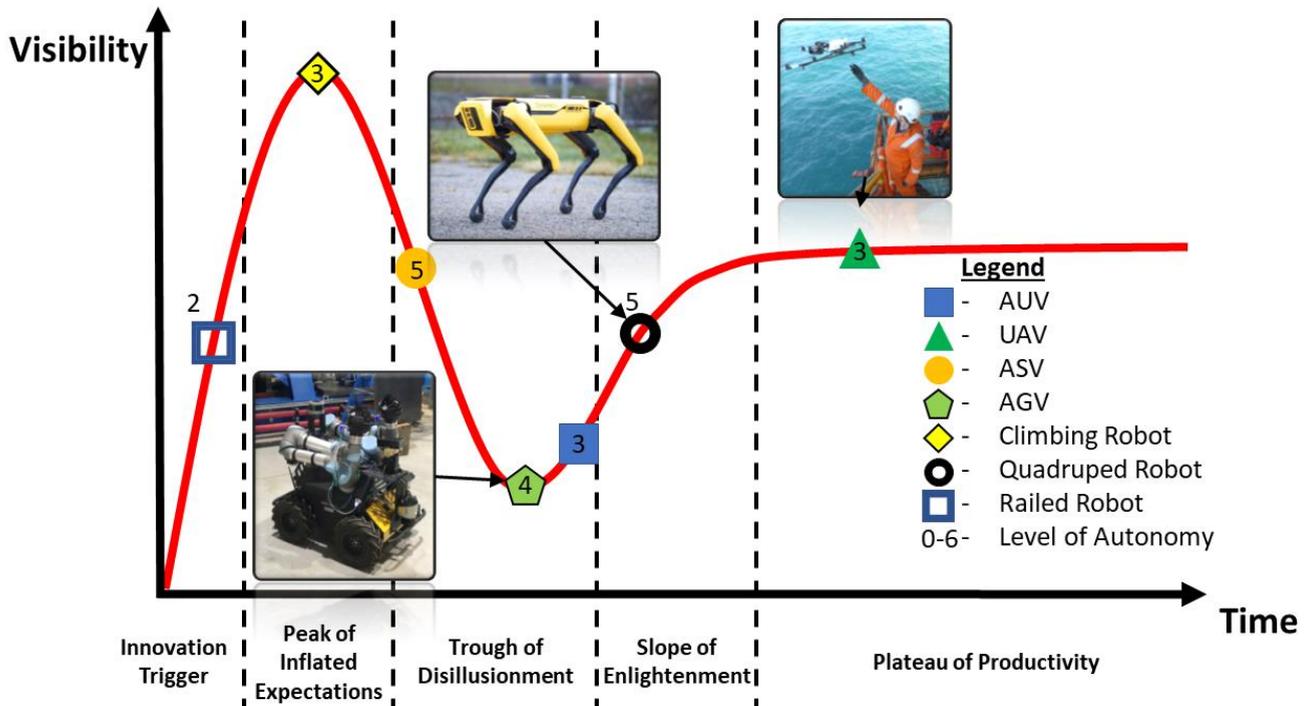

Figure 17 **Robotic platforms positioned relative to a hype curve highlighting their level of visibility and autonomy with respect to time of innovation.**

UAVs are currently at a developmental barrier, as full value of autonomy occurs when a robotic platform can achieve true BVLOS operation. To achieve this goal, the system must possess the ability to adapt and respond to unforeseen challenges in real time, while also improving the energy density to lift useful payloads, in tandem with improved flight durations. These capabilities will require significant advancements in reliability, resilience and endurance. Another key barrier in UAV inspection includes limits in effective sensory payloads, which are limited to surface measurements via visual or thermal imagery. Other sensors capable of subsurface structural evaluation require contact with wind turbine blades. These are sensor types unsuitable as they are susceptible to noise and require operations within a highly risk portion of the flight envelope [149]–[151].

A major limitation is currently impeding RAI deployment, due to robots not being integrated with the surrounding offshore architecture. There currently exists an imbalance between delivery of service and autonomy, which is primarily due to a product centric mindset where system of systems architecture design has not been tailored towards a design for autonomy as a service.

## 5 COMMERCIAL AND PATENT DATABASE SURVEY

The Espacenet and Scopus databases utilized in this study include over 120 million published patent applications and registered patents, both worldwide and in the UK [152]–[154]. This study investigates how many patents are relevant to the ORE sector and the future emergence of energy and RAI in this sector. Growth and development in RAI for the ORE sector will result in improvements which are transferrable to a range of sectors. By focusing on the ORE sector, we target knowledge and research gaps. The methodology for comparing the search is identified within Figure 18.

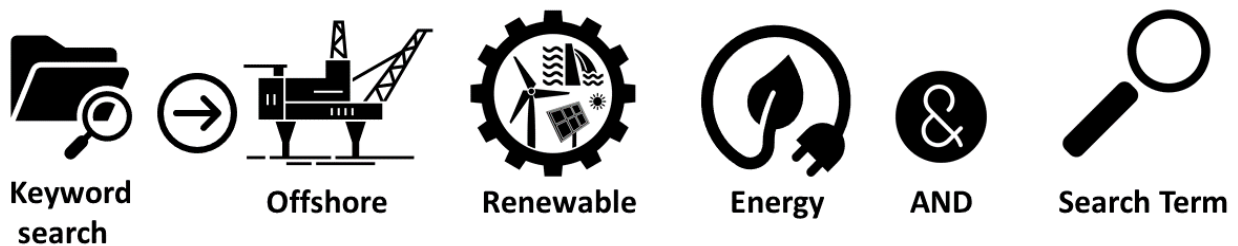

Figure 18 **Methodology of how the keyword search was completed on Espacenet and Scopus.**

The critical path of RAI has resulted in a clear growth in the number of approved patents worldwide identifiable in Figure 19, with an increase from 2018 onwards. The robotic technologies most widely adopted, and with the most patents approved worldwide include railed robots, AGVs and crawling platforms. This has been due to the technological advancements in hardware resulting in cheaper device production, wider accessibility and with improvements in payloads that are applicable to several functions. When conducting the search, a high number of patents for UAVs was expected. However, our results show that UAV developments have reached saturation, due to a plateau of productivity via the utilization of UAVs in other similar sectors and design maturity. Of the platforms identified in this search, none claimed to be compliant to the EU ATEX directive. All future field systems in robotics will need to be compliant with standards required for that specific sector E.g. O&G systems often have to be ATEX compliant. In the instance of ORE, it will be dependent on the deployment, for example radio frequency signals within a substation will need to be given more consideration to ensure reliable communications. In addition, companies which could utilize RAI are now looking to identify OPEX costs associated with each platform; reducing the risk of investment in a robotic platform as companies want to ensure their investment increases productivity and efficiency.

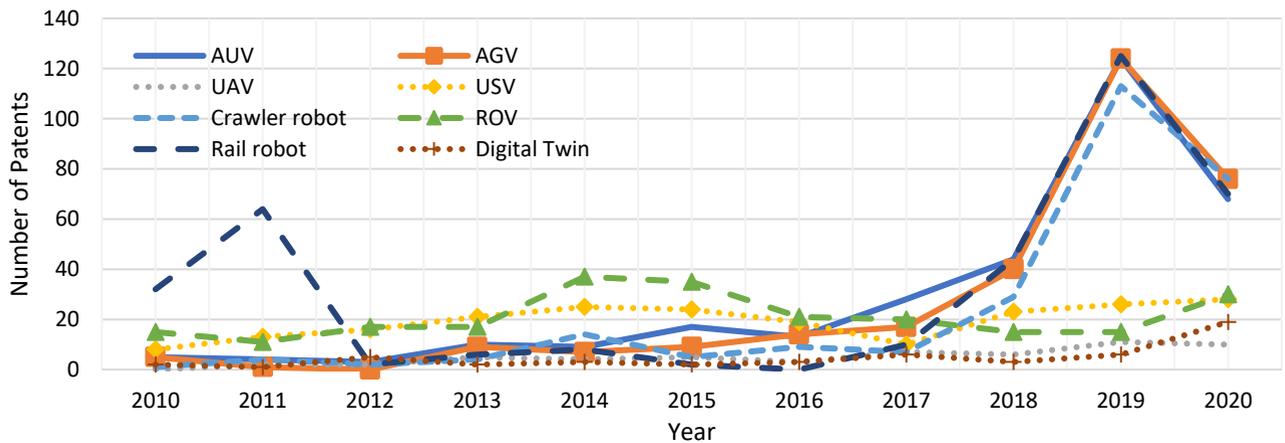

**Figure 19** Number of patents approved per year for each robotic platform type [152]

To identify the purpose and functionality of robotic platforms, key word searches were made for different mission deployment scenarios. Figure 20 indicates a marked increase in the number of patents in inspection, repair, navigation and mapping from 2017, with a clear peak in 2019. This increase is due to industry identification of the challenges and engineer innovation to develop solutions. In addition to the UK outlining its future approach, via the UK Digital Strategy 2017, which began to outline the UK method to aid the development of robotic technologies capable of autonomous operation within hazardous environments, such as radioactive areas within the nuclear sector [155]. Engineers recognize that RAI has the potential to extend to more complex tasks, specifically operations and maintenance. A noticeable dip in 2020 is attributable to the COVID-19 pandemic, with and the requirement for workforce isolation and remote working. This trend is observable within all the figures presented in this section. A corresponding increase in patents is expected in 2021 as the backlog of applications is cleared. The small increase in logistics is likely to be due to the requirement for human supervision of autonomous systems when deployed on-site. Various levels of autonomy in robotic platforms have been developed by academia and industry, which include remotely operated and semi to persistent autonomy. However, it is often the case that these technology prototypes are available to third parties, but a lack in regulation of the technology impedes deployment. As sensors and other onboard equipment reduce in size, along with enhanced functional integration with AI, it is anticipated that an uptake of autonomous robotic platforms will increase in the ORE sector.

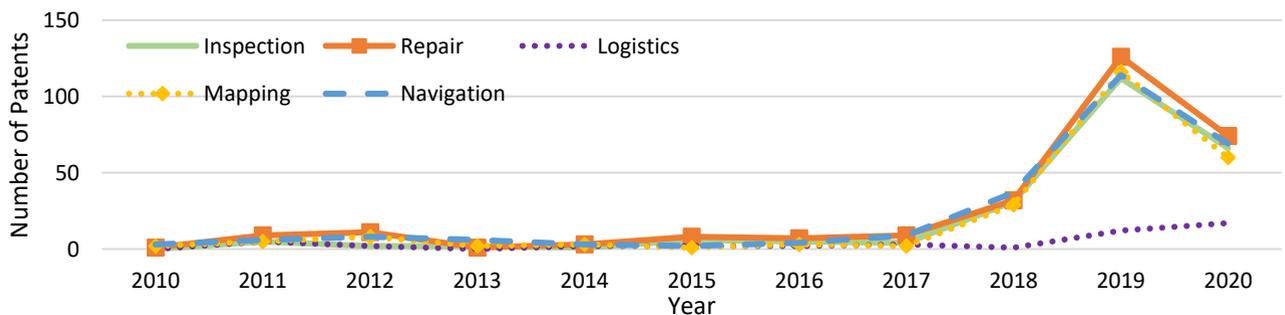

**Figure 20** Number of patents alongside the primary functions of robotic platforms [152].

AI holds the potential to advance automation for not only service robots, but also in how data is collected, secured and used to increase trust, safety and efficiency. Using a key word search of 'Offshore Renewable Energy AND AI', the number of patents where AI played a significant role in the ORE sector, pertained to intelligent and resilient systems. Figure 21 highlights where AI is used in the ORE role and identifies that an increase in numbers of AI related patents commences in 2018. The AI sector deal outlined by the UK government resulted in an increase in the number of AI related patents and with a corresponding increase signifying industry acceptance of the benefits of implementing AI [36].

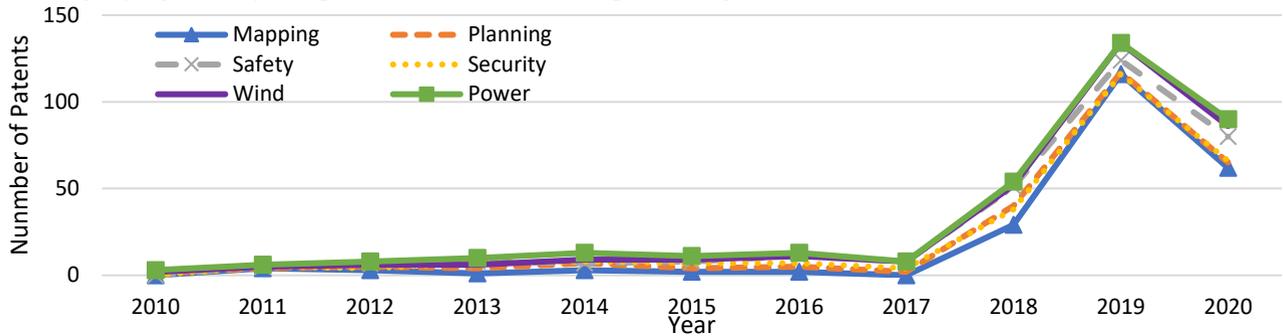

Figure 21 **Number of patents with an artificial intelligence has focused** [152]

Addressing the ground up capability challenges has been a driving force for industry and academia to date. The number of patents reveal key areas which, until developed, will inhibit the safe deployment of robotics. AI, ML and UAVs are well documented within other land-based sectors, however, Figure 22 shows that the number of patents approved in the ORE sector identify safety, AI, ML, decision making and AUVs with the most approved patents, highlighting the current ORE research focus. In the future development of RAI in the ORE sector, we identify several challenges from our search in the patent database. To ensure seamless integration of RAI within infrastructure and environments, we identify several overlooked areas that will inhibit the future deployment of intelligent systems. These are viewed in Figure 22 and include UAVs, self-certification, resilience, ontology, trust and digital twins alongside their definitions in Figure 23.

Current standard operating practice for autonomous systems requires deployment by human operators on-site. As a result, minimal focus is placed on systems possessing the ability to self-certify their onboard systems; a task typically carried out by the deploying human operator under Visual Line of Sight (VLOS) flight rules [156], [157]. However, there are variables that could negatively impact the success of a mission, such as the battery life, operationally induced and physical properties of the robot, software consuming resources or weather conditions. The deployment of fully autonomous systems requires an additional variable – trust. Under a BVLOS deployment, trust becomes an essential element of the relationship between the autonomous platform, human and infrastructure. Consequently, we identify trusted autonomy becoming a mandatory requirement as missions are deployed further offshore. Trust and self-certification will be crucial for autonomous systems which are resident to remote operating areas, such as offshore platforms.

Digital twins are gaining a significant amount of interest from academia and industry [158]–[160]. These systems are highly desirable for small to large companies and can be integrated more easily with other systems. A real-time digital twin of robotic platforms allows remote operators to supervise and have an overview of a robotic platform/platforms in real-time. This enhances the trust between robotic platform and human observer as information about the robotic platform, environment and infrastructure can be displayed, enhancing the BVLOS information which could include self-certification, mission status and data collected from inspection payloads. Integration of operating data with a digital twin becomes increasingly important as robots are required to enter zones which are inaccessible to humans or areas where limited access is available, such as in the ORE sector.

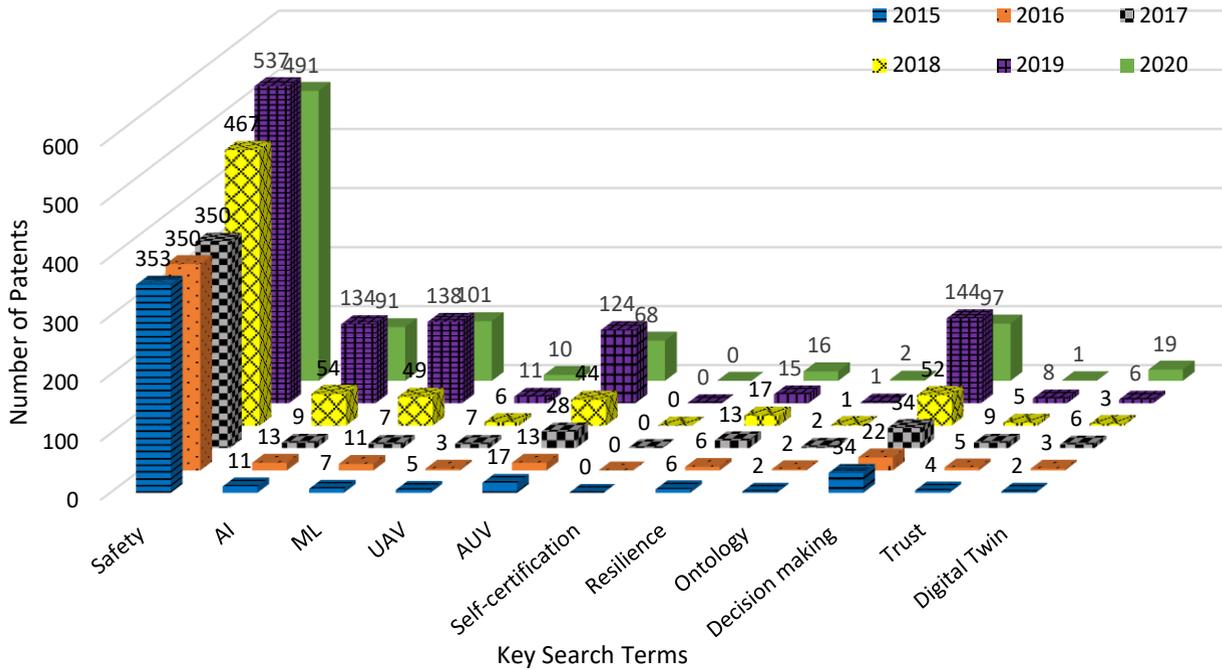

**Figure 22** Key search terms highlighting the number of patents in the advancement of robotics in the ORE sector [152].

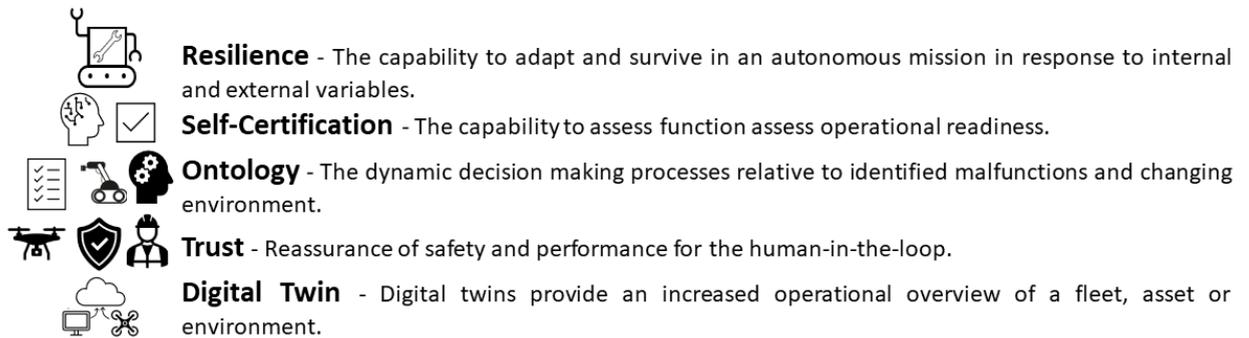

**Figure 23** Key search terms which have been overlooked in academia and industry alongside their definitions.

By focusing on market summaries and identified trends during this period we can determine why RAI has seen an increase in the ORE sector. Firstly, we recognize the importance of the ORE sector as a whole, however, the increase in ORE technologies has mainly been attributed to offshore wind. In 2018, 90% of global offshore wind capacity was installed in the North Sea and Atlantic Ocean however, capacity was also growing worldwide in areas such as China. Turbine technology, wind farm development and O&M were identified as areas which could drive down the cost of electricity from offshore wind. New technology developments, which include floating foundations, would increase the economic potential of offshore wind technology, allowing the development of larger areas in deeper water, which are currently unfeasible for fixed-bottom structures. The emergence of other forms of ocean energies which include tidal barrage, tidal current, wave energy and thermal gradient will further ensure the resilience of ORE and encourage the development of new robotic platforms to ensure these systems are maintained [161]. From this analysis we identify key points, trends and challenges which include:

- 2018 was a key turning point, where many applications for patents in RAI were completed.
- Several critical areas require development, including self-certification, resilience, ontologies, trust and digital twins.
- Logistics has several patents, however, requires other technologies in RAI to advance so that logistics platforms can be used to rapidly deploy autonomous platforms BVLOS.

The number of RAI related patents in the ORE sector has increased due to solutions created through robotic implementation. As wind turbines move further offshore, BVLOS missions will be planned onshore and completed autonomously by resident robots at the asset field. Larger wind turbine designs capture higher wind speeds when positioned further offshore, representing significant risks to offshore workers and limited time periods for O&M procedures and results in increased delays and costs for wind farm operators. In contrast, the deployment of autonomous robotic platforms in the offshore environment can reduce the need for human presence, reducing the risk to humans whilst maintaining asset productivity and providing increased resilience to weather resulting in increased windows of operations. Lastly, robots enable productivity to increase during offshore inspection due to the rapid deployment and increasing number of payloads which can be used to identify faults and complete maintenance in certain scenarios.

# 6   RESEARCH DATABASE

## 6.1   Strategic Robotic Initiatives for Offshore Energy

Several academic institutions have been awarded funding to support the development of the offshore wind sector. The Holistic Operation and Maintenance for Energy from Offshore wind farms (HOME Offshore) project includes Research and Development (R&D) of advanced sensing techniques, robotics, virtual reality models and AI to reduce O&M costs of offshore wind farms [91], [162].

The foundation of the project includes cutting edge technologies resulting from concurrent research developments and provide a novel approach to O&M. The HOME offshore project mission statement is the integration of the following elements into a single framework will reduce costs whilst also improving previous maintenance cycles. The emergent disruptive technologies include [162]:
- Multi-physics strength-stress models, which are challenging for systems but still feasible.
- Drone technology with improved run-time duration and levels of sophistication.
- High-fidelity data streams from advanced sensing techniques, which are low cost.
- Post processing data analytics of large volumes of data or 'big data'.
- Emergence of improved robotic platforms and capabilities in the form of improved robot manipulation and self-certification.

The Multi-Platform Inspection, Maintenance and Repair in Extreme Environments (MIMREE) project, represents a two-year, £4.2 million, cross-sector innovation project tasked with revolutionizing offshore wind operations via RAI. The project is demonstrated how autonomous motherships and robotic crews are proficient in the planning and execution of O&M operations, such as inspection and repair missions, without the requirement for an offshore human presence. Realizing these objectives would lead to the world's first fully autonomous robotic inspection and repair solution where the ORE Catapult has predicted that the project will save £26 million across the lifetime of an offshore wind farm [163], [164].

MIMREE had defined a clear set of goals to achieve within the two years of research and development. These included [163], [165]:
- Utilize a mothership USV) to support the launch of robotic platform and recovery trials.
- Create a moving blade inspection system which would be mounted on the mothership to capture blur-free images in varying sea states.
- Utilize a UAV and launch pad with the capabilities to transport and deploy blade crawlers to wind turbine blades and for retrieval to the mothership.
- Develop mission planning tools for coordinating the robotic assets, underpinning their autonomous behaviors and MIMREE's planning-monitoring-execution cycle.
- Demonstration of the MIMREE sub-systems (mothership, drone, blade crawler, mission optimization, moving blade imaging capabilities) and verify their performance and reliability.

The Offshore Robotics for the Certification of Assets (ORCA) Hub was founded in 2017 to support the energy transition and growth of renewable energy generation. The Hubs primary aim is to utilize RAI to enhance asset integrity monitoring for the offshore energy sector by using robotic solutions which can operate and interact safely autonomously in complex or cluttered environments. The ORCA Hub was initially funded £93 million and has recently extended its funding by £2.5 million from UK Research & Innovation (UKRI). The funding extension aims to expand its work with industry partners whilst also expanding into new sectors using the lessons learned during the project period [166]–[168]. The ORCA Hub focusses on areas including:

- Intelligent Human-Robot Interaction
- Mapping, Surveying and Inspection
- Robot and Asset Self-Certification
- Planning, Control and Manipulation

The National Robotarium, currently under construction at Heriot-Watt University will be a major contributor to world-leading research for RAI, creating innovative solutions to global challenges by progressing cutting-edge research into technologies to create disruptive innovation in a rapidly increasing global market. This facility is expected to open in 2022, where the facilities will add to Heriot-Watt laboratories in Ocean Systems, Human Robotic Interaction and Assisted Living and Self-Certification and Safety Compliance work packages. The consortium was awarded £3 million dedicated to researching trust in robotics [169], [170].

### 6.2 Academic Database

To identify the focus of these organizations and other academic institutions, an investigation of academic papers published in the ORE sector has been completed. This identifies areas considered as gaps which require further research focus. Using the same methodology as represented within Figure 18, we observe that the number of documents in the ORE sector with the keyword of robot is increasing over time, as displayed in Figure 24. The search terms of the most popular papers in the ORE sector including robotics are presented in Figure 25-Left, which highlights wind turbines, wind power, controllers, wave energy conversion and oil well production as the topics with the most published papers. While offshore oil well production is not a renewable source, there is an inherent overlap in the robotic technologies used for both ORE and in the O&G sector. Figure 25-Right displays the three leading countries of published papers including China, the United States of America and the UK. China has emerged in a key position to maximize exports of offshore wind turbines, with the Chinese government supporting local developers and has built a local value chain, which is expected to grow considerably by 2025. This has included the Chinese government changing the energy unit set price to a guide price (no higher than 0.8 yuan per kilowatt hour), with new tariffs for newly approved projects to be set competitively [171]. These policies have placed Chinese manufacturers, such as MingYang, Envision and Goldwind in strong economic positions. Rystad Energy predict that by 2025, Asian installed offshore wind capacity will increase sixfold to 52GW, where China will provide 94% of Asian capacity [172]. This expected increase is attributable to the recent upscaling of manufacturing of wind turbine construction in China and is expected to level with European offshore installed wind capacity by 2025 [173].

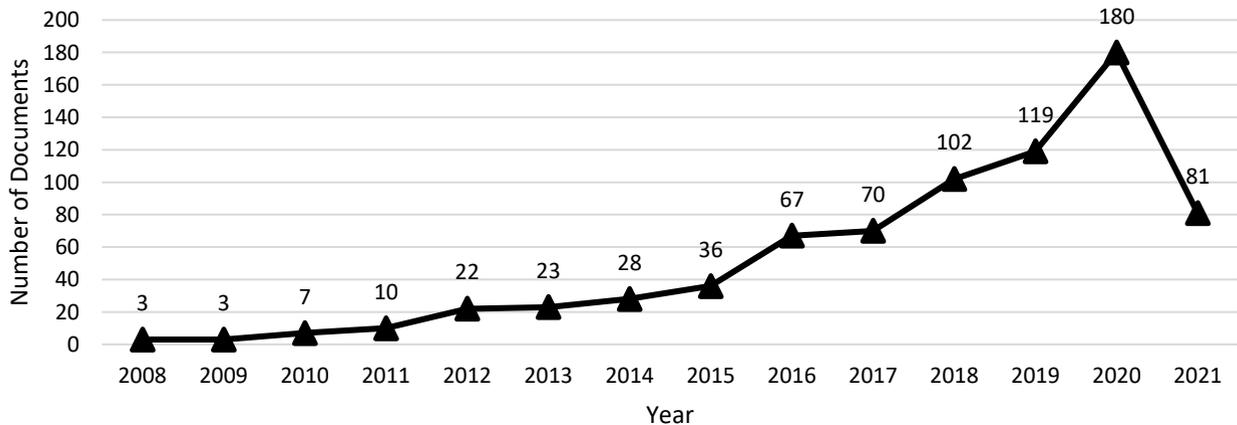

**Figure 24** Number of published documents by year in the ORE sector, keyword 'robot' [153].

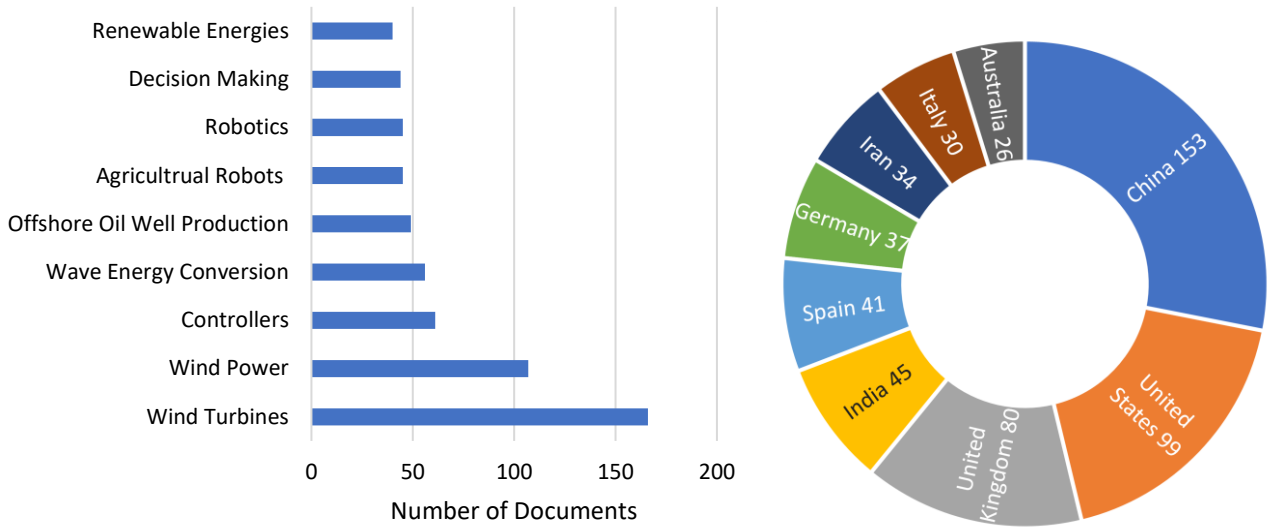

**Figure 25** Left- Key search term 'Robot' in the ORE sector, Right- Documents published by country of origin (2008-2021), key search term 'robot' [153].

A year-on-year increase in the rate of commissioned British onshore and offshore wind capacity can be observed from 2010, as identified in Figure 26. This increase follows the trend shown in Figure 24, where an increase in the number of turbines equates to increased O&M requirements and the resultant need for inspection and maintenance robots [174], [175].

The Scopus database identified 2016 as the threshold, where the number of papers published involving robotic platforms increases significantly (Figure 27). In 2016, onshore and offshore wind power became the second largest power generating capacity in the European Union (EU), with 154GW representing a share of 16.7% of installed capacity [176]. This analysis found that UAVs, ROVs and AUVs were the most published paper topics in the ORE sector from 2016 onwards. This increase in publications can be attributed to the following:
- The identification of the challenges faced in the ORE sector.
- 2016 representing the year prior to UK offshore wind licensing round, resulting in significant increases in projected UK offshore wind capacity, increasing the requirement to find solutions to the challenges faced in industry.
- Growth in government support for RAI as the technology matures, as predicted in the *"Eight Great Technologies"* outlined by the UK government in 2013 and the UK Digital Strategy 2017 [102], [155].
- The continued development of academic solutions to the immediate requirements of the early-stage lifecycle of offshore wind farms, as represented within wind farm planning and environmental surveys.

AUVs and ROVs were the first robotic platforms developed for use in offshore wind farms. This represents the current need to enhance environmental surveys in the wind farm lifecycle and ensure accurate and more efficient wind farm planning. Hence, an increase in the number of documents for the platform types as displayed in Figure 27, as many wind farms are at the early stages of their lifecycle.

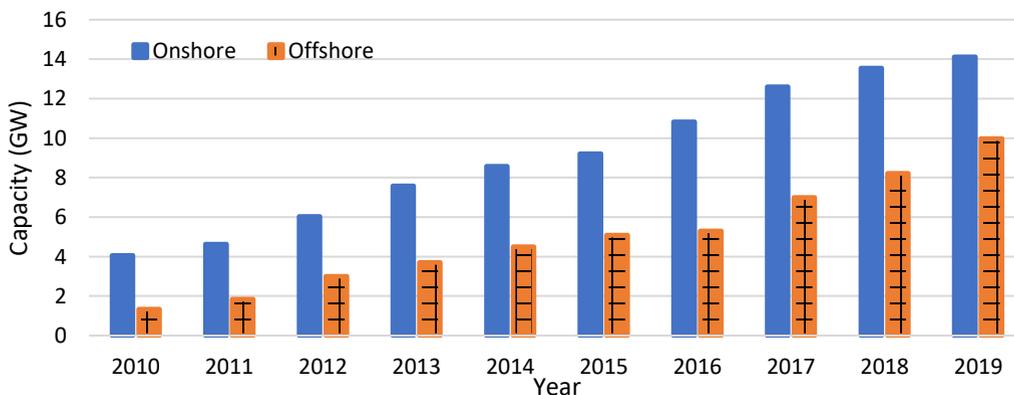

**Figure 26** UK onshore and offshore wind capacity 2010 to 2019 [174], [175]

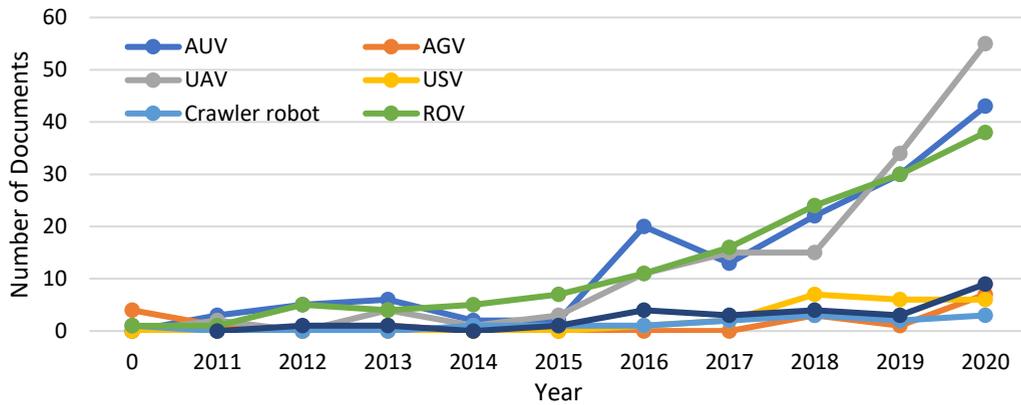

**Figure 27** Types of platforms prioritized in the ORE sector [153].

To identify the types of missions that these robotic platforms were designed to complete, a key word search was performed for mission type, where a steady trend for all types of missions is observed. The number of publications each year has increased, as illustrated in Figure 28, highlighting the capabilities which exist within RAI deployment, allowing for the identification of a timeline for new sensing technology creation. This represents the innovation trigger alongside a rapid saturation of the technology the year after.

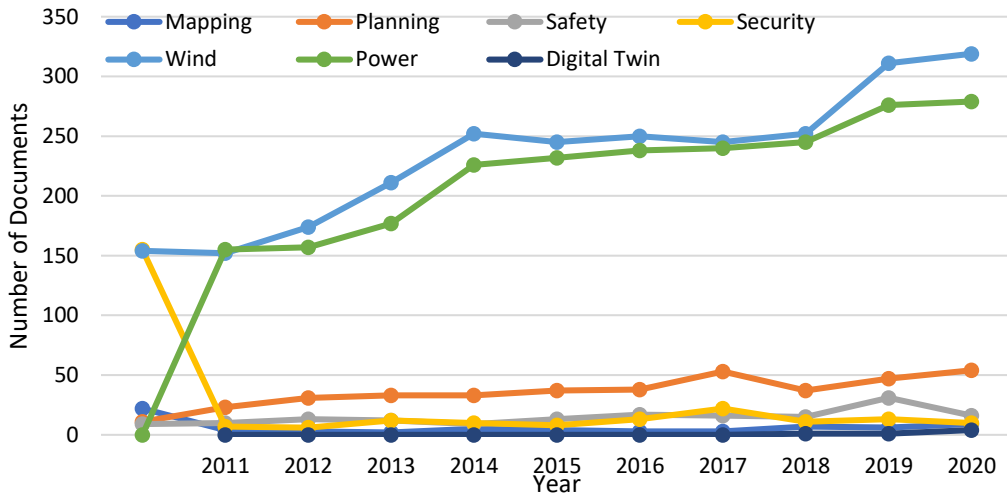

**Figure 28** The functions which robotic platforms are being utilized for in the ORE sector [153].

To identify how AI is being used within the ORE sector, a key word search was performed to ascertain the focus of academic research in the area. As displayed in Figure 29 several AI-themes exist in the public domain for wind and power related applications. This infers an increasing trend and identifies a potential gap in AI use to increase safety and for Digital Twin (DT) capabilities. As wind turbines move further from shore, safety and the application of DTs are critical, as the workforce will be required to monitor the offshore environmental DT from onshore facilities. Further development in this area will be critical in the roadmap to a fully autonomous offshore environment, where a remote operator has an enhanced operational overview of all assets including infrastructure, environment and robotic platforms.

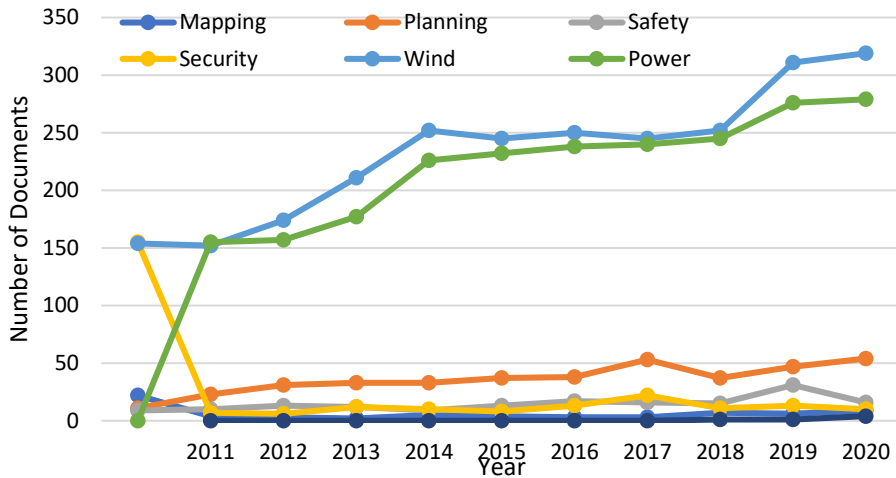

Figure 29 **Number of documents published on AI in the ORE sector** [153].

A key word search covering the period 2017 - 2021 (Figure 30) provides an overview of relevant areas of academic research and identifies the future needs to ensure success in the offshore environment. We identify safety and decision making as crucial areas which have been developed for the ORE sector. Of concern is the number of documents published in areas identified as future challenges, which must be addressed in the offshore sector. These areas include self-certification, resilience, trust, BVLOS and DTs, and ensure that RAI can operate at their full capabilities.

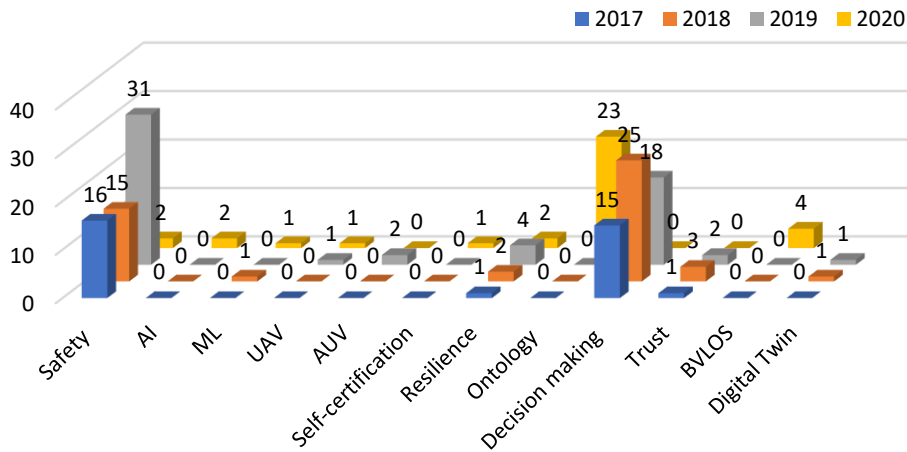

Figure 30 **Key search terms in published documents in the ORE sector** [153].

## 7    CAPABILITY CHALLENGES BASED ON IDUSTRIAL INSIGHTS

To avoid the risk of robotic systems becoming another operational overhead, requiring additional asset management, systems must be resilient and reliable. The industrialization of RAI will require autonomy as a service. A central challenge to achieving this is the generalization of AI and field robotic implementations to use case specific deployments. This review identifies key areas for development to ensure RAI is readily available, adaptable and reliable within offshore workspaces. This includes the self-certification of autonomous systems, DTs for human-robot interaction and asset integrity inspection.

### 7.1    *Self-Certification of Autonomous Systems*

As autonomous robots are deployed further offshore in BVLOS scenarios, operators must ensure that missions are planned effectively to safeguard the success of a mission; wherein a mission must be executed perfectly first time, every time. To ensure full confidence in RAS, operators must have access to reliability information, to assess run-time resilience. Run-time reliability ontologies must be designed for each platform, where diagnostic data is applied to the ontology, which monitors for warnings, faults and malfunctions that risk the operational resilience of the robot. To achieve this, data streams will be required to be combined with computational models to encode safety critical parameters of the robot and asset as they are deployed.

Live updates of robot capabilities and mission objectives can provide answers to the questions posed in Figure 31. This facilitates trust between a robotic platform and operator, leading to enhanced interaction via access to vital asset information in real-time at any point during the mission via the self-certification data.

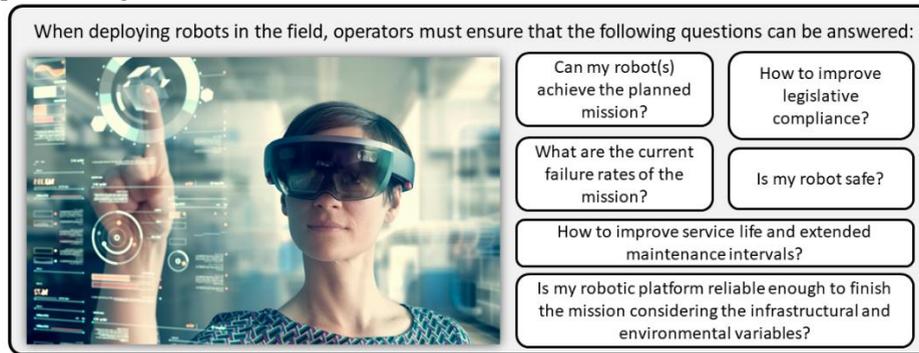

**Figure 31** Mixed reality interaction alongside questions a remote operator requires answers to when deploying BVLOS robots in the field.

By implementing a self-certification framework for robotic platforms, it is possible to ensure the benefits outlined are captured and that risks can be minimized by analyzing real-time and historical data. Diagnosis, prognosis and health management are the most effective techniques to safeguard a high success rate of autonomous missions. This includes converting fault data into actionable information to allow a remote operator to create a maintenance plan for an autonomous system, or even deploy another robot to complete a mission.

Run-time reliability ontologies are useful in this research topic as they allow robots to monitor their own state of health and recommend decisions for the robot based on any warnings or fault thresholds being identified. A primary motivation for the use of ontologies within robotics is that these knowledge-based approaches offer an expandable and adaptable methodology for capturing the semantic features to model robot cognitive capabilities. This offers an agile and rapidly tunable capability to the challenge of capturing dynamic safety compliance and mission operability requirements. Furthermore, when considering a fleet distribution of robotic platforms, or swarms, the ontology provides a cyber-physical interface to cloud or web-based service robots that enable a platform to collect and share knowledge of a mission [177], [178]. This knowledge enabled architecture provides a means of sharing across different robots and subsystems in a machine understandable and consistent presentation (i.e., symbolic presentation).

The development and implementation of the ontology facilitates the autonomous self-certification of a robotic platform and supports adaptive mission planning enabling for front-end resilience, run-time diagnosis, prognosis and decision making. System ontologies should be designed for easy integration across a range of robotic platform types and should collect data from onboard sensors and actuators [179]. This information from a robotic platform can be relayed to a digital twin to update a remote operator on the state of the robotic platform and should be fully synchronized via bidirectional communications, ensuring an accurate representation of the state of health, mission state and operation of an autonomous system.

### 7.2  Digital Twins for Human-Robot Interaction

A DT provides key benefits for wind farm operators by allowing for enhanced levels of trust in the status of an asset or infrastructure. Data sharing between robotic platforms, infrastructure and remote human operators represents a critical challenge in a fully operational symbiotic offshore environment. As discussed in Section 7.1, system self-certification remains a challenge in robotic platforms, however, a challenge also exists in the transfer of this information to a remote human operator operating remotely (BVLOS) from the deployed system/mission. A digital representation of the real-world asset presents a solution to this issue, where collected data is transmitted to a DT to update an operator about the state of the asset. DTs are not only limited to the health management of robotic platforms but also infrastructural health monitoring, via the integration of multi-sensor data. This would include data from sensors integrated on mobile inspection robots and sensors embedded within major structures.

DTs are a tool which can be applied to remote run-time asset interaction [143], [180]. The framework presented by the research group included a digital tool for asset management which addressed common issues of most existing DT models, such as inefficiencies in data synchronization and data representation, trustworthiness issues and limited support for flexible scenarios. By presenting a DT with two-way control and simulation, autonomous assets may be deployed with a higher level of trust.

Similarly, a platform- and application-agnostic approach allows the framework to be scaled across diverse fields. Despite the sophisticated level of this twin, advances in technology and Machine Learning (ML)/AI techniques will lead to significant technological development.

The current DT implementation provides support for remote and run-time decision making; however, AI and ML can be utilized to support improved decision making and autonomous mission planning. To increase the trust in a system, a secondary system can be used to provide suggestions and pre-determined options for decision making based on mission parameters and real-time asset status. This would allow for autonomous decision making onboard a cyber physical system alongside an update to the human-in-the-loop where they have the option to alter the mission if necessary. AI and ML algorithms could further improve the accuracy and timeliness of these predictions if fed with historical data to compare to the run-time data [143], [180].

Furthermore, advances in the field of virtual and mixed reality can enable more in-depth human-asset interaction [181]. A high-fidelity virtual workspace replicating that of the asset, such as a real-time scan of the asset environment, would increase the level of interactivity between the operator and their system. New graphical user interface techniques could provide more intuitive and more relevant information to aid operators in their understanding of a dynamic situation, thereby reducing the risks involved.

## 8 ROADMAP TO RESILIENT INFRASTRUCTURE

Offshore infrastructure is undergoing significant innovation through the rapid development of RAI as discussed in Sections 2-4. Key areas of engagement and progress have been made with a view to improving robotic platforms via development of physical qualities of the asset. These can be categorized as ground-up approaches and include maneuverability, speed, performance and sensing mechanisms. The authors of this paper believe that many of the robotic platforms required for implementation within offshore wind farms are currently available, and therefore a shift in focus is required in the future to allow operators to adopt and integrate autonomous systems in multiple areas symbiotically. The main areas which will be required to achieve this exist within the full lifecycle and operational challenges of offshore wind farms.

### 8.1 Lifecycle of Offshore Wind Farms

Offshore wind farms have matured to the point where it is possible to address problems within all areas of the lifecycle, as shown in Figure 32. This identifies the key stages of an offshore wind farm and illustrates a circular economy for sustainability. Key areas include:
- The raw materials used within the design and manufacture of wind turbine components.
- The logistics required to transport these components and personnel to offshore locations.
- The construction of offshore wind farms, including the installation of subsea cables, foundations, nacelle, tower and blades.
- The O&M requirements that facilitate optimal operating conditions to reduce downtime.
- The decommissioning of wind turbines.
- The identification of components for repurposing, recycling or disposal.

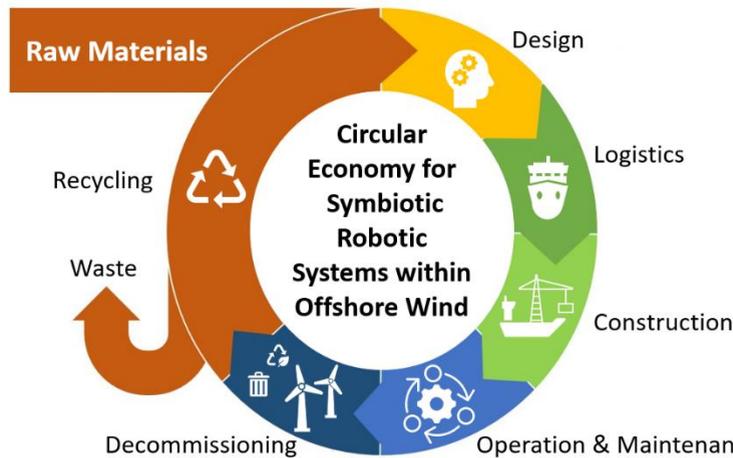

**Figure 32** RAI Circular Economy which can be used to ensure a sustainable and symbiotic offshore wind sector.

### 8.1.1 QUALITY CONTROL OF MANUFACTURING DEFECTS

The manufacturing of wind turbine blades includes the utilization of raw materials, including fiberglass, balsa, polyurethane, composites and resins. To achieve a balance between flexibility and strength, several structural tests are performed before wind turbine blades are certified as safe for deployment. This includes matrix failure, dynamic fatigue tests and optimization tests to ensure high levels of efficiency [182], [183]. Large blades can be manufactured in approximately two days and require ~100 employees. Blade construction is achieved via two main manufacturing processes, utilizing an air tight mold for the transfusion of resin and wet layup [184]. The longest offshore wind turbine blade to date was created by LM Wind Power, at 107m, arriving at the ORE Catapult world-leading blade test facility in 2019. The blade validations included a range of structural tests to demonstrate its readiness for operation at sea and the ability to withstand peak wind conditions [185].

The future of the manufacturing process will utilize increased automation, with improved quality verification via implementation of advanced sensing paired with AI. This will verify the integrity of subsurface features, such as uniform resin distribution and the presence of any air pockets; processes which cannot currently be performed for manufacturing defects. Inspection and maintenance procedures for quality assurance will be adopted between the manufacturing and testing procedure to ensure that blades are designed to a high standard, verifying the structural integrity of blades commissioned from manufacturing centers via a range of emergent NDE sensors.

Rotor blade testing services include determination of damping loss factors, natural frequencies and modal constants in addition to mechanical testing, such as static tests, fatigue tests, post-fatigue tests and collapse tests [186]. Future investigations will be very similar, however, non-destructive sensors will be increasingly used to assess the integrity of the structure prior to failure. There are two proposed solutions to this problem:

1. In the near future, NDE sensors will be directed to the outside of blades to monitor the materials during these tests, which can monitor the return signal at the points where a failure occurred. This would highlight the limits of the blade and allow for measures to be taken to avoid reaching that point of stress.
2. A solution further into the future, once development makes these devices more affordable, includes *in-situ* NDE sensors positioned within the hollow blade structure. This would have benefits for wind turbine blade testing facilities and operators in the long-term as the sensors would be used to assess the integrity of significant areas of weakness.

The NDE sensor signals will have the ability to penetrate through the composite materials wherein the return signal would enable integrity analysis of the internal structure of the blade. This would provide prognostic information about a prospective delamination or defect before a catastrophic structural failure occurs. Value is created for wind farm operators as data from these sensors can be fed into a digital twin, where historical data can be accessed to make a more effective and informed decision about the remaining useful life of a blade via real time structural health monitoring and condition monitoring.

### 8.1.2 LOGISTICS AND CONSTRUCTION

Logistics and construction objectives include the timely installation and commissioning of the wind farm whilst achieving high safety standards for the workforce. Carbon emissions from Crew Transfer Vessels (CTVs) and jack up vessels are an area which has, so far, received little consideration within offshore wind farms. New technology is required to reduce fuel consumption and emissions from these types of platforms, whilst reducing maintenance requirements of CTVs and enhancing operations [187]. The circular economy of offshore wind looks to create a fully sustainable offshore lifecycle, a significant barrier faced by the industry is due to the greenhouse gases produced by these types of vessels during this phase of the lifecycle. If CTVs can address this scenario by using a sustainable fuel source, such as the electricity created from the offshore wind farm itself or from another low carbon fuel source such as hydrogen, this would enable for a more sustainable segment of the lifecycle of an offshore wind farm.

In future, we will see autonomous systems support green field construction of new and/or replacement of wind turbines E.g. maritime logistics to deliver sub-systems and automated pick and place assembly. This will also be represented in subsea construction E.g. submarine cable ploughing [188]. We expect that advances in land based construction from companies such as *Built Robotics* and *Cat* who develop autonomous systems for excavators and bulldozers, will see similar advances in the ORE sector as these advances in systems can be adapted sector-wide [189]–[191]. This will lead to semi-persistent autonomy in heavy jack up vessels, cable laying vessels and CTVs.

### 8.1.3 OPERATION & MAINTENANCE

Operation and Maintenance currently relies on RAI for mostly inspection. Robotic platforms will be better equipped for inspection missions with advanced sensing techniques which increase the effectiveness of asset integrity inspections. This includes inspections of wind turbine towers, blades, foundations and subsea cables. Of significant importance is the development of non-destructive and non-contact sensors which can allow for more prognostic information to be collected, resulting in early surface and subsurface defect detection. In addition, the robots will be better equipped to complete repair missions including cut out and fill for laminate repairs or wet layup [192]. Majority of repairs will include converting the methods used from rope-access technicians to be utilized by robotic platforms, hereby reducing the requirement for workers in hazardous conditions at height. However, due to the nature and variance of damage, we expect that rope-access technicians will still be required for the purposes of medium surface repairs.

We expect to see installation of robotics resident to the field, which will enable readily deployable systems. Drones will be deployed resident to the environment it operates in where a UAV would be responsible for several wind turbines within its area. To ensure the resilience of this system, each UAV will have areas of overlap in case a UAV faces a malfunction and is undeployable. It is also beneficial to ensure this overlap as environmental conditions play a significant role in decreasing flight times. Therefore, conditions such as wind speed should be relayed to a de-centralized digital twin, which can evaluate possible factors leading to mission failure, such as battery life and whether the UAV can operate under strong wind conditions. Underwater vessels can also be integrated resident to the field, with underwater garages capable of the storage and recharging of the vessels. The location of garages would be optimally positioned near to foundations of the substations and could allow for modular designs of tools to be stored and utilized by the underwater vessel. Due to the difficult terrain on offshore platforms and substations, the authors expect that quadruped robots will be utilized for day-to-day inspection tasks on substations. Substations are decreasing significantly in size with Siemens offering an Offshore Transformer Module (OTM) at one-third the size of regular OTMs [193]. This requires smaller robotic platforms with ATEX compliance to reliably navigate confined spaces. We expect railed robots to be utilized more frequently in the future to perform autonomous missions where mission failure due to navigation can detrimentally affect the plant. These systems would typically collect SCADA data of the internal functions either the nacelle or offshore transformer module and can be easily powered via the railed infrastructure.

As O&M is the most developed section of offshore wind in terms of robotics, in the future we expect to see a shift towards symbiotic systems where robots can communicate together to ensure that mission goals are met. For example, in the scenario that a UAV has completed 50% of a section of an integrity inspection, the UAV can share information with the digital twin that it has deployed to base for charging and ask for another nearby UAV to finish the inspection mission. This will ensure the reliability of a robot and resilience of a mission.

*8.1.3.1    New and Resilient Sensing Technologies*

**Wind Turbine Blades**

A tunable Non-Destructive Evaluation (NDE) sensor with an operating range of approximately 2-10 meters from a wind turbine blade would significantly reduce the downtime of an asset within O&M. Frequency Modulated Continuous Wave (FMCW) radar sensing is a novel, non-contact sensing mechanism with the potential to significantly improve the efficiency of O&M procedures on composite structures. The millimeter-wave sensor exhibits minimal limitations, when compared to current employed methods for external and internal diagnosis and prognosis of faults, such as rope access teams or physical contact sensors that often require a couplant. A significant advantage of K-band FMCW radar sensing is the ability to penetrate low dielectric materials and interrogate the return signal amplitude for contrasts in material integrity via analytical techniques. The sensing mechanism is also resilient to the harsh offshore environment, capable of unaffected operation in conditions which can include smoke, steam, mist and rain. The radar sensor is low power, lightweight and emits non-harmful radiation, representing an ideal payload for UAVs and other robotic platforms [149].

The FMCW sensor system has been investigated on a decommissioned wind turbine blade containing a type 4 delamination defect as in Figure 33A & B [149], [194]–[196]. The FMCW radar sensor was able to differentiate between healthy and delaminated sections of the wind turbine blade structure, where the delamination defect was situated on the internal skin structure. The sensor was also capable of detecting contrasts for the same internal delamination defect with two conditions; both dry and with 3 milliliters of water ingress. The sensor was able to detect contrasts from a non-contact position near the external surface of the blade, as pictured in Figure 33B. The acquired data was then transferred to an interactive Asset Integrity Dashboard (AID), as illustrated in Figure 33C, allowing a remote wind farm operator to view their asset in a synthetic environment to attain multiple tiers of fault information.

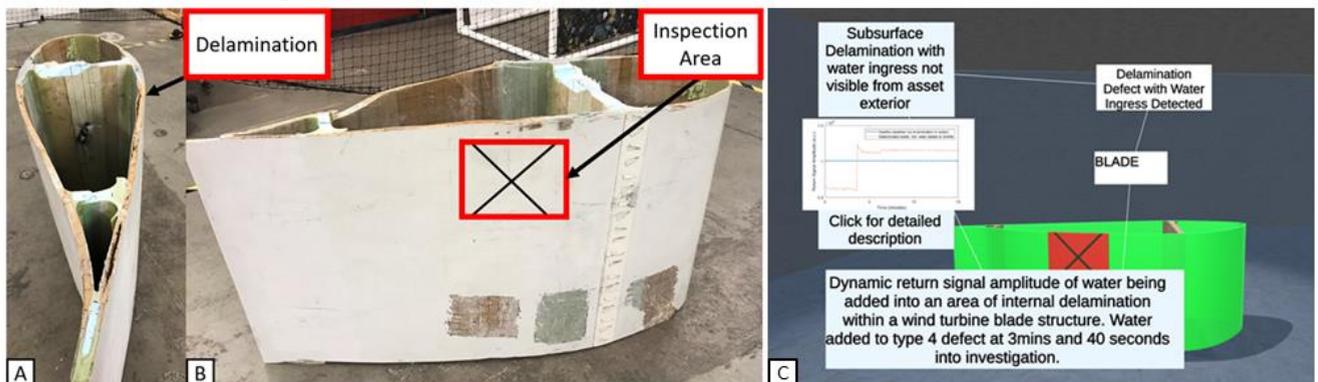

**Figure 33** A- A side view of the decommissioned wind turbine blade displaying the delamination on the interior of the blade. B- Straight on view of the exterior of the decommissioned wind turbine blade highlighting the inspection area. C- AID indicating digital twin of acquired sandwich wind turbine blade in green highlighting a defective area of the blade with red hatching and options for a human operator to attain more information about the diagnosed [194], [196], [197].

**Subsea Cables**

Subsea power cables are critical assets for the security in ORE power transmission. Sustainable power supply and the economic viability of offshore windfarms is highly dependent on the reliability of subsea cables. Consequently, as demand for renewable energy grows, the market for subsea power cables also grows, where subsea cable demand grew at a compound annual growth rate of 7% over 2017-2021, and is estimated to total 24,103km globally [198]. It is estimated that UK power outages caused by faults in subsea cables may stand at £5.4 million per month [199]. In addition, its reported that the costs to locate and replace sections of damaged subsea cables can range from £0.6 million to £1.2 million [200], [201]. Insurance underwriter G-Cube states that offshore wind insurance claims relating to subsea power cables equated to €60 million in 2015 alone [77]. Given the importance of subsea cable assets, it is necessary to develop an integrated solution for degradation and reliability monitoring, and efficient subsea cable maintenance.

Dinmohammadi *et al.* evaluated a Failure Mode Mechanism and Effect Analysis (FMMEA) using 15 years of historical failure data and reports on subsea power cables, with the identification that most of the failures in subsea power cables were attributed to external, environmental factors [81]. In addition, current commercial monitoring systems are unable to account for 70% of cable failure modes, which are not caused by internal failure modes, such as partial discharge, but by environmental conditions and third-party damage. For example, environmental conditions, such as strong tidal velocities, can abrade, stress and fatigue subsea cables through mechanical displacement. This incurs abrasion and corrosion of the cables due to interactions with the seabed topography e.g., rocks [81].

To overcome these limitations, an innovative sensor technology that can provide *in-situ* integrity analyses of subsea cables is discussed. According to Capus *et al.*, bio-inspired wideband sonar can be deployed to efficiently detect and track underwater targets [202]. Furthermore, Dmitrieva *et al.* illustrated that employing bio-sonar technology in combination with time-frequency domain representation, and a convolution neural network system, allows for accurate classification of echo response data of underwater objects [203]. These findings led researchers to explore the capability of low frequency wideband sonar to undertake detailed assessment of subsea cables and how to use state-of-the-art machine learning techniques to evaluate cable integrity data, for example using sonar echo data. The research showed the capabilities for integrity analysis of applying low frequency sonar on subsea power cables, also demonstrating the feasibility of supervised learning approaches on cable classification with different inner structure and failure types [204], [205]. This novel approach enabled the successful distinction of different cable samples to achieve an overall 95%+ accuracy rate to detect different cable diameters and degradation stages. These findings support the suitability of low frequency wideband sonar for the analysis of the complex internal multi-layers of subsea power cables, which is to use wideband sonar to assess the cable integrity and therefore locate potential failure sites.

### 8.1.4 DECOMMISSIONING

The offshore wind sector, though still emergent, can no longer be considered a frontier energy sector. Since 2018, seven offshore wind sites have been decommissioned globally [206], [207]. The Blyth offshore facility represents the first decommissioned in the UK [208]. This represents a new stage in the lifecycle and circular economy of an offshore wind asset and presents unique challenges for decommissioning teams in the harsh operating environment offshore. Of key concern is the embryonic nature of the offshore wind decommissioning infrastructure, with a focus on the disposal and recycling of key asset components. In spite of the high recyclability of wind turbine components, cited as 85% in some cases, the observed current trend displays that the recycling infrastructure in place is poorly equipped to process the volumes of damaged and decommissioned wind turbine blades that require disposal [209]. This has led to the questionable practice of blade burial in landfill sites and, where regulation prevents, their destruction via burning in kilns leading to the atmospheric release of pollutants. This is a contravention of the environmental mission statement of the Paris Agreement, inevitably and unnecessarily increasing the carbon footprint of low carbon energy generation technology. Therefore, to accelerate and secure a circular economy we need to develop large scale recycling solutions for wind turbine blades and composites in general. This will make costs more competitive between disposal and recycling of wind turbine blades [210].

### 8.1.5 TRUSTED DESIGN FOR ROBOTICS

Similar to an offshore wind farm array, robotic platforms have their own lifecycle and remaining useful life. With respect to the timeline of robotic platforms they are designed and manufactured by the robotic companies and then deployed. This features rapid advancement when compared to administrative procedures. Therefore, legislation and regulation of robotic platforms represents a barrier to trusted design for robotics and must gear up in advancement for the wide range of autonomous deployments. These gaps in administrative processes restrict multiple sectors as issues/questions around safety and responsibility in the event of an accident are yet to be determined. This is in addition to a lack of regulation on the effective maintenance of robotic platforms and how they are certified as deployable for autonomous operation both in VLOS and BVLOS scenarios. Figure 34 displays the organization required in the deployment of robotic platforms however, there exists a disparity in the rates of progression for robotic systems versus legislation and regulation resulting in impedance for the design of trusted robotics.

The robotic platform segment features rapid deployment as industry has focused on overcoming challenges from the ground up including navigation, awareness and collision avoidance. This segment can still be improved where robots have the capability to inspect and repair each other to ensure optimal operation. Data driven approaches can also be used to optimize these robots to improve aspects such as battery life, endurance and safety ratings. In terms of the administrative segment, this requires a focus for its development where considerations should be made for resilience, adaptive mission planning to ensure the certification, legislation and regulation of autonomous systems. This will ensure robots operate as intended throughout their lifecycle enhancing safety protocols. Through the adoption of data driven design technologies, the offshore supply chain will make fundamental improvements in the design for reliability for offshore products and services leading to trusted design for robotics.

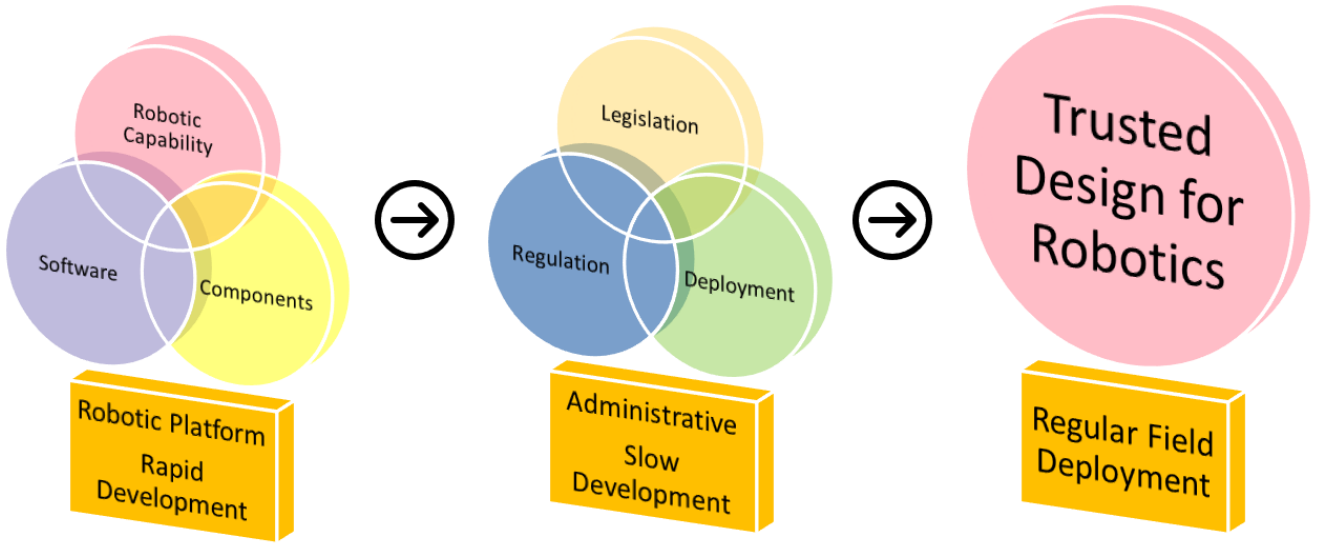

**Figure 34** Design for robotic platforms highlighting robotic platform, administrative and regular field deployment.

## 9 VIEWPOINT OF FUTURE INFRASTRUCTURE

This review presents a future view of offshore infrastructure, which includes robot compliant environments, wireless communication, rack and railed systems, docking stations, cloud platforms and digital infrastructure under a Symbiotic System Of Systems Approach (SSOSA). The digital infrastructure allows for rapid deployment of resident robotic fleets which are all interconnected via a systems engineered symbiotic digital architecture which utilizes knowledge sharing for mutualistic gain. As displayed in Figure 35, a challenge in the future deployment of RAI is ensuring multi-robot fleets have the capability to Collaborate, Cooperate and Corroborate ($C^3$) where $C^3$ governance ensures that these systems can work together to provide value for the ORE sector. Robotic systems provide insight due to the increased mobility to deliver an increased operational overview via knowledge being fed to a DT. DTs have already proven that operators can gain an increased overview of their assets, therefore opportunities exist in remotely obtaining the status of assets including robotic deployment, infrastructure, environment and deployed personnel. This would lead to increases in productivity, safety and operational insights throughout a plant. The early findings as represented in Sections 2-4 (literature review and hype curve) from RAI are promising as it delivers the capabilities we desire and require going forward. This leads to trusted deployment of BVLOS RAI and a hyper-enabled operational overview of the offshore environment as the DT is designed to replicate the offshore environment [143]. However, there still remains a gap - which a SSOSA can fill - where multi-robot fleets are interconnected to give the ability for robots to leverage each other's capability to give positive enforcement to autonomous missions [211].

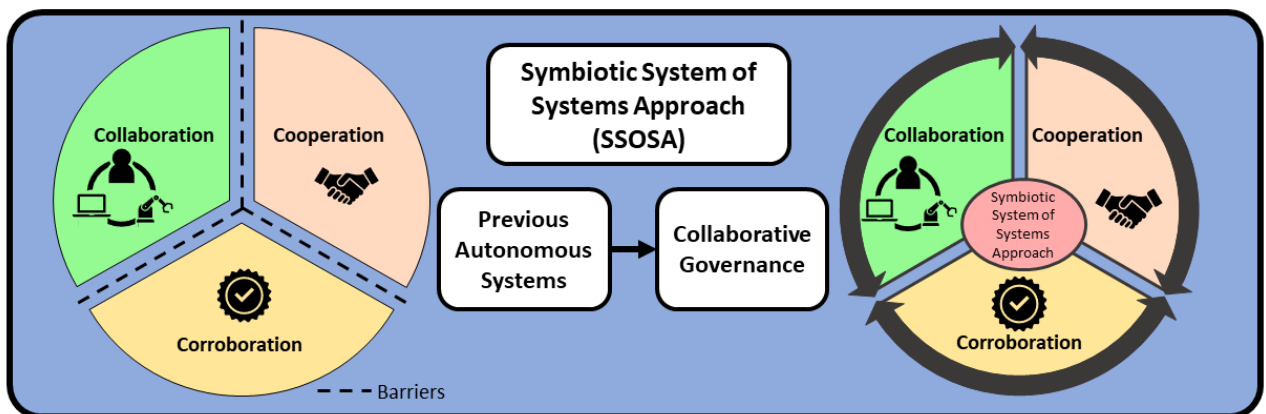

**Figure 35** The challenges in the current deployment of trusted autonomy and the route to achieve collaborative governance [143], [211].

These systems will enable for a holistic view and approach to problems which occur in an offshore wind farm. A view of a future offshore wind farm is illustrated within Figure 36 highlighting the different types of robotic platforms which will be required in an operational wind farm. These systems are deployed to live resident to the wind turbines that they inspect, therefore they are resident to the asset. This allows for rapid and remote deployment of robots for inspection or maintenance. Figure 36 presents the following where a remote operator can plan missions from the shoreline for the various robotic platforms from a DT:

- A- Represents a crawler robot and UAV positioned resident to the wind turbine. The UAV is responsible for completing periodic inspections of the surrounding wind turbines.
- B- Presents a quadruped robotic platform in deployment resident to an offshore substation. The robot can be teleoperated by a human operator or can be assigned autonomous missions to ensure the plant is running smoothly. The platform can routinely read gauges, inspect for problems and map the substation.
- C- Subsea vessels such as ROVs and AUVs are identified in image C. The robotic platforms are resident to the field and complete routine inspections of foundations, cable integrity and cable displacement.
- D- Displays an ASV which completes logistics from the shoreline to the offshore wind farm. The ASV acts as the mothership and base for the deployment of a UAV which then completes an inspection of the wind turbine blade. The UAV has a payload of a camera for surface images of defects and NDE sensor which can detect subsurface defects on the wind turbine blade.
- E- Displays a crawling robotic platform completing an inspection on the wind turbine blade alongside a UAV. The crawler robot can also complete any required remedial maintenance of surface defects.

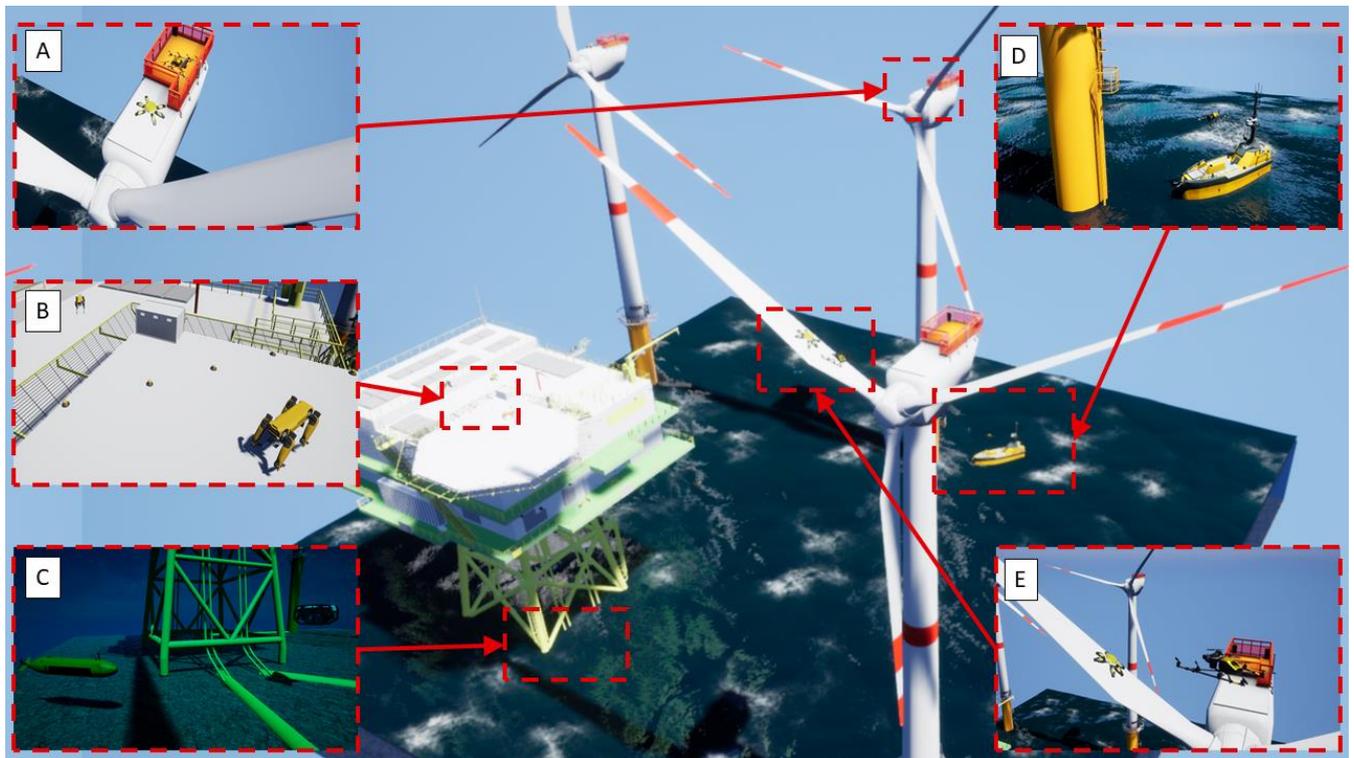

**Figure 36** 3D representation of a view of future offshore infrastructure highlighting symbiotic autonomous assistants in an offshore wind farm where robots are resident to the field or deployed by humans remotely [211].

## 10 DISCUSSION

This paper has reviewed RAI across the lifecycle of an offshore wind farm. A 3D illustration (Figure 5) was presented, which highlighted the challenges currently faced by engineers in offshore wind farms. These challenges include cost, risk and safety due to the dangerous offshore environment. Subsequently, the lifecycle of an offshore wind farm was presented which encompassed several phases from planning to decommissioning. The state-of-the-art was presented alongside the required RAI to complete each phase of the lifecycle of an offshore wind farm. This review highlighted current timeframes of the phases and the RAI utilized for air, land and sea, as displayed in Figure 4, where the following key points were made:

**Timeframes of Lifecycle Phases**- The development of feasible offshore wind farms has led to significant optimization of the *'In Development'* phase to streamline the process of the lifecycle, which has mainly been attributed to the Crown Estate resulting in a more established process in leasing and licensing offshore land.

**Environmental Survey**- There is a drive to push for a more ecologically centric strategy by developing sustainability centric planning. To achieve this, there is a provision to improve the key features of AUVs by ensuring battery endurance and operating depth.

**Wind Farm Planning-** Requiring vessels for large area assessment to autonomously map the seabed and classify objects present in the area. Effectively harnessing this data can result in reduction of accidents, which can cost millions in revenue and increase the awareness of an area where a wind farm array is to be commissioned.

**Wind Turbine Design-** A clear trend is identified as wind farms move further from the shoreline and the swept area of the blades increases to capture higher wind speeds. A key area that has reached the innovation trigger is railed robotics, which can reduce the requirement of human access within nacelles in the future. However, this would require commitment from manufacturers to retrofit the railed robots or include in future design which may be costly.

**Logistics-** All of the sensing technology exists in this area. A key challenge exists in achieving reliable sensor fusion, control algorithms and communication within and across logistics vessels. Academia and industry have supported ASVs to advance their autonomy, resulting in a mature autonomous system when compared to other robotic platforms. However, a barrier exists in allowing for logistics vessels to become efficient offshore. An example would be the level of autonomy and symbiosis between a UAV and a supporting ASV is underdeveloped presenting challenges if a UAV was tasked with inspecting another area on an offshore wind farm. In addition, there is minimal implementation of AI within heavy transfer vessels and SOVs to improve maneuvering and transfer of vessels between wind turbine assets.

**Operational Support-** A wide range of RAI is being deployed; UAVs, crawlers, AUVs, quadrupeds and DTs. These cyber physical systems are harnessing the ability to perform inspection, however, barriers exist in the deployment of these types of robots to perform maintenance missions. Platforms which can overcome this barrier to complete even the smallest of remedial action have the potential to significantly improve the efficiency of wind turbines alongside reducing risk and cost of deploying personnel to complete the remedial action.

**Training-** The offshore sector has inherent risks associated which can result in major injury and loss of life. Many procedures which currently exist offshore rely on human deployment and will likely still exist in the future due to humans working alongside robots. Virtual reality is a safe method to present dangerous scenarios to engineers at all deployment levels, where these vital exercises can be used to verify that training and procedures are effective and safe. This emergent technology can act as an enabler to more thorough training and should be rapidly developed to remove risk via simulated high-pressure scenarios, which engineers may face in the field.

**Decommissioning-** The first generation of offshore wind turbines are reaching the terminal phase of their lifecycle and require decommissioning or replacement. Engineers are faced with evolving challenges due to the unique complexity of the decommissioning phases of offshore infrastructure. Engineering practices from the mature field of decommissioning of O&G assets and can be implemented to the ORE sector. Solutions are also required to ensure the sustainability of the disposal of wind turbine blades, which can be achieved via major upscaling of the recycling or repairing blades for reuse on wind farms.

The state-of-the-art reveals key opportunities within the offshore wind industry. We expect that the ORE sector will benefit from the myriad of industries which are rapidly integrating autonomous systems within their current procedures. The development of RAI in each sector will result in a ripple effect sector-wide, especially as the number of service robots sold increases yearly. In addition to these benefits, the offshore wind sector is being backed by regulation and a number of investment opportunities from the UK government due to Covid-19 recovery packages and COP 26 [212]. This has been supplemented alongside opportunities for small to medium enterprises to scale up their innovations for the offshore wind sector alongside opportunities via new partnerships.

The commercial and patent database key search terms identify knowledge gaps and opportunities within industry. The number of RAI related patents in the ORE sector has increased due to the solutions created through the implementation of robotics. The search identified 2018 as a key year, where an increase in applications for patents in RAI were made which could be attributed to the AI sector deal. This highlighted significant investment from the UK government in RAI for the offshore energy sector. As identified within this review, several areas require development within industry and, until addressed, will restrict offshore deployment of robotics. These include self-certification, resilience, ontologies, trust and DTs representing key enablers in the development of trusted BVLOS autonomy.

Robotics for the ORE sector is gaining increased interest from academics. Similar challenges and opportunities exist within academia and are also represented within industry. Safety and decision making have a large number of published papers for the ORE sector. However, challenges are presented as a number of gaps exist in the research, including AI, ML, UAV, AUV, self-certification, resilience, ontologies, trust, BVLOS and DTs. The incorporation of all the aforementioned technologies under a single systems architecture will represent a SSOSA and will unlock the future potential in the deployment of fully autonomous, resident multi-robot fleets.

As wind turbines become larger, this presents challenges to offshore operators including more inspections and increased risks. Key questions in the deployment of RAI offshore include how to deploy robotic platforms remotely and further from the shoreline? How do we ensure the communication between in house systems and proprietary systems? How do we deal with acquired data and can we streamline data to reduce onboard computational times? How to scale RAI for multiple applications and across a larger area of turbines? This review identifies several robotic systems designed for single applications therefore, a challenge exists in creating interoperable systems. Given all of the available technology, does this restrict the workflow within current and future O&M procedures?

A SSOSA represents the next steps required to seamlessly integrate human-machine partnerships and multiple robotic platforms for mutualistic collaboration. This is reinforced by bidirectional communications, encouraging knowledge sharing across cyber physical systems. A SSOSA will alleviate the difficulties of collecting data from several robotic platforms, which have been designed under their own cloud-based architectures or alternative approaches. A SSOSA presents some challenges in the first instance however, becomes less of an issue once platforms achieve system of systems integration and multi-robot fleets. The benefits include $C^3$ governance, increased operational overview and hyper enabled control, via a fully synchronized real-time DT. AI supports the improvement of accuracy and decision making, wherein a system can make its own decisions relative to the scenario or environment. Improved RAI supports an increase in automated BVLOS missions wherein a robot can be assigned a mission and deployed remotely to an operator. The combination of RAI and a SSOSA leads to advanced decision making for autonomous systems wherein a robot can *'Adapt and Thrive'* when faced with unforeseen challenges. This provides an improved operational overview for a remote operator due to increased accessibility of deployed robot properties such as mission status, inspection status and data collected all within a highly accessible DT which allows for human readable data.

As represented in this paper, the offshore wind sector will be a leading energy supplier for the UK and many other countries. The sector faces a pivotal moment where sufficient experience has been gained from the development to operational phase for improvements to the efficiency of procedures and to shape the future improvement of the lifecycle of offshore assets, which include new procedures such as decommissioning and repowering. RAI provide, a key opportunity to optimize the primary lifecycle, support functions and circular economy of an offshore wind farm. This paper paves the route forward in the roadmap of resident autonomous systems where the analysis completed in this review identifies that many of the RAI are in development or already exist. A key requirement in the future of autonomous offshore systems is how to create a framework from which data can be harnessed for a remote operator onshore. This framework must have the capability to ensure trusted and safe BVLOS autonomous missions alongside an increased operational overview of the deployed robots and offshore assets. This can be achieved via $C^3$ governance to a DT where RAI can cooperate and collaborate through the framework and inspection data can be fed to a DT to corroborate the health status of vital assets and services.

Another key barrier inhibiting the future advancement of autonomous services includes safety and regulation of these autonomous systems. Advancements are being made within robotics however, we need parallel advancements in a new class of certification which verifies and validates the operation of a robot. This will allow for the BVLOS deployments of robots resulting in trusted design for robotics.

## 11 CONCLUSION

This paper presents one of the first systematic comprehensive reviews of RAI for the offshore wind energy sector with analysis into the academic and industrial aspects as well as identification/u of key technology and regulatory barriers, enablers and disruptors. This review identifies risk and cost as the current challenges faced within the current lifecycle of an offshore wind farm and presents future challenges, which restrict the safe role out of fleets of RAI. This review differentiates from others due to the detailed evaluation of regulation, investment and skills development within the offshore wind sector. RAI presents itself in a fast paced, dynamic position where AI assists within many aspects of data analysis, decision making and autonomy.

In conducting this review, a keyword search of patents and publications was performed with analysis of the work in the ORE sector being completed by industry and academia. The depth and analysis of this stage of the review highlighted several areas which require development to ensure the safe deployment of RAI in the offshore environment and include self-certification, ontologies, DTs and resilience.

The technological priorities based on the previously addressed insights were discussed and include the self-certification of autonomous systems, digital twins for human-robot interaction and asset integrity inspection. This highlights significant areas requiring development in achieving a fully autonomous offshore environment where a remote operator can deploy resident robots for O&M whilst achieving an increased operational overview about mission status, failure rates onboard a RAI and the results from the inspected asset.

The implementation of RAI has inherited issues due to a narrow focus on case specific scenarios. To refocus these efforts, a wider view must be implemented by industry and academia to reach a common goal for a SSOSA. This encompasses the seamless integration of data harnessed from sensors, robotics and infrastructure via bidirectional communications to an overarching DT for increased control and operational overview resulting in sector-wide benefits for RAS. Most robotic platforms and AI are already available to implement into procedures in the offshore environment. AI has the potential to provide capability enhancements for automated decision making, ensuring mission resilience and regulatory safety compliance across the offshore wind sector. This has been recognized in industry as the technology is highly scalable and transferrable, resulting in an increase in the number of patents emerging for AI and robotic systems across all relevant fields. However, the disparity between the advancement of the design of robotic platforms versus the regulatory and legislative frameworks is substantial. Two examples include the designation of responsibility if an autonomous system creates damage to an asset or injures personnel. The implementation of autonomy as a service where autonomous systems are continuously updated and serviced under a regulated framework to provide sector-wide commercialization of cyber physical systems.

In the pathway of RAI into offshore wind, a thorough evaluation has been completed by including a review of investment, skills development and regulation alongside a technology centric evaluation. RAI encompasses a wider picture than just safety and reliability as the incorporation of robotics offshore further reduces carbon emissions, improves productivity, reduces costs and increases observability via the orchestration of sensors, robots and assets offshore. The circular economy for RAI in offshore wind farms has been identified alongside challenges and opportunities. A fully symbiotic system will maximize relationship efficiency between humans and machines resulting in trusted BVLOS deployment and optimization of the ORE sector.

## 12 REFERENCES

[1] "Facts about the Climate Emergency | UNEP - UN Environment Programme." https://www.unep.org/explore-topics/climate-action/facts-about-climate-emergency (accessed Oct. 05, 2021).

[2] World Energy Council and University of Cambridge, "Climate Change: Implications for the Energy Sector," Jun. 2014. https://www.worldenergy.org/assets/images/imported/2014/06/Climate-Change-Implications-for-the-Energy-Sector-Summary-from-IPCC-AR5-2014-Full-report.pdf (accessed Nov. 03, 2021).

[3] "Energy and climate change — European Environment Agency." https://www.eea.europa.eu/signals/signals-2017/articles/energy-and-climate-change (accessed Sep. 21, 2020).

[4] "What will a post-pandemic world mean for climate change?" https://www.pwc.co.uk/services/sustainability-climate-change/insights/post-pandemic-world-and-climate-change.html (accessed Aug. 12, 2020).

[5] "UK: Offshore Wind Big in CCC's Progress Report to Parliament | Offshore Wind." https://www.offshorewind.biz/2020/06/26/uk-offshore-wind-big-in-cccs-progress-report-to-parliament/ (accessed Aug. 12, 2020).

[6] "Crown Estate Scotland - Media centre - News, media releases & opinion - Green light for multi-billion pound investment in Scotland's net zero economy." https://www.crownestatescotland.com/media-and-notices/news-media-releases-opinion/green-light-for-multi-billion-pound-investment-in-scotlands-net-zero-economy (accessed Aug. 13, 2020).

[7] "Offshore Wind Generates 9.9% of UK Energy in 2019 | Offshore Wind." https://www.offshorewind.biz/2020/03/27/offshore-wind-generates-9-9-of-uk-energy-in-2019/ (accessed Aug. 12, 2020).

[8] "Drax Electric Insights." https://electricinsights.co.uk/#/dashboard?period=3-months&start=2020-03-12&&_k=ookxqg (accessed Jun. 14, 2020).

[9] I. Staffell, R. Green, T. Green, and M. Jansen, "Electric Insights Quarterly," Accessed: Oct. 05, 2021. [Online]. Available: https://reports.electricinsights.co.uk/wp-content/uploads/2021/05/Drax-Electric-Insights-Q1-2021-Report.pdf.

[10] "Crown Estate offers up 7 GW through Round 4 | 4C Offshore News." https://www.4coffshore.com/news/crown-estate-offers-up-7-gw-through-round-4-nid14203.html (accessed Aug. 13, 2020).

[11] A. B. Wilson and L. Killmayer, "Briefing- Offshore Wind Energy in Europe," 2020. Accessed: Jan. 20, 2021. [Online]. Available: https://www.europarl.europa.eu/RegData/etudes/BRIE/2020/659313/EPRS_BRI(2020)659313_EN.pdf.

[12] European Commiion, "An EU Strategy to Harness the Potential of Offshore Renewable Energy for a Climate Neutral Future," Brusse, 2020. Accessed: Jan. 20, 2021. [Online]. Available: https://ec.europa.eu/environment/nature/natura2000/management/pdf/guidance_on_energy_transmission_infrastr.

[13] Visiongain, "Submarine Power Cables Market Forecast 2020-2030," 2019, [Online]. Available: https://www.visiongain.com/report/global-submarine-power-cable-market-report-2019-2029/.

[14] M. Noonan, G. Smart, and ORE CATAPULT, "The Economic Value of Offshore Wind Benefits to the UK of Supporting the Industry," 2017.

[15] The Guardian and J. Ambrose, "Offshore wind energy investment quadruples despite Covid-19 slump | Environment," 2020. https://www.theguardian.com/environment/2020/jul/13/offshore-wind-energy-investment-quadruples-despite-covid-19-slump (accessed Sep. 21, 2020).

[16] "Climate Change – United Nations Sustainable Development." https://www.un.org/sustainabledevelopment/climate-change/ (accessed Sep. 14, 2020).

[17] HM Government, "The Ten Point Plan for a Green Industrial Revolution," 2020. Accessed: Dec. 19, 2020. [Online]. Available: https://assets.publishing.service.gov.uk/government/uploads/system/uploads/attachment_data/file/936567/10_POINT_PLAN_BOOKLET.pdf.

[18] "Kincardine Floating Offshore Wind Farm, Scotland, UK." https://www.nsenergybusiness.com/projects/kincardine-floating-offshore-wind-farm-scotland/ (accessed Dec. 19, 2020).

[19] Gov.uk, "UK becomes first major economy to pass net zero emissions law - GOV.UK." https://www.gov.uk/government/news/uk-becomes-first-major-economy-to-pass-net-zero-emissions-law (accessed Feb. 14, 2020).

[20] "Offshore wind Sector Deal - GOV.UK." https://www.gov.uk/government/publications/offshore-wind-sector-deal/offshore-wind-sector-deal (accessed Jul. 25, 2020).

[21] Wind Europe, "Offshore Wind in Europe- Key Trends and Statistics 2019," 2019. Accessed: Aug. 14, 2020. [Online]. Available: https://windeurope.org/wp-content/uploads/files/about-wind/statistics/WindEurope-Annual-Offshore-Statistics-2019.pdf.

[22] D. Milborrow, "Big turbines push down O&M costs | Windpower Monthly," 2020. https://www.windpowermonthly.com/article/1682020/big-turbines-push-down-o-m-costs (accessed Aug. 15, 2020).

[23] "ROMEO Targets Offshore Wind O&M Cost Reduction | Offshore Wind," 2017. https://www.offshorewind.biz/2017/06/16/romeo-targets-offshore-wind-om-cost-reduction/ (accessed Aug. 15, 2020).

[24] Offshore Renewable Energy Catapult, "Operations & Maintenance: The Key to Cost Reduction," 2016. Accessed: Jan. 20, 2021. [Online]. Available: https://ore.catapult.org.uk/analysis-insight.

[25] "UN Decade of Ocean Science for Sustainable Development."

[26] "HSE Offshore: Health risks." https://www.hse.gov.uk/offshore/healthrisks.htm (accessed Aug. 13, 2021).

[27]    H. Bailey, K. L. Brookes, and P. M. Thompson, "Assessing environmental impacts of offshore wind farms: Lessons learned and recommendations for the future," *Aquatic Biosystems*, vol. 10, no. 1. BioMed Central Ltd., p. 8, Sep. 14, 2014, doi: 10.1186/2046-9063-10-8.

[28]    "Overcoming Recruitment Challenges in Today's Competitive Hiring Environment - PSG Global Solutions." https://psgglobalsolutions.com/onshore-recruiters-recruitment-challenges/ (accessed Aug. 13, 2021).

[29]    "The future of work in oil and gas | Deloitte Insights." https://www2.deloitte.com/us/en/insights/industry/oil-and-gas/future-of-work-oil-and-gas-chemicals.html (accessed Aug. 13, 2021).

[30]    Bloomberg, "Wind Turbine Blades Can't Be Recycled, So They're Piling Up in Landfills," 2020. https://www.bloomberg.com/news/features/2020-02-05/wind-turbine-blades-can-t-be-recycled-so-they-re-piling-up-in-landfills (accessed Jan. 20, 2021).

[31]    P. Belton, "What happens to all the old wind turbines? - BBC News," 2020. https://www.bbc.co.uk/news/business-51325101 (accessed Jan. 20, 2021).

[32]    D. Mueller, "Getting Offshore Wind Power on the Grid | T&D World," 2019. https://www.tdworld.com/renewables/article/20972636/getting-offshore-wind-power-on-the-grid (accessed Jan. 20, 2021).

[33]    B. W. Tuinema, R. E. Getreuer, J. L. Rueda Torres, and M. A. M. M. van der Meijden, "Reliability analysis of offshore grids—An overview of recent research," *Wiley Interdiscip. Rev. Energy Environ.*, vol. 8, no. 1, p. e309, Jan. 2019, doi: 10.1002/wene.309.

[34]    "Big data: can it reduce the cost of wind turbine operations and maintenance? The need for big data in offshore windfarm management." Accessed: Jan. 20, 2021. [Online]. Available: www.chpv.co.uk.

[35]    D. Mitchell, "Challenges and Opportunities for Artificial Intelligence and Robotics in the Offshore Wind Sector - YouTube," 2021. https://www.youtube.com/watch?v=GAp88iKPK1Q (accessed Dec. 03, 2021).

[36]    "AI Sector Deal - GOV.UK," 2019. https://www.gov.uk/government/publications/artificial-intelligence-sector-deal/ai-sector-deal (accessed Nov. 05, 2020).

[37]    "UK backs robots and smart machines in £22bn r&d plan - Drives and Controls Magazine." https://drivesncontrols.com/news/fullstory.php/aid/6788/UK_backs_robots_and_smart_machines_in__A322bn_r_d_plan.html (accessed Oct. 28, 2021).

[38]    "Government aims for UK to become 'science superpower' with £7m funding for robotics projects | Bdaily." https://bdaily.co.uk/articles/2021/05/25/government-aims-for-uk-to-become-science-superpower-with-7m-funding-for-robotics-projects (accessed Oct. 28, 2021).

[39]    "Greater Levels of Collaborative Robot Safety in the Workplace." https://www.automate.org/blogs/collaborative-robot-safety-trends (accessed Oct. 06, 2021).

[40]    Offshore Robotics for Certification of Assets (ORCA) Hub, "ORCA Robotics - Vision." https://orcahub.org/about-us/vision (accessed Jul. 08, 2020).

[41]    "Financial Times - UK Offshore Wind Infographic." https://equinor.ft.com/infographics/how-offshore-wind-could-power-the-UK-energy-transition.

[42]    "What are the advantages and disadvantages of offshore wind farms? | American Geosciences Institute." https://www.americangeosciences.org/critical-issues/faq/what-are-advantages-and-disadvantages-offshore-wind-farms (accessed Nov. 29, 2021).

[43]    "GE's Giant 13MW Offshore Wind Turbines to Debut at." https://www.oedigital.com/news/481843-ge-s-giant-13mw-offshore-wind-turbines-to-debut-at-dogger-bank-a-b (accessed Oct. 22, 2020).

[44]    "Ørsted Burbo Bank Extension Offshore Wind Farm," 2019, Accessed: Nov. 29, 2021. [Online]. Available: https://orstedcdn.azureedge.net/-/media/www/docs/corp/uk/updated-project-summaries-06-19/190514_ps_burbo-bank-extension-web_aw.ashx?la=en&rev=704980bb76be4f66b08c39b2ac0ce565&hash=249AF9E2446AF47A46F41D5F15F3FB82.

[45]    Gravas Oyvind, "Is wind power's future in deep water? - BBC Future." https://www.bbc.com/future/article/20201013-is-wind-powers-future-in-deep-water (accessed Nov. 29, 2021).

[46]    The Crown Estate and Offshore Renewable Energy Catapult, "Guide to an offshore wind farm," Apr. 2019. Accessed: Sep. 15, 2020. [Online]. Available: www.thecrownestate.co.uk.

[47]    Renewables UK, "Wind Projects Currently in Operation," 2016. Accessed: May 18, 2021. [Online]. Available: www.RenewableUK.com.

[48]    Renewables UK, "Wind Projects Currently in Operation," 2020. Accessed: May 18, 2021. [Online]. Available: www.RenewableUK.com.

[49]    "Wind Energy - RenewableUK." https://www.renewableuk.com/page/WindEnergy (accessed Oct. 28, 2020).

[50]    K. Symon, "ScottishPower plans to invest £10 billion over the next five years to aid the UK's green recovery - Business Insider," 2020. https://www.insider.co.uk/news/scottishpower-plans-invest-10-billion-22964410 (accessed Nov. 05, 2020).

[51]    "OWIC | Sector Deal." https://www.owic.org.uk/osw-sector-deal (accessed Oct. 28, 2020).

[52]    "Offshore Wind Growth Partnership Launches | News | ORE Catapult." https://ore.catapult.org.uk/press-releases/hundreds-of-uk-companies-to-benefit-from-new-initiative-to-maximise-offshore-wind-supply-chain-opportunities/ (accessed Oct. 28, 2020).

[53]    "New guidance for offshore wind industry on geophysical surveys for unexploded ordnance and boulders | Carbon Trust." https://www.carbontrust.com/news-and-events/news/new-guidance-for-offshore-wind-industry-on-geophysical-surveys-for-unexploded (accessed Dec. 18, 2020).

[54]    "Rampion wind farm: Unexploded wartime bombs found in sea - BBC News." https://www.bbc.co.uk/news/uk-england-sussex-35868978 (accessed

Dec. 18, 2020).
[55] "UXB Found Near Offshore Wind Farm Development | Zetica UXO." https://zeticauxo.com/uxb-found-near-offshore-wind-farm-development/ (accessed Dec. 18, 2020).
[56] Alpha Associates Special Risks Consultancy, "Unexploded Ordnance (UXO) Threat & Risk Assessment with Risk Mitigation Strategy for Cable Installation | Project: NorthConnect | Report Number P5530," 2017. [Online]. Available: http://marine.gov.scot/sites/default/files/northconnect_-_uxo_threat.pdf.
[57] Wind Europe, "Offshore Wind in Europe- Key trends and statistics 2019," 2018. Accessed: Feb. 03, 2021. [Online]. Available: https://windeurope.org/wp-content/uploads/files/about-wind/statistics/WindEurope-Annual-Offshore-Statistics-2019.pdf.
[58] "Blyth offshore wind farm to use floating turbines - BBC News." https://www.bbc.co.uk/news/uk-england-tyne-55842657 (accessed Feb. 03, 2021).
[59] Scottish Renewables and Renewable UK, "Floating Wind The UK Industry Ambition," 2019. Accessed: Feb. 03, 2021. [Online]. Available: https://www.scottishrenewables.com/assets/000/000/475/floating_wind_the_uk_industry_ambition_-_october_2019_original.pdf?1579693018.
[60] "BS EN IEC 61400-21-1:2019+A11:2020 - Wind energy generation systems. Measurement and assessment of electrical characteristics. Wind turbines." https://shop.bsigroup.com/ProductDetail/?pid=000000000030428748 (accessed Dec. 18, 2020).
[61] Siemens, "Thoroughly tested, utterly reliable Siemens Wind Turbine SWT-3.6-120." Accessed: Feb. 03, 2021. [Online]. Available: https://www.siemens.com.tr/i/Assets/Enerji/yenilenebilir_enerji/E50001-W310-A169-X-4A00_WS_SWT_3-6_120_US.pdf.
[62] S. Gamesa, "Offshore Wind Turbine SWT-7.0-154 I Siemens Gamesa." https://www.siemensgamesa.com/en-int/products-and-services/offshore/wind-turbine-swt-7-0-154 (accessed Feb. 17, 2020).
[63] "Beatrice Reaches Full Power | Offshore Wind." https://www.offshorewind.biz/2019/05/15/beatrice-reaches-full-power/ (accessed Feb. 03, 2021).
[64] "World's Most Powerful Offshore Wind Turbine: Haliade-X 12 MW | GE Renewable Energy." https://www.ge.com/renewableenergy/wind-energy/offshore-wind/haliade-x-offshore-turbine (accessed Oct. 07, 2020).
[65] "Innovations | Offshore Wind Turbines | MHI Vestas™." https://mhivestasoffshore.com/innovations/ (accessed Oct. 07, 2020).
[66] "Offshore Wind Turbine SG 8.0-167 DD I Siemens Gamesa." https://www.siemensgamesa.com/products-and-services/offshore/wind-turbine-sg-8-0-167-dd (accessed Oct. 07, 2020).
[67] "GE's Giant Haliade-X Offshore Wind Turbine." https://www.oedigital.com/news/482639-ge-s-giant-haliade-x-offshore-wind-turbine-prototype-now-operates-at-13mw?utm_source=AOGDigital-ENews-2020-10-22&utm_medium=email&utm_campaign=OEDigital-ENews (accessed Oct. 22, 2020).
[68] "Offshore Wind Turbine SG 11.0-200 DD I Siemens Gamesa." https://www.siemensgamesa.com/en-int/products-and-services/offshore/wind-turbine-sg-11-0-200-dd (accessed Oct. 07, 2020).
[69] "Offshore Wind Turbine SG 14-222 DD I Siemens Gamesa." https://www.siemensgamesa.com/en-int/products-and-services/offshore/wind-turbine-sg-14-222-dd (accessed Oct. 07, 2020).
[70] "Planit : Job Profiles : Wind Turbine Technician Offshore and Energy." https://www.planitplus.net/JobProfiles/View/782/53 (accessed Oct. 07, 2021).
[71] "Robotic technologies in offshore wind." https://www.power-technology.com/features/robotic-technologies-in-offshore-wind/ (accessed Oct. 07, 2021).
[72] "Servicing complex offshore needs I Siemens Gamesa." https://www.siemensgamesa.com/en-int/products-and-services/service-wind/offshore-logistics (accessed Oct. 27, 2021).
[73] "Life at sea by world's largest offshore wind farm in North Sea - BBC News." https://www.bbc.co.uk/news/av/science-environment-58761725 (accessed Oct. 07, 2021).
[74] "Overcome a sea of challenges - Offshore service logistics - YouTube." https://www.youtube.com/watch?v=RDivnmaBSB4 (accessed Oct. 27, 2021).
[75] D. Lee, S. Oh, and H. Son, "Maintenance Robot for 5MW Offshore Wind Turbines and its Control," *IEEE/ASME Trans. Mechatronics*, vol. 21, p. 1, Oct. 2016, doi: 10.1109/TMECH.2016.2574711.
[76] D. Lattanzi and G. Miller, "Review of Robotic Infrastructure Inspection Systems," *J. Infrastruct. Syst.*, vol. 23(3), no. 04017004, 2017.
[77] GCube Underwriters, "An Insurance Buyer's Guide to Subsea Cabling Incidents," 2015.
[78] European Marine Energy Centre Ltd, The Crown Estate, and UK Government, "PFOW enabling actions project: Sub-sea cable lifecycle study," 2015.
[79] "Report for the Department for Business Energy and Industrial Strategy (BEIS), LessonsLearnt from MeyGen Phase 1a Part 1/3: Design Phase," 2017.
[80] D. Flynn, C. Bailey, P. Rajaguru, W. Tang, and C. Yin, "PHM of Subsea Cables," in *Prognostics and Health Management of Electronics*, John Wiley and Sons Ltd, 2018, pp. 451–478.
[81] F. Dinmohammadi *et al.*, "Predicting Damage and Life Expectancy of Subsea Power Cables in Offshore Renewable Energy Applications," *IEEE Access*, vol. 7, pp. 54658–54669, 2019, doi: 10.1109/ACCESS.2019.2911260.
[82] T. Anderson and M. Rasmussen, "Aging Management: Monitoring of Technical Condition of Aging Equipment," *ICMES, Helsinki*, 2003.
[83] "North East paramedics scale new heights in wind turbine rescue training - ORE Catapult." https://ore.catapult.org.uk/press-releases/north-east-paramedics-scale-new-heights-wind-turbine-rescue-training/ (accessed Oct. 15, 2020).

[84]   Energy & Utility Skills, "SKILLS AND LABOUR REQUIREMENTS OF THE UK OFFSHORE WIND INDUSTRY," 2018. https://greenporthull.co.uk//uploads/files/Aura_EU_Skills_Study_Summary_Report_October_2018.pdf (accessed Nov. 05, 2020).

[85]   "Offshore wind sector set to increase female workforce." https://www.openaccessgovernment.org/offshore-wind-sector-female-workforce/60328/ (accessed Nov. 09, 2020).

[86]   "400,000 new energy workers needed to power UK to net zero | National Grid Group." https://www.nationalgrid.com/uk/stories/community-spirit/400000-new-energy-workers-needed-power-uk-net-zero (accessed Nov. 16, 2021).

[87]   E. Topham and D. McMillan, "Sustainable decommissioning of an offshore wind farm," *Renew. Energy*, vol. 102, pp. 470–480, Mar. 2017, doi: 10.1016/j.renene.2016.10.066.

[88]   K. Branner, Ghadirian, and Amin, "General rights Database about blade faults," 2014. https://backend.orbit.dtu.dk/ws/files/118222161/Database_about_blade_faults.pdf (accessed Jun. 15, 2020).

[89]   "Fiberglass Recycling | Global Fiberglass Solutions." https://www.globalfiberglassinc.com/ (accessed Mar. 01, 2021).

[90]   C. Röckmann, S. Lagerveld, and J. Stavenuiter, "Operation and Maintenance Costs of Offshore Wind Farms and Potential Multi-use Platforms in the Dutch North Sea," in *Aquaculture Perspective of Multi-Use Sites in the Open Ocean: The Untapped Potential for Marine Resources in the Anthropocene*, 2017, pp. 97–113.

[91]   "HOME-Offshore: Holistic Operation and Maintenance for Energy from Offshore Wind Farms." https://gow.epsrc.ukri.org/NGBOViewGrant.aspx?GrantRef=EP/P009743/1 (accessed Oct. 27, 2020).

[92]   "Government announces landmark campaign to inspire next generation of engineers - GOV.UK." https://www.gov.uk/government/news/government-announces-landmark-campaign-to-inspire-next-generation-of-engineers (accessed Nov. 18, 2021).

[93]   "Skills shortage could undo UK government's net zero plans." https://theconversation.com/skills-shortage-could-undo-uk-governments-net-zero-plans-170806 (accessed Nov. 18, 2021).

[94]   "Dangers of Commercial Diving and How to Stay Safe | Divers Institute of Technology." https://www.diversinstitute.edu/dangers-of-commercial-diving-and-how-to-stay-safe/ (accessed Oct. 27, 2021).

[95]   Ø. Netland, I. B. Sperstad, M. Hofmann, and A. Skavhaug, "Concept illustration of a remote inspection robot inside a simplified,... | Download Scientific Diagram," *Energy Procedia*, vol. 53, 2014, doi: 0.1016/j.egypro.2014.07.233.

[96]   G. Hassan, "A Guide to UK Offshore Wind Operations and Maintenance," 2013. Accessed: Oct. 09, 2020. [Online]. Available: http://questfwe.com/wp-content/uploads/2018/02/Wind-farm-operations-and-maintenance-GL-Garrad-Hassan.pdf.

[97]   "World's longest wind turbine blade arrives in the UK! | LM Wind Power." https://www.lmwindpower.com/en/stories-and-press/stories/news-from-lm-places/107-worlds-longest-blade-arrives-in-the-uk (accessed Dec. 14, 2020).

[98]   M. Nancekievill, "Robotics for Harsh Environments - Nuclear Decommissioning," 2020. https://www.wevolver.com/article/robotics.for.harsh.environments.nuclear.decommissioning (accessed Oct. 20, 2021).

[99]   Siemens Global, "Leading as the Digital Enterprise - Electronics Works Amberg Digitalization Practices."

[100]  Siemens Global, "Future technologies are contributing to the success of the Siemens Electronics Works Amberg." https://new.siemens.com/global/en/company/stories/industry/electronics-digitalenterprise-futuretechnologies.html (accessed Feb. 12, 2021).

[101]  F. Insights, "Revolution On The Siemens Factory Floor." https://www.forbes.com/sites/insights-teradata/2019/07/08/revolution-on-the-siemens-factory-floor/?sh=7610dc4d5648 (accessed Feb. 12, 2021).

[102]  "Eight great technologies - GOV.UK." https://www.gov.uk/government/speeches/eight-great-technologies (accessed Feb. 12, 2021).

[103]  "Executive Summary World Robotics 2019 Service Robots." Accessed: Aug. 19, 2020. [Online]. Available: https://www.ifr.org/downloads/press2018/Executive_Summary_WR_Service_Robots_2019.pdf.

[104]  S. Wyatt, S. Bieller, C. Muller, D. Qu, and X. Song, "World Robotics 2019 Service Robots," *IFR Press Conference*, Sep. 18, 2019. https://www.ifr.org/downloads/press2018/IFR World Robotics Presentation - 18 Sept 2019.pdf (accessed Sep. 15, 2020).

[105]  M. Fisher *et al.*, "Verifiable Self-Certifying Autonomous Systems," in *2018 IEEE International Symposium on Software Reliability Engineering Workshops (ISSREW)*, 2018, pp. 341–348, doi: 10.1109/ISSREW.2018.00028.

[106]  "Levels of Autonomous Driving, Explained." https://www.jdpower.com/cars/shopping-guides/levels-of-autonomous-driving-explained (accessed Nov. 23, 2021).

[107]  "Autonomous Underwater Vehicle, HUGIN - Kongsberg Maritime." https://www.kongsberg.com/maritime/products/marine-robotics/autonomous-underwater-vehicles/AUV-hugin/?OpenDocument (accessed Mar. 12, 2021).

[108]  "Naval handling systems Custom naval handling systems for undersea sensors and unmanned vehicles."

[109]  T. Gunnlaugsson and J. Donovan, "Beaufort Sea States." Accessed: Apr. 21, 2021. [Online]. Available: https://www.wdcs.org/submissions_bin/WDCS_Shorewatch_Seastate.pdf.

[110]  Ocean Infinity, "Ocean Infinity pioneers advanced AUV battery technology - Ocean Infinity." https://oceaninfinity.com/2019/11/ocean-infinity-pioneers-advanced-auv-battery-technology/ (accessed Feb. 02, 2021).

[111]  "ROV Observation, Inspection & Subsea Intervention Services." https://www.i-tech7.com/capabilities/assets/rovs (accessed Oct. 15, 2020).

[112]  "Innovative Engineering & Technologies For The Subsea Industry." https://irm.subsea7.com/?start=30 (accessed May 25, 2021).

[113]  I-Tech 7, "Centurion SP Work Class ROV datasheet." Accessed: May 25, 2021. [Online]. Available: www.interventiontechnology.com.

[114]  Ø. Netland, G. D. Jenssen, and A. Skavhaug, "The Capabilities and Effectiveness of Remote Inspection of Wind Turbines," *Energy Procedia*, vol. 80, pp. 177–184, 2015, doi: https://doi.org/10.1016/j.egypro.2015.11.420.

[115]  Ø. Netland, G. Jenssen, H. M. Schade, and A. Skavhaug, "An Experiment on the Effectiveness of Remote, Robotic Inspection Compared to Manned," in *2013 IEEE International Conference on Systems, Man, and Cybernetics*, 2013, pp. 2310–2315, doi: 10.1109/SMC.2013.395.

[116]  Ø. Netland and A. Skavhaug, "Prototyping and evaluation of a telerobot for remote inspection of offshore wind farms," in *2012 2nd International Conference on Applied Robotics for the Power Industry (CARPI)*, 2012, pp. 187–192, doi: 10.1109/CARPI.2012.6473351.

[117]  "Remote and Autonomous Ships The next steps Remote and Autonomous Ship-The next steps."

[118]  "Safer seas with autonomous unmanned vessels for mine countermeasures | Thales Group." https://www.thalesgroup.com/en/united-kingdom/news/safer-seas-autonomous-unmanned-vessels-mine-countermeasures (accessed Oct. 08, 2021).

[119]  "Thales crosses a milestone at sea | Thales Group." https://www.thalesgroup.com/en/united-kingdom/news/thales-crosses-milestone-sea (accessed Oct. 08, 2021).

[120]  "New UK project eyes autonomous vessels in offshore wind - SAFETY4SEA." https://safety4sea.com/new-uk-project-eyes-autonomous-vessels-in-offshore-wind/ (accessed Oct. 15, 2020).

[121]  "C-Worker 7 ASV | L3Harris™ Fast. Forward." https://www.l3harris.com/all-capabilities/c-worker-7-asv (accessed Apr. 21, 2021).

[122]  S. Cheeseman and K. Stefaniak, "The Windfarm Autonomous Ship Project," 2020.

[123]  "Windfarm Autonomous Ship Project (WASP) | Projects | ORE Catapult." https://ore.catapult.org.uk/stories/wasp/ (accessed Mar. 01, 2021).

[124]  "Thales ready for Royal Navy test of its unmanned systems | Thales Group." https://www.thalesgroup.com/en/worldwide/defence/press-release/thales-ready-royal-navy-test-its-unmanned-systems (accessed Oct. 08, 2021).

[125]  "Cyberhawk » Power Generation." https://thecyberhawk.com/power-generation/ (accessed Oct. 20, 2020).

[126]  C. Stout and D. Thompson, "UAV Approaches to Wind Turbine Inspection Reducing Reliance on Rope-Access," 2019.

[127]  "CYBERHAWK SECURES FIVE-YEAR MULTIMILLION-DOLLAR SOFTWARE CONTRACT WITH SHELL." https://insights.thecyberhawk.com/news_and_blog/cyberhawk-secures-five-year-multimillion-dollar-software-contract-with-shell?utm_campaign=2020 Oil %26 Gas Campaign&utm_content=132427757&utm_medium=social&utm_source=linkedin&hss_channel=lcp-987040 (accessed Oct. 20, 2020).

[128]  "MIMRee's Autonomous Inspect and Repair Mission to Offshore Wind Farms | Science & Technology | News." https://www.oceannews.com/news/science-technology/mimree-s-autonomous-inspect-and-repair-mission-to-offshore-wind-farms (accessed Nov. 30, 2020).

[129]  "BladeBUG | Case Studies | Offshore Renewable Energy Catapult." https://ore.catapult.org.uk/stories/bladebug/ (accessed Oct. 20, 2020).

[130]  "BladeBUG: Advanced robotics for turbine maintenance." https://bladebug.co.uk/ (accessed Oct. 20, 2020).

[131]  "First robotic 'blade walk' on a wind turbine opens door to significant cost cuts in offshore renewables - ORE." https://ore.catapult.org.uk/press-releases/bladebug-completes-worlds-first-blade-walk-on-offshore-turbine/ (accessed Nov. 30, 2020).

[132]  P. Chattopadhyay and S. Ghoshal, "Adhesion technologies of bio-inspired climbing robots: A survey," *Int. J. Robot. Autom.*, vol. 33, Nov. 2018, doi: 10.2316/Journal.206.2018.6.206-5193.

[133]  "Robotics in oil and gas: Five developments revolutionising the industry." https://www.offshore-technology.com/features/robotics-oil-gas/ (accessed Nov. 14, 2020).

[134]  "Reshaping Underwater Operations — Eelume." https://eelume.com/ (accessed Nov. 28, 2020).

[135]  P. Liljebäck and R. Mills, "Eelume: A flexible and subsea resident IMR vehicle," in *OCEANS 2017 - Aberdeen*, 2017, pp. 1–4, doi: 10.1109/OCEANSE.2017.8084826.

[136]  "VIDEO: 'World's first' autonomous offshore robot tested | Windpower Monthly." https://www.windpowermonthly.com/article/1497649/video-worlds-first-autonomous-offshore-robot-tested (accessed Mar. 01, 2021).

[137]  "Spot® | Boston Dynamics." https://www.bostondynamics.com/spot (accessed Nov. 17, 2020).

[138]  "Aker BP deploys robotic dog on North Sea FPSO - Offshore Energy." https://www.offshore-energy.biz/aker-bp-deploys-robotic-dog-on-north-sea-fpso/ (accessed Feb. 27, 2021).

[139]  Boston Dynamics, "BP Application of Spot." https://www.bostondynamics.com/spot/applications/bp (accessed Nov. 17, 2020).

[140]  Reuters and Wire Service Content, "Boston Dynamics Dog Robot 'Spot' Learns New Tricks on BP Oil Rig | Technology News | US News," 2020. https://www.usnews.com/news/technology/articles/2020-11-13/boston-dynamics-dog-robot-spot-learns-new-tricks-on-bp-oil-rig (accessed Nov. 17, 2020).

[141]  "JPT Case Study: Drone Technology Inspection of UK North Sea Facility." https://pubs.spe.org/en/jpt/jpt-article-detail/?art=3632&utm_source=newsletter&utm_medium=email-link&utm_campaign=JPT&utm_content=21NOV_DroneInspection&mkt_tok=eyJpIjoiWVdRM00yRmxPR1V3TW1NeCIsInQiOiJwajQxeGdYVSswZUJxcWJVMW9NdTQwQ2xxZ211T1N5dnlkVWNNVSytZODB2RGxkSmdndVJtNnVUQ2l3c0JQU0hUGZQcmlDUkxxSUdVTNjgwRGxYS1ZPMk80UnBHYUFFRMVZKRRkJvSXlRVVlZUWwzYmlwWmlJFWjFaT215TWtFaXXdmeSJ9 (accessed Oct. 20, 2020).

[142]  A. El Saddik, "Digital Twins: The Convergence of Multimedia Technologies," *IEEE Multimed.*, vol. 25, no. 2, pp. 87–92, Apr. 2018, doi: 10.1109/MMUL.2018.023121167.

[143]  D. Mitchell *et al.*, "Symbiotic System of Systems Design for Safe and Resilient Autonomous Robotics in Offshore Wind Farms," Oct. 2021, doi:

[144] "Newcastle College launches virtual reality training | Press releases." https://ore.catapult.org.uk/press-releases/newcastle-college-launches-virtual-reality-offshore-wind-training-facility/ (accessed Oct. 15, 2020).

[145] M. Heggo et al., "The Operation of UAV Propulsion Motors in the Presence of High External Magnetic Fields," *Robot. 2021, Vol. 10, Page 79*, vol. 10, no. 2, p. 79, Jun. 2021, doi: 10.3390/ROBOTICS10020079.

[146] M. Heggo, K. Kabbabe, V. Peesapati, R. Gardner, S. Watson, and B. Crowther, "Evaluation and Mitigation of High Electrostatic Fields on Operation of Aerial Inspections Vehicles in HVDC Environments." 2019, Accessed: Oct. 18, 2021. [Online]. Available: https://www.research.manchester.ac.uk/portal/en/publications/evaluation-and-mitigation-of-high-electrostatic-fields-on-operation-of-aerial-inspections-vehicles-in-hvdc-environments(274426ef-59e2-4dbd-a7a8-2004027ec241)/export.html.

[147] University of Houston Energy Fellows, "Can Robots Transform Offshore Energy? Standardization, Regulations And Workforce Are The Keys," Apr. 15, 2021. https://www.forbes.com/sites/uhenergy/2021/04/15/can-robots-transform-offshore-energy-standardization-regulations-and-workforce-are-the-keys/ (accessed Oct. 21, 2021).

[148] "Understanding Gartner's Hype Cycles | Gartner." https://www.gartner.com/en/documents/3887767/understanding-gartner-s-hype-cycles (accessed Apr. 21, 2021).

[149] J. Blanche, D. Mitchell, R. Gupta, A. Tang, and D. Flynn, "Asset Integrity Monitoring of Wind Turbine Blades with Non-Destructive Radar Sensing," in *2020 11th IEEE Annual Information Technology, Electronics and Mobile Communication Conference (IEMCON)*, 2020, pp. 498–504, doi: 10.1109/IEMCON51383.2020.9284941.

[150] H. Guo, Q. Cui, J. Wang, X. Fang, W. Yang, and Z. Li, "Detecting and Positioning of Wind Turbine Blade Tips for UAV-Based Automatic Inspection," in *IGARSS 2019 - 2019 IEEE International Geoscience and Remote Sensing Symposium*, 2019, pp. 1374–1377, doi: 10.1109/IGARSS.2019.8899827.

[151] A. N. Nichenametla, S. Nandipati, and A. L. Waghmare, "Optimizing life cycle cost of wind turbine blades using predictive analytics in effective maintenance planning," in *2017 Annual Reliability and Maintainability Symposium (RAMS)*, 2017, pp. 1–6, doi: 10.1109/RAM.2017.7889682.

[152] "Espacenet - Advanced search." https://worldwide.espacenet.com/advancedSearch?locale=en_EP (accessed Feb. 10, 2021).

[153] "About Scopus - Abstract and citation database | Elsevier." https://www.elsevier.com/solutions/scopus (accessed Feb. 12, 2021).

[154] "Search for a patent - GOV.UK." https://www.gov.uk/search-for-patent (accessed Feb. 10, 2021).

[155] "UK Digital Strategy - GOV.UK." https://www.gov.uk/government/publications/uk-digital-strategy (accessed Apr. 26, 2021).

[156] "An introduction to unmanned aircraft systems | UK Civil Aviation Authority." https://www.caa.co.uk/Consumers/Unmanned-aircraft/Our-role/An-introduction-to-unmanned-aircraft-systems/ (accessed Aug. 24, 2021).

[157] U. Civil Aviation Authority, "Unmanned Aircraft System Operations in UK Airspace – Guidance," Accessed: Aug. 24, 2021. [Online]. Available: www.caa.co.uk/CAP722.

[158] "Centre for Digital Built Britain |." https://www.cdbb.cam.ac.uk/ (accessed Aug. 24, 2021).

[159] "Home - DT Hub Community." https://digitaltwinhub.co.uk/ (accessed Aug. 24, 2021).

[160] ORCA-Hub, "Research in Robotics, Artificial Intelligence and Autonomous Systems for the Offshore Sector," 2020. Accessed: Oct. 27, 2020. [Online]. Available: https://smartsystems.hw.ac.uk/wp-content/uploads/ORCA-Hub-Booklet-min-2020-Research-in-Robotics-and-Artificial-Intelligence-and-Autonomous-Systems.pdf.

[161] The International Renewable Energy Agency (IRENA), *Offshore innovation widens renewable energy options*. 2018.

[162] M. Barnes et al., "Technology Drivers in Windfarm Asset Management Position Paper," Home Offshore, Jun. 2018. doi: 10.17861/20180718.

[163] "MIMRee explores future of offshore wind O&M | Press Release | ORE Catapult." https://ore.catapult.org.uk/press-releases/mimree-inspection-repair-solution/ (accessed Oct. 09, 2020).

[164] "The MIMREE Project." https://www.mimreesystem.co.uk/project (accessed Nov. 30, 2020).

[165] "Objectives." https://www.mimreesystem.co.uk/objectives (accessed Nov. 30, 2020).

[166] "ORCA Robotics - ORCA Hub secures £2.5m of further funding." https://orcahub.org/articles/orca-hub-secures-2-5m-of-further-funding (accessed Aug. 24, 2021).

[167] "ORCA Robotics - Vision." https://orcahub.org/about-us/vision (accessed Aug. 24, 2021).

[168] "ORCA Robotics - Focus Areas." https://orcahub.org/innovation/focus-areas (accessed Aug. 24, 2021).

[169] M. Andoni et al., "Blockchain technology in the energy sector: A systematic review of challenges and opportunities," *Renewable and Sustainable Energy Reviews*, vol. 100. Elsevier Ltd, pp. 143–174, Feb. 01, 2019, doi: 10.1016/j.rser.2018.10.014.

[170] "New multi-million academic project will research trust in robots - Business Insider." https://www.insider.co.uk/news/new-multi-million-academic-project-22947751 (accessed Nov. 05, 2020).

[171] "Offshore wind takes off in China - China Dialogue." https://chinadialogue.net/en/energy/china-offshore-wind-power-growth/ (accessed Apr. 30, 2021).

[172] "Rystad Energy - China's growth set to help Asia's installed offshore wind capacity catch up with Europe in 2025." https://www.rystadenergy.com/newsevents/news/press-releases/chinas-growth-set-to-help-asias-installed-offshore-wind-capacity-catch-up-with-europe-in-2025/ (accessed Apr. 30, 2021).

[173] "China fueling Asian offshore wind market growth | Offshore." https://www.offshore-mag.com/renewable-energy/article/14188902/china-fueling-

[174] "Special feature-Wind powered electricity in the UK." Accessed: Mar. 15, 2021. [Online]. Available: www.iea.org/reports/offshore-wind-outlook-2019.

asian-offshore-wind-market-growth (accessed Feb. 28, 2021).

[175] "• UK onshore and offshore wind power capacity 2019 | Statista." https://www.statista.com/statistics/240205/uk-onshore-and-offshore-wind-power-capacity/ (accessed Mar. 15, 2021).

[176] I. Pineda, P. Tardieu, Wind Europe, A. Nghiem, and A. Mbistrova, "Wind in power- 2016 European statistics," 2017. Accessed: Mar. 15, 2021. [Online]. Available: https://windeurope.org/wp-content/uploads/files/about-wind/statistics/WindEurope-Annual-Statistics-2016.pdf.

[177] M. Waibel *et al.*, "RoboEarth - A World Wide Web for Robots," *Robot. Autom. Mag. IEEE*, vol. 18, pp. 69–82, Jul. 2011, doi: 10.1109/MRA.2011.941632.

[178] M. Beetz *et al.*, "Robotic agents capable of natural and safe physical interaction with human co-workers," in *2015 IEEE/RSJ International Conference on Intelligent Robots and Systems (IROS)*, 2015, pp. 6528–6535, doi: 10.1109/IROS.2015.7354310.

[179] O. Zaki, M. Dunnigan, V. Robu, and D. Flynn, "Reliability and Safety of Autonomous Systems Based on Semantic Modelling for Self-Certification ," *Robotics* , vol. 10, no. 1. Jan. 03, 2021, doi: 10.3390/robotics10010010.

[180] O. F. Zaki *et al.*, "Self-Certification and Safety Compliance for Robotics Platforms," 2020, p. OTC-30840-MS, doi: 10.4043/30840-ms.

[181] S. Harper, A. Sivanathan, T. Lim, S. McGibbon, and J. M. Ritchie, "Control-display affordances in simulation based education ," *38th Computers and Information in Engineering Conference 2018* , vol. 1B  BT-. American Society of Mechanical Engineers , 2018, doi: 10.1115/DETC2018-85352.

[182] "Wind turbine blade of the year: LM 107.0 P | LM Wind Power." https://www.lmwindpower.com/en/stories-and-press/stories/news-from-lm-places/blade-of-the-year-lm-107 (accessed Dec. 14, 2020).

[183] "Blades - Testing & procedures | GE Renewable Energy." https://www.ge.com/renewableenergy/stories/ge-blades-in-the-wild (accessed Dec. 15, 2020).

[184] "Innovative wind turbine blade manufacturing | GE Renewable Energy." https://www.ge.com/renewableenergy/stories/lm-castellon-wind-turbine-blade-manufacturing (accessed Dec. 15, 2020).

[185] "LM 107m blade arrives at ORE Catapult Blyth | Press Releases." https://ore.catapult.org.uk/press-releases/lm-blade-arrives-ore-catpult-blyth/ (accessed Dec. 15, 2020).

[186] "Turbine Blade Test Facilities - ORE." https://ore.catapult.org.uk/what-we-do/testing-validation/turbine-blade-test-facilities/ (accessed Dec. 17, 2020).

[187] "Riviera - News Content Hub - Diverse propulsion options sought to make crew transfer vessels greener." https://www.rivieramm.com/news-content-hub/diverse-propulsion-options-sought-to-make-crew-transfer-vessels-greener-24217 (accessed Dec. 16, 2020).

[188] "Royal IHC - Offshore wind equipment - Royal IHC." https://www.royalihc.com/en/products/offshore-wind/offshore-wind-equipment (accessed Jun. 09, 2021).

[189] "Autonomous Construction Solutions Advance as Industry Specific Challenges Addressed - Robotics Business Review." https://www.roboticsbusinessreview.com/construction/autonomous-construction-solutions-advance-as-industry-specific-challenges-addressed/ (accessed Jun. 09, 2021).

[190] "Built Robotics Scores $33M for Autonomous Construction Equipment." https://www.roboticsbusinessreview.com/construction/built-robotics-scores-33m-for-autonomous-construction-equipment/ (accessed Jun. 09, 2021).

[191] "The State of the Autonomous Construction Industry | Grading & Excavation Contractor." https://www.gxcontractor.com/technology/article/21121180/the-state-of-the-autonomous-construction-industry (accessed Jun. 09, 2021).

[192] "The challenge of wind turbine blade repair - Renewable Energy Focus." http://www.renewableenergyfocus.com/view/21860/the-challenge-of-wind-turbine-blade-repair/ (accessed Dec. 16, 2020).

[193] Abold and Emily, "Datasheet,Transformers for the offshore transformer module (OTM®) made by Siemens (new)," 2017. Accessed: Dec. 16, 2020. [Online]. Available: https://assets.new.siemens.com/siemens/assets/api/uuid:dafeadace97f7b263942b46ae1229245b8e96321/transformers-for-the-offshore-transformer-module-new.pdf.

[194] D. Mitchell *et al.*, "Symbiotic System of Systems Design for Safe and Resilient Autonomous Robotics in Offshore Wind Farms- Full Video," 2021. https://youtu.be/3YzsEtpQMPU.

[195] D. Mitchell, J. Blanche, S. Harper, W. Tang, T. Lim, and D. Flynn, "Asset Integrity Dashboard Version 2 & Machine Learning – Smart Systems Group." https://smartsystems.hw.ac.uk/asset-integrity-dashboard-version-2/ (accessed Jun. 07, 2021).

[196] D. Mitchell, J. Blanche, S. Harper, T. Lim, and D. Flynn, "Asset Integrity Dashboard Version 1 – Smart Systems Group," 21AD. https://smartsystems.hw.ac.uk/prognostics-and-health-management-phm/ (accessed Jun. 07, 2021).

[197] Smart Systems Group, D. Mitchell, J. Blanche, S. Harper, and D. Flynn, "Asset Integrity Dashboard for FMCW Radar Inspection of Wind Turbine Blades - YouTube," 2020. https://www.youtube.com/watch?v=1-vkn9uRmKA (accessed Feb. 27, 2021).

[198] D. Westwood, "Offshore Wind Driving 2017-2021 Subsea Cable Market Growth | Offshore Wind." https://www.offshorewind.biz/2017/02/24/offshore-wind-driving-2017-2021-subsea-cable-demand/ (accessed Feb. 08, 2021).

[199] J. Warnock, D. McMillan, J. A. Pilgrim, and S. Shenton, "Review of offshore cable reliability metrics," in *13th IET International Conference on AC and DC Power Transmission (ACDC 2017)*, 2017, pp. 71 (6 .)-71 (6 .), doi: 10.1049/cp.2017.0071.

[200]  J. Beale, "Transmission cable protection and stabilisation for the wave and tidal energy industries," *9th Eur. Wave Tidal Energy Confernce, Univ. Southampton, UK*, 2011.

[201]  "Reliable Offshore Power Connection." http://www.electricalreview.co.uk/features/10153-reliable-offshore-power-connections.

[202]  C. Capus, Y. Pailhas, K. Brown, and D. Lane, *Detection of buried and partially buried objects using a bio-inspired wideband sonar*. 2010.

[203]  M. Dmitrieva, M. Valdenegro Toro, K. Brown, G. Heald, and D. Lane, "Object classification with convolution neural network based on the time-frequency representation of their echo," *2017 IEEE 27th International Workshop on Machine Learning for Signal Processing*. IEEE, pp. 1–6, Dec. 07, 2017, doi: 10.1109/MLSP.2017.8168134.

[204]  W. Tang, K. Brown, D. Flynn, and H. Pellae, "Integrity Analysis Inspection and Lifecycle Prediction of Subsea Power Cables," *PHM*. IEEE, pp. 105–114, 2018, doi: 10.1109/PHM-Chongqing.2018.00024.

[205]  W. Tang, D. Flynn, K. Brown, R. Valentin, and X. Zhao, "The Application of Machine Learning and Low Frequency Sonar for Subsea Power Cable Integrity Evaluation," in *OCEANS 2019 MTS/IEEE SEATTLE*, 2019, pp. 1–6, doi: 10.23919/OCEANS40490.2019.8962840.

[206]  "Wind farm No. turbines MW MW/turbine Turbines Year Country Commissioned* Offshore Wind Farms Worldwide."

[207]  World Forum Offshore Wind, "Global Offshore Wind Report- 1st half 2020," 2020. Accessed: Dec. 14, 2020. [Online]. Available: https://wfo-global.org/wp-content/uploads/2020/08/WFO_Global-Offshore-Wind-Report-HY1-2020.pdf.

[208]  "2019 - Blyth decommissioning | The Crown Estate." https://www.thecrownestate.co.uk/en-gb/media-and-insights/stories/2019-blyth-decommissioning/ (accessed Dec. 14, 2020).

[209]  "Wind turbine blades are piling up in landfills - Los Angeles Times." https://www.latimes.com/business/story/2020-02-06/wind-turbine-blades (accessed Dec. 14, 2020).

[210]  "NREL Discusses Circular Economy With High-Profile Partners in Fourth Partner Forum | News | NREL." https://www.nrel.gov/news/features/2021/nrel-discusses-circular-economy-with-high-profile-partners-in-fourth-partner-forum.html (accessed Mar. 01, 2021).

[211]  T. Semwal and F. Iqbal, *CYBER-PHYSICAL SYSTEMS : solutions to pandemic challenges.*, 1st ed. ROUTLEDGE- Taylor & Francis Group, 2022.

[212]  "COP26 Outcomes - UN Climate Change Conference (COP26) at the SEC – Glasgow 2021." https://ukcop26.org/the-conference/cop26-outcomes/ (accessed Nov. 29, 2021).